\documentclass[times,twocolumn,final,authoryear]{elsarticle}
\usepackage{jasr}
\usepackage{framed,multirow}

\usepackage{amssymb}
\usepackage{latexsym}

\usepackage[switch]{lineno}
\usepackage{url}
\usepackage{xcolor}
\definecolor{newcolor}{rgb}{.8,.349,.1}

\usepackage[citebordercolor=white]{hyperref}
\usepackage{jasr}

\usepackage[english]{babel}
\usepackage{amsmath,amsfonts,amssymb}
\usepackage{amsthm}
\usepackage{graphicx, epstopdf}
\usepackage[colorinlistoftodos]{todonotes}

\usepackage{subcaption}
\usepackage{amsfonts}

\usepackage{algorithm}
\usepackage{algpseudocode}
\usepackage[T1]{fontenc}
\usepackage{mathtools}

\usepackage{multirow}
\usepackage{tabularx} % allow set table width
\usepackage{booktabs}
\usepackage{threeparttable}
\usepackage{tablefootnote}

\usepackage{soul}
\usepackage{xspace}
\usepackage{float}

\newcommand*{\etc}{\emph{etc.}\@\xspace}
\newcommand*{\eg}{\emph{e.g.}\@\xspace}
\newcommand*{\ie}{\emph{i.e.}\@\xspace}

\journal{Advances in Space Research}

\begin{document}

\verso{Chee-Kheng Chng \textit{etal}}

\begin{frontmatter}

\title{Direct initial orbit determination}%

\author[1]{Chee-Kheng \snm{Chng}}
\cortext[cor1]{Corresponding author: 
  Email: cheekheng.chng@adelaide.edu.au}
  
\author[2]{Trent \snm{Jansen-Sturgeon}}
\ead{trent.jansen-sturgeon@lmco.com}
%% Third author's email
% 
\author[2]{Timothy \snm{Payne}}
\ead{timothy.m.payne@lmco.com}
\author[1]{Tat-Jun \snm{Chin}\corref{cor1}}
\ead{tat-jun.chin@adelaide.edu.au}

% \fnref{fn1}}
% \fntext[fn1]{This is author footnote for second author.}

\address[1]{Australian Institute for Machine Learning, The University of Adelaide, Adelaide, SA 5000, Australia}
\address[2]{STELaRLab - Lockheed Martin Australia}

% \received{1 May 2013}
% \finalform{10 May 2013}
% \accepted{13 May 2013}
% \availableonline{15 May 2013}
% \communicated{S. Sarkar}

\begin{abstract}
Initial orbit determination (IOD) is an important early step in the processing chain that makes sense of and reconciles the multiple optical observations of a resident space object. IOD methods generally operate on line-of-sight (LOS) vectors extracted from images of the object, hence the LOS vectors can be seen as discrete point samples of the raw optical measurements. Typically, the number of LOS vectors used by an IOD method is much smaller than the available measurements (\ie, the set of pixel intensity values), hence current IOD methods arguably under-utilize the rich information present in the data. In this paper, we propose a \emph{direct} IOD method called D-IOD that fits the orbital parameters directly on the observed streak images, without requiring LOS extraction. Since it does not utilize LOS vectors, D-IOD avoids potential inaccuracies or errors due to an imperfect LOS extraction step. Two innovations underpin our novel orbit-fitting paradigm: first, we introduce a novel non-linear least-squares objective function that computes the loss between the candidate-orbit-generated streak images and the observed streak images. Second, the objective function is minimized with a gradient descent approach that is embedded in our proposed optimization strategies designed for streak images. We demonstrate the effectiveness of D-IOD on a variety of simulated scenarios and challenging real streak images.
%%%
% We evaluated the robustness of D-IOD against different simulated scenarios and challenging real streak images.
%%%%
\end{abstract}


\end{frontmatter}

%% For linenumbers
% \linenumbers
\section{Introduction}\label{sec:intro}
Initial orbit determination (IOD) was proposed more than two centuries ago to determine the orbits of celestial bodies, given their ephemerides. Today, IOD is a key step in tracking Resident Space Objects (RSOs)~\citep{stokes2000lincoln, chambers2016pan, drake2009first}, a capability that contributes to Space Domain Awareness (SDA). SDA is essential for safe utilisation of space, due to the the ever-growing population of RSOs.

\begin{figure}[!htb]
    \centering
    \subfloat{\includegraphics[height=4cm]{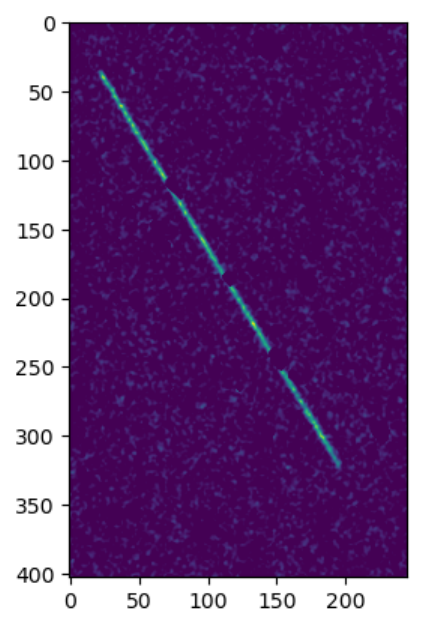}}
    \subfloat{\includegraphics[height=4cm]{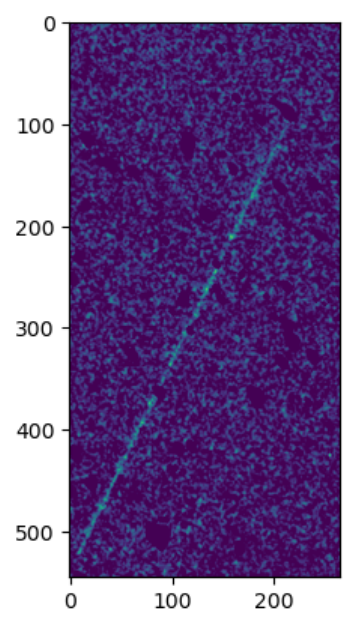}}
    \subfloat{\includegraphics[height=4cm]{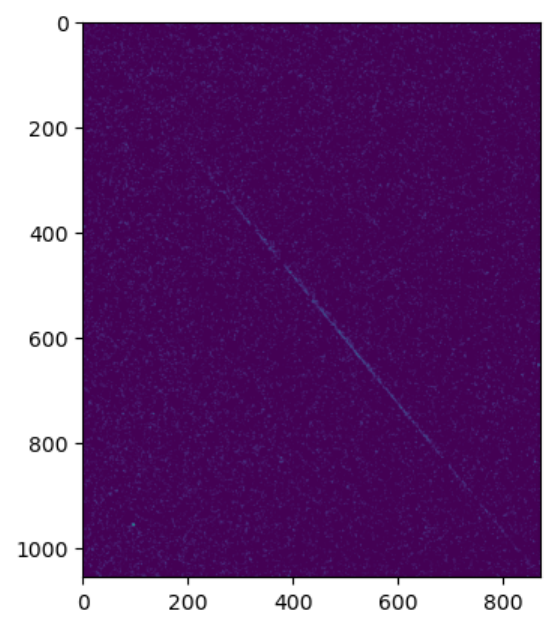}}\\
    \subfloat{\includegraphics[width=6cm]{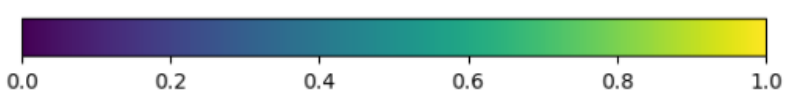}}
    \caption{Real streak images of RSOs captured under a 5-second exposure. The discontinuities in the streaks are due to background star removal (see Sec.~\ref{sec:pre-processing} on pre-processing). For visual clarity, the images are scaled to unity. Note the remaining significant background noise. (Best viewed on electronic devices.)}
    \label{fig:real_chip}
\end{figure}

\begin{figure}[!htb]
    \centering
    \begin{subfigure}[t]{0.5\textwidth}
        {\centering  \includegraphics[width=\textwidth]{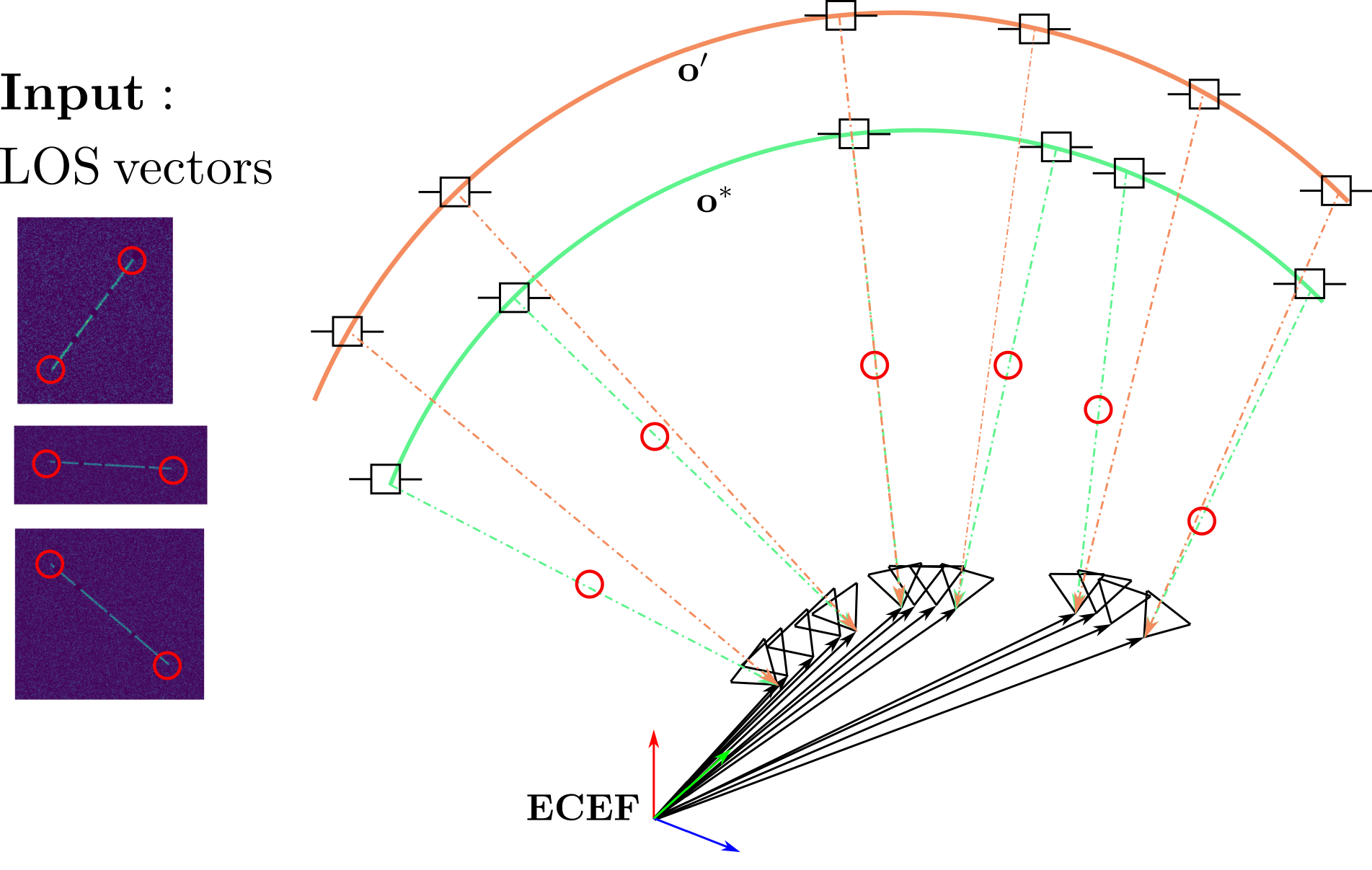}}\caption{}\label{fig:los_pipe}
    \end{subfigure}\\
    \begin{subfigure}[t]{0.5\textwidth}
        {\centering  \includegraphics[width=\textwidth]{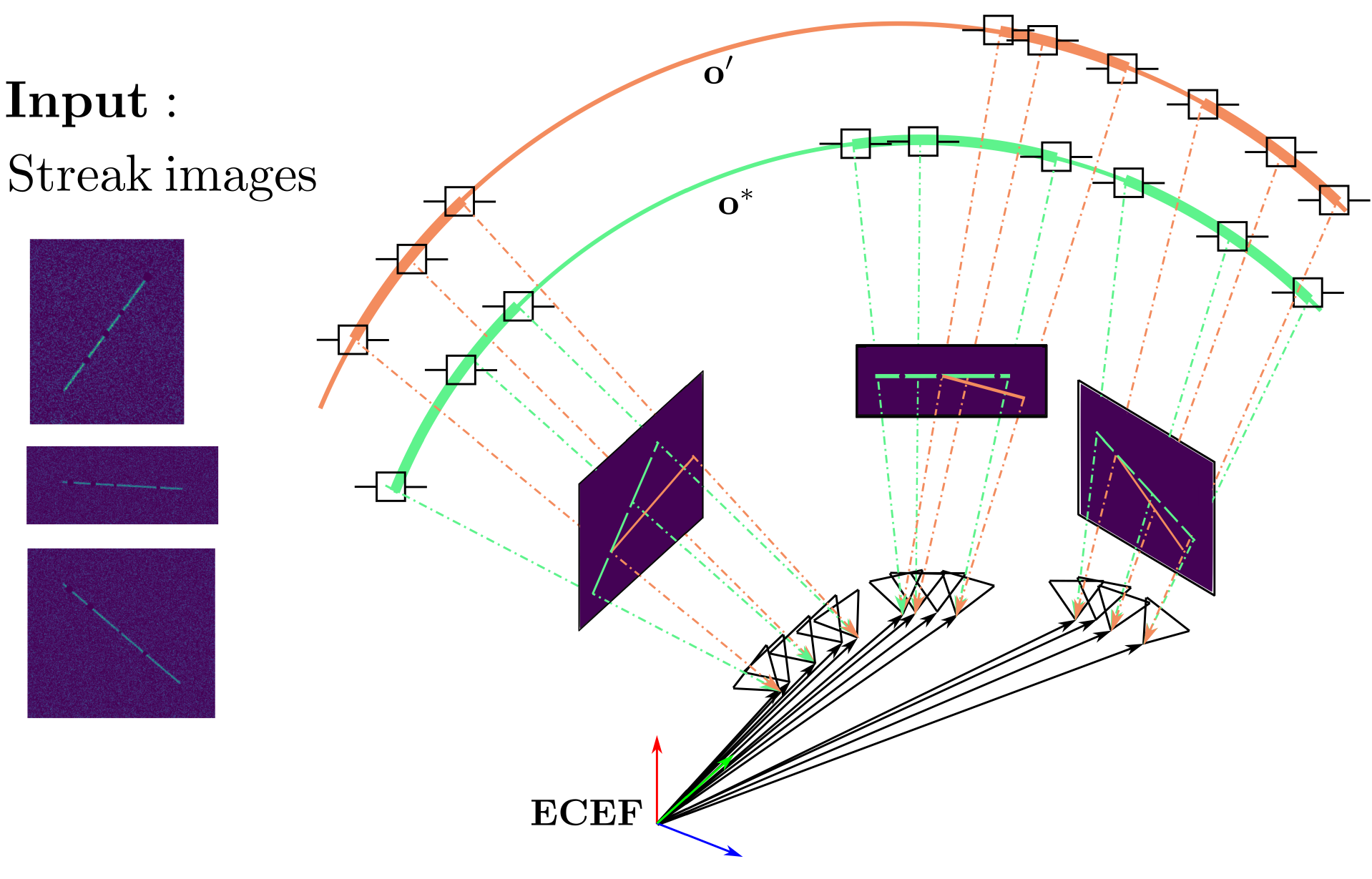}}\caption{}\label{fig:diod_pipe}
    \end{subfigure}\\
    \caption{Comparison of the LOS-based IOD method~(\ref{fig:los_pipe} and D-IOD~(\ref{fig:diod_pipe}). The former first extracts timestamped endpoints and project them to obtain the LOS vectors. Then, the LOS vectors are fed to either a closed-formed IOD solver or an iterative method to determine the optimum orbit ($\mathbf{o}^*$). The depicted method here resembles the Double-r method which iteratively refines the initial estimate ($\mathbf{o}'$). D-IOD, on the other hand, fits the orbital parameters directly onto the streak images without requiring an LOS extraction step.}
    \label{fig:pipeline_comp}
\end{figure}

Many SDA systems employ optical sensors due to their lower cost and ability to capture rich information. When an RSO passes through the field-of-view of a telescope-equipped camera conducting long-exposure imaging, a \textit{streak} is formed in the resulting image. The images that contain streaks (``streak images''; see Fig.~\ref{fig:real_chip}) are the input data to IOD. Most IOD methods, including the classical techniques of Gauss, Laplace, and Double-r~\cite[Chapter~7]{vallado2001fundamentals} as well as more advanced techniques from~\cite{wishnek2021robust} and~\cite{ansalone2013genetic}, operate on Line-of-Sight (LOS) vectors of the RSO. Thus the streak observations must be converted to LOS vectors before the application of the IOD methods. Often, this is accomplished by finding the endpoints of streaks, since these can be associated with the start and end times of exposure. At least three timestamped LOS vectors from multiple images are then fed into the IOD solver to compute an orbit solution. The top row of Fig.~\ref{fig:pipeline_comp} illustrates LOS-based IOD.

\begin{figure}[!]
    \centering
    \subfloat{\includegraphics[width=0.45\textwidth]{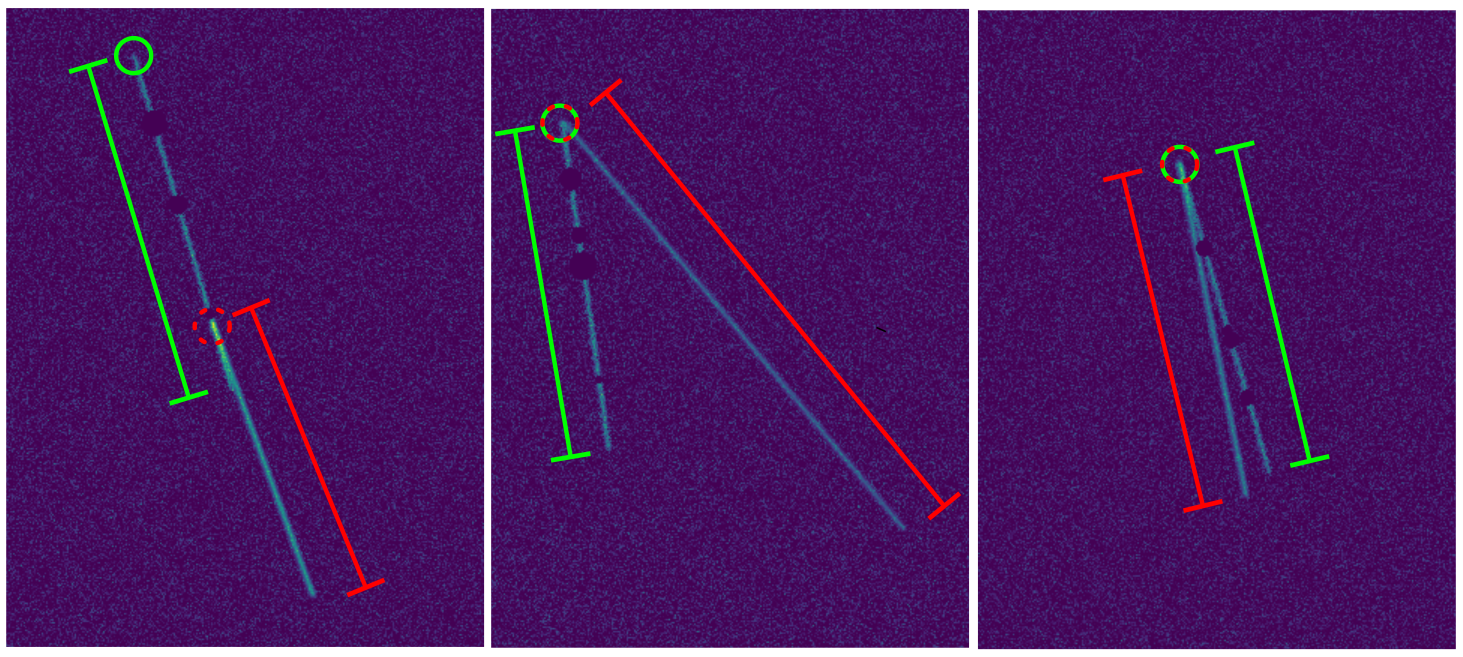}}
    \caption{Projections of a poor orbital solution (highlighted by red lines) on three streak images. The green lines and circles represent the observed streaks and the ground-truth endpoints, respectively. The poor orbital solution is produced by the IOD solver as a result of an ill-localized endpoint (red dashed circle), as depicted in the first figure (left).}
    \label{fig:bad_sol}
\end{figure}

An obvious weakness of the above ``two-stage'' process is that errors in LOS extraction will propagate to the estimated orbit. It is well known that bright stars and low signal-to-noise ratio can greatly challenge the precise detection of the imaged streak~\citep{tagawa2016orbital, levesque2007image, du2022trailed, virtanen2016streak}. Indeed, a single poorly localized LOS can lead to an arbitrarily wrong orbital solution; see Fig.~\ref{fig:bad_sol} for examples.

\paragraph{Contributions} 

In this paper, we introduce a novel Direct Initial Orbit Determination (D-IOD) method. D-IOD fits the orbital parameters directly on the streak images \emph{without} requiring LOS extraction. This is achieved by minimizing the intensity differences between the observed streaks and the generated images of the RSO trajectory propagated from the candidate orbit state vector; see bottom row of Fig.~\ref{fig:pipeline_comp}. By avoiding the usage of LOS vectors, D-IOD is not susceptible to LOS errors. More fundamentally, D-IOD maximizes the use of all available measurements (\ie, the pixel intensities), as opposed to LOS-based IOD that operates on discrete point samples (\ie, the LOS vectors).

D-IOD is inspired by direct image registration techniques~\citep{lucas1981iterative, tomasi1991detection, baker2004lucas} in computer vision that employ all pixels in performing image registration, which stand in contrast to feature-based methods~\citep{nister2004efficient, stewenius2006recent, hartley1992estimation} that utilize higher-level features such as edges, corners and keypoints resulting from a feature extraction step.

\paragraph{Utilization modes and assumptions}

\begin{figure*}
  \centering
  \includegraphics[width=\textwidth]{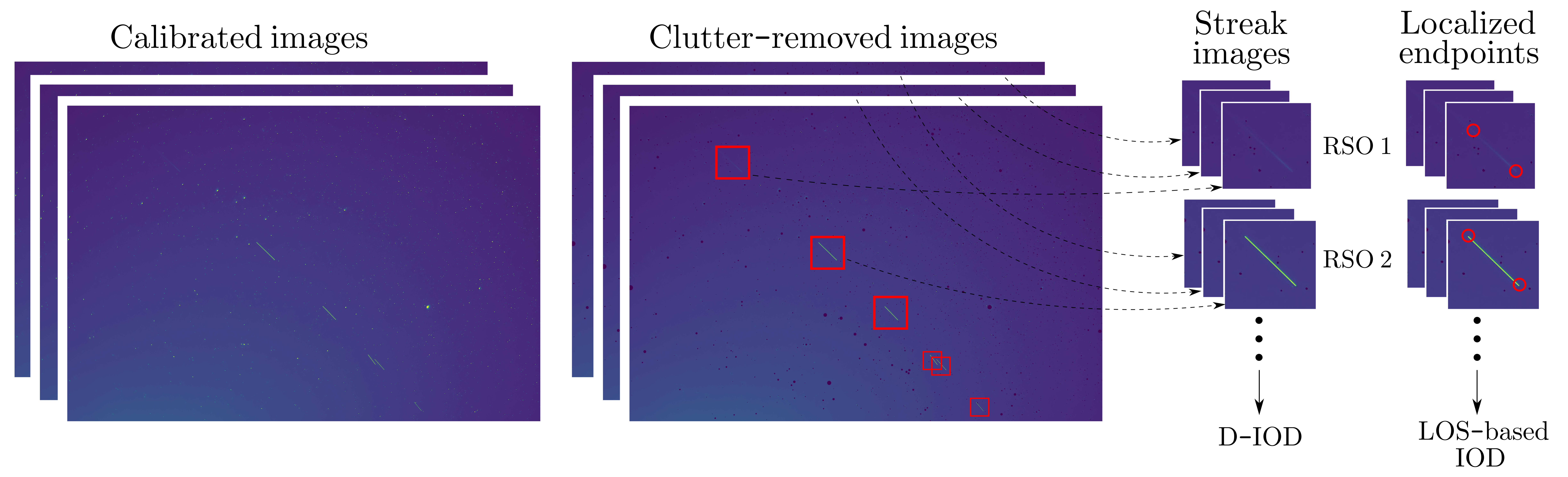}
  \caption{Pre-processing of overall image frames. The red squares are the outputs of streak detection. The (cropped) streak images are associated to different RSOs, denoted as RS0 1 and RSO 2, and then passed on as input to D-IOD. Meanwhile, the endpoints of the streaks are localized for the LOS-based IOD. See text for details.}
    \label{fig:pre_prop}
\end{figure*}

A typical optical sensing pipeline for SDA produces high-resolution images of the night sky that contain clutter (\eg, background stars, clouds) and potentially multiple streaks per frame. In this work, we assume that the ``raw'' images have been preprocessed to extract individual streak images (as exemplified by Fig.~\ref{fig:real_chip}) for D-IOD. Fig.~\ref{fig:pre_prop} depicts the pre-processing steps, while Sec.~\ref{sec:pre-processing} will discuss pre-processing techniques.

Given the streak images with metadata (\ie, timestamps, extrinsic and intrinsic parameters), Two operation modes can be conceived for D-IOD:
\begin{itemize}
    \item End-to-end mode, whereby D-IOD takes only the streak images and outputs an orbit state vector estimate. Enabling end-to-end operation is a novel built-in initialization scheme that constructs a viable initial orbit state vector from the input streak images.
    \item Refine mode, whereby D-IOD takes the streak images and an initial orbit state vector (\eg, a low-accuracy result of an LOS-based IOD method), then conducts direct fitting to polish the initial solution.
\end{itemize}
For D-IOD to yield sensible results, it is assumed that
\begin{itemize}
    \item[A1] The streak images contain the same RSO that has undergone a consistent orbital trajectory.
    \item[A2] Each streak image contains only one streak, as exemplified by Fig.~\ref{fig:real_chip}.
\end{itemize}
Violations to the above assumptions (due to, \eg, erroneous data association, maneuvering RSO, incorrectly localized streaks) will manifest as high fitting error in the D-IOD result without causing program failure.

\paragraph{Paper organization}

Sec.~\ref{sec:survey} surveys related works. Sec.~\ref{sec:problem_formulation} formulates direct orbit fitting, including  streak image generation from propagated orbits and the objective function of D-IOD. Sec.~\ref{sec:D-IOD} describes the optimization algorithm of D-IOD, including strategies to circumvent the difficulty of lack of intensity gradients in streak images. In Sec.~\ref{sec:experiments}, we report the performance of D-IOD under different simulated scenarios, such as different quality of initial estimates, various orbit types, ranging time intervals and signal-to-noise ratios. Additionally, we also showcase the practicality of D-IOD with challenging real streak images. The limitation of D-IOD is presented in Sec.~\ref{sec:limitation}, and Sec.~\ref{sec:conclusion} concludes this paper.

\section{Literature review}\label{sec:survey}
\subsection{Pre-processing}\label{sec:pre-processing}

As mentioned above, optical sensing pipelines for SDA produce raw images of the night sky that contain significant clutter and potentially multiple RSO streaks per image. In general, the first step is to calibrate the raw images with flat-field correction and dark current subtraction. Then, pre-processing of the calibrated images is necessary before IOD can proceed. The major pre-processing steps are as follows.

\begin{itemize}
\item Clutter removal---given an overall image frame, remove (\eg, by zeroing the intensities) blob-like regions corresponding to background stars, galaxy, nebula, \etc.
\item Streak detection---finding individual streaks in the clutter-removed images, where the outputs are typically subimages that contain one streak each.
\item Endpoint localization---given a streak (sub)image, find the endpoints of the streak. Backproject the endpoints to form LOS vectors.
\end{itemize}

Fig.~\ref{fig:pre_prop} illustrates the pre-processing steps. While we distinguish the pre-processing steps above, existing pre-processing methods often conduct all or some of the steps jointly. Below details several representative algorithms.

% Bekte\v{s}evi\'c and Vinkovi\'c

\cite{bektevsevic2017linear} proposed a pipeline that starts from removing background noise with a sequence of image-processing operations (\eg erosion, dilation and histogram equalization). Then, the method runs a combination of Canny edge detection and contour detection to produce bounding boxes for potential streak candidates. Lastly, Hough transformed is executed to obtain the orientation of the streak inside the bounding box. Endpoint localization was not described in the paper.

The ESA-funded streak detection algorithm (StreakDet) \citep{virtanen2016streak} operates on two versions of the image frame - black and white (BW) and grayscale (GS). The method first performs clutter removal and streak detection on the binarized (BW) image to increase computational efficiency. Subsequently, it refines the streak parameters (\ie, localizing the endpoints) in the GS image with a 2D Gaussian point-spread-function fitting method introduced by \cite{verevs2012improved}.

The following methods do not emphasize the clutter removal procedure. However, note that all of them mentioned the usage of some standard noise reduction and removal steps. 

\cite{tagawa2016orbital} proposed a novel technique centered around image shearing and compression to first detect streaks from the full frame image. Then, the detected streak region are handed over to an intensity-thresholding process to localize the endpoints of the streak. \cite{nir2018optimal} leveraged the Fast Radon Transform to detect streaks efficiently. The author reformulated the original Radon Transform to incorporate the endpoints as part of the parameters to be solved for. A similar approach is seen the work of~\cite{cegarra2022real}, where a variant of the Hough Transform, namely the Progressive Probabilistic Hough Transform~\citep{matas2000robust}, is used in performing streak detection and endpoint localization simultaneously.
Filter matching is another line of method that performs both streak detection and endpoint localization simultaneously~\citep{schildknecht2015streak, dawson2016blind, du2022trailed}. In essence, these methods perform convolution over the clutter-removed image with a predefined streak model and seek regions with the maximum response.

More recently, deep-learning-based object detection has been applied to streak detection~\citep{varela2019streak, duev2019deepstreaks, jia2020detection}. Such methods can accurately estimate the bounding boxes of streaks in the overall frame, even for faint streaks. Unlike the Hough/Radon-transform and image processing approaches, deep-learning-based object detection recovers the streak images independently of streak orientation estimation and endpoint localization.

We highlight that endpoint localization is a necessity for the LOS-based IOD methods, as depicted in Fig.~\ref{fig:pipeline_comp}. In contrast, D-IOD only require loosely cropped streak images as input. Since D-IOD fits segments of the candidate orbit to streaks, it also achieves endpoint localization as a byproduct. As alluded to above, conducting direct orbit fitting maximises the use of all available data and prevents premature endpoint localization that is inconsistent with the orbital motion.

\subsection{IOD methods}

\paragraph{Classical IOD methods}
Gauss's and Laplace's analytical solutions have been regarded as the first two practical algorithms for the IOD problem \citep{vallado2001fundamentals}. Both methods were developed for the type of data available back then---LOS vectors from slow-moving celestial bodies acquired with a sextant. Three linearly independent LOS vectors are required to determine the six degree-of-freedom (DOF) orbital parameters since each LOS vector has two DOF. Modern numerical methods such as Double-r and Gooding's method \citep{gooding1996new} have fewer restrictions on the measurement geometry.

To deal with inaccurate LOS vectors, \cite{der2012new} presented a Double-r method that allows the LOS vectors to be jointly optimized with the range parameters. The superiority of the Double-r method over classical IOD solvers is clearly demonstrated in cases with LOS errors. Our proposed D-IOD is also not affected by LOS errors; however, D-IOD refines the orbital estimate based on differences in the image intensities directly instead of adjusting the LOS vectors.

\paragraph{Data association and too-short-arc problem}

Too-short-arc (TSA) measurements capture segments of an orbit that are too short geometrically to yield a reliable orbit estimate, especially if the measurements are noisy~\citep{gronchi2004classical}. It is essential to collect more measurements over time---by progressively associating new measurements to previous measurements---to improve the numerical accuracy of the resulting orbit. On the other hand, data association is informed by knowledge of the orbit. This chicken-and-egg problem---called the TSA problem---motivates solving data association and IOD jointly.

\cite{milani2004orbit} proposed one of the earliest methods to solve the TSA problem. The authors presented a method to determine an Admissible Region (AR) based on a set of physical constraints, e.g., orbits with negative energies. Each point in the region is an orbital solution candidate, which enables propagation, in turn allowing data association. 

The AR-based method is well-received by the community and has been furthered by numerous works. \cite{maruskin2008orbit} discussed a conceptual algorithm for intersecting two AR regions to eliminate infeasible solution candidates before the subsequent (expensive) least square orbit correction process. \cite{fujimoto2012correlation} proposed a technique to correlate tracks with probability distributions in the Poincaré space to improve computational efficiency. \cite{fujimoto2014boundary} made another attempt to reduce the computational cost by incorporating an extra data domain - angle-rate. The added constraint reduces the number of minimal track associations needed from three to two. Meanwhile, \cite{demars2013probabilistic} incorporated the Gaussian mixture models to approximate the admissible region, which allows subsequent refinement when new data is available. 

\cite{gronchi2010orbit} proposed a closed-formed solution to the data association problem via two-body integrals. The authors further improved the original 48-degree polynomial equation to 20 degrees and 9 degrees in later works \citep{gronchi2011orbit,gronchi2015orbit}. The reduction of polynomial degrees is accompanied by an improvement in computational efficiency.

Note that assumption A1 for D-IOD described in Sec.~\ref{sec:intro} implies that data association has been solved prior to the method. However, noting that the best-fit orbit returned by D-IOD would yield a high loss given a set of ill-associated streak images, it is possible to extend D-IOD to solve the TSA problem. We leave this as future research.

\section{Problem formulation}\label{sec:problem_formulation}

We formulate the orbit fitting problem in this section. We first discuss the given data in Sec.~\ref{subsec:data}, followed by the interpolation of timestamps (required by the modeling) in Sec.~\ref{subsec:timestamps}. Then, we present the modeling of an intensity image as a function of the initial state vector in Sec.~\ref{subsec:chip_modelling}. This section ends with the objective function of D-IOD in Sec.~\ref{subsec:orbit_fitting_formulation}.

\subsection{Data}\label{subsec:data}
The data for our problem are: 

\begin{enumerate}
    \item A set of streak images (as seen in Fig.~\ref{fig:real_chip}),
    \item The starting and ending timestamps of each image throughout the long-exposure imaging process, and
    \item The extrinsic and intrinsic parameters of the telescoped-equipped camera in-used. 
\end{enumerate}

Each streak image is denoted as $\mathbf{D}^{(m)} \in \mathbb{R}^{X_{m} \times Y_{m}}$. The extrinsic parameters include the camera location in the Earth-Centered, Earth-Fixed (ECEF) coordinates and the pointing direction. We denote the extrinsic parameters at the starting timestamp of each image ($t^{(m)}_0$)  as $\mathcal{E}^{(m)}_{t_0} \coloneqq \{\mathbf{r}^{(m)}_{t_0} \in \mathbb{R}^3 , \mathbf{a}^{(m)}_{t_0} \in \mathbb{R}^3, \parallel \mathbf{a}^{(m)}_{t_0} \parallel = 1\}$, where $\mathbf{r}$ is the camera location and $\mathbf{a}$ is the pointing direction of the camera. The intrinsic parameters include the focal length of the telescope, distortion parameters, pixel scale, etc. These are all provided in a FIT file as a standard practice in Astrometry~\citep{calabretta2002representations}. We denote these constant intrinsic parameters as $\mathcal{I}$. These parameters are used to perform pixel-to-world and world-to-pixel projections (more details in Sec.~\ref{subsubsec:img_projection}).

\begin{figure}[!]
    \centering
    \subfloat{\includegraphics[width=0.5\textwidth]{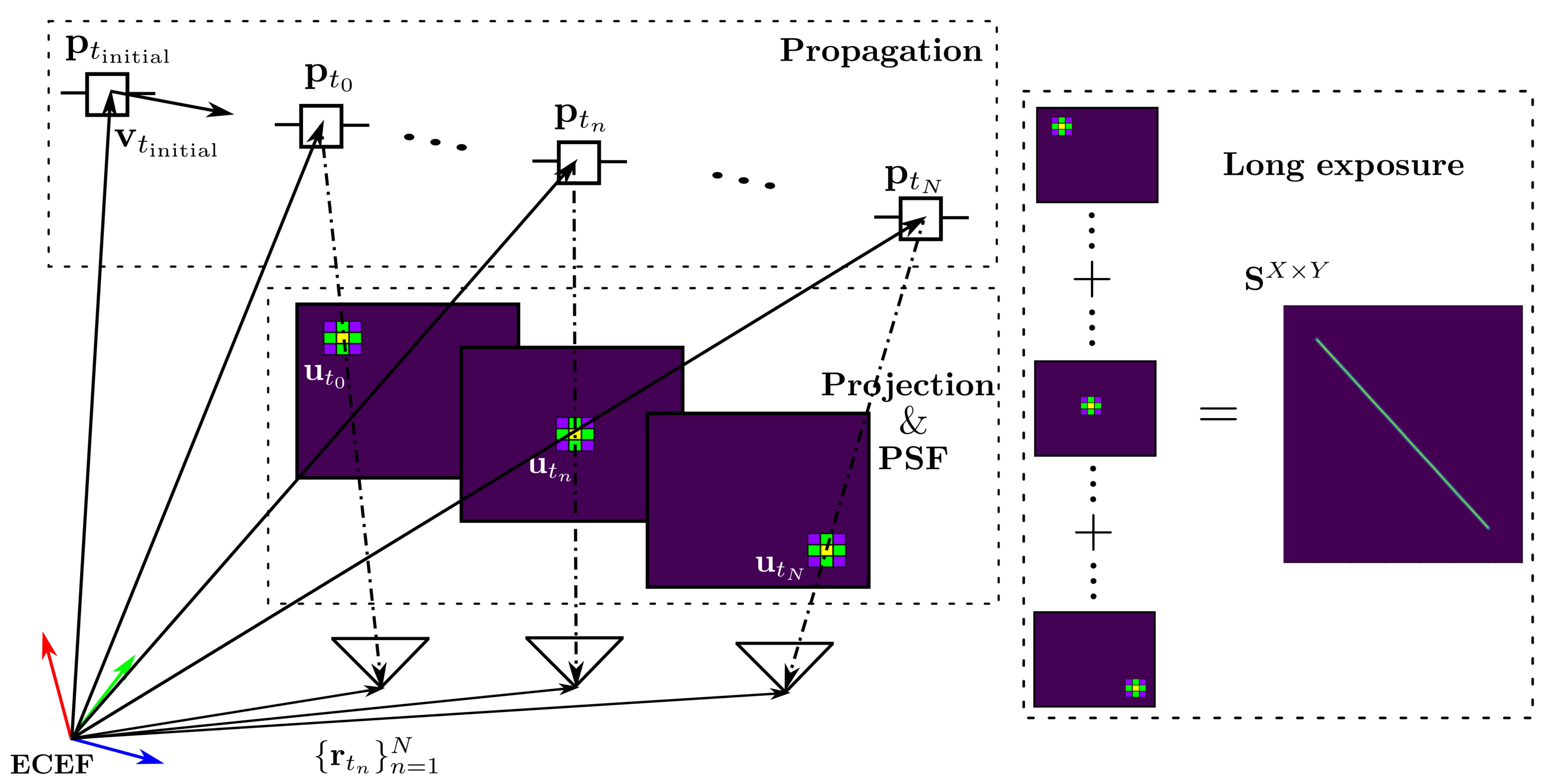}}
    \caption{Our proposed model that maps the initial state vector ($\mathbf{p}_{t_{\rm{initial}}}$ and $\mathbf{v}_{t_{\rm{initial}}}$) to one of the intensity images ($\mathbf{S} \in \mathbb{R}^{(X \times Y)})$. Each propagated position vector ($\mathbf{p}_{t_n}$) is projected to the pixel coordinate $\mathbf{u}_{t_n}$ given the camera location $\mathbf{r}_{t_n}$ and other camera parameters (see text). The highlighted $3\times3$ matrices (next to $\mathbf{u}_{t_0}$, $\mathbf{u}_{t_n}$, and $\mathbf{u}_{t_N}$) depicts the point spread function (PSF). The long-exposure imaging process is modelled by the summation of all intensity images corresponding to the discretized timestamps.}  
    \label{fig:chip_modelling}
\end{figure}

\subsection{Interpolation of timestamps}\label{subsec:timestamps}
We need to interpolate the \textit{in-between} timestamps since the imaged streak is a function of time. Specifically, the RSO state, the camera position, and the accumulation of photons at each pixel bin change continuously within the time exposure window.

\begin{eqnarray}\label{eqn:timestamp_inter}
    \tau^{(m)} \coloneqq \{t^{(m)}_n \: \vert \: t^{(m)}_n = \frac{n \: \Delta t^{(m)}}{d^{(m)}} + t_0 \: , n = 0, 1, ..., N  \} \: .
\end{eqnarray}

Given $\tau$, we obtain a set of camera extrinsic parameters $\{\mathcal{E}_{t^{(m)}_n}\}^N_{n=0}$ as a function of Earth's motion, i.e., position, rotation, nutation and precession~\citep{vallado2001fundamentals}. We employ an existing off-the-shelf tool, Astropy~\citep{price2018astropy}, for this task.

\subsection{Modeling}\label{subsec:chip_modelling}
We model the streak image as a function of the initial state vector $\mathbf{o}_{\rm{initial}} \coloneqq \{\mathbf{p}_{{t_{\rm{initial}}}} \in \mathbb{R}^3, \mathbf{v}_{{t_{\rm{initial}}}} \in \mathbb{R}^3\}$ as follows,

\begin{eqnarray}\label{eqn:image_formation}
    \mathbf{S}^{(m)} = F(\mathbf{o}_{t_{\rm{initial}}}; \mathbf{C}^{(m)})
\end{eqnarray}

\noindent where $\mathbf{S}^{(m)} \in \mathbb{R}^{X_{m} \times Y_{m}}$ is an intensity matrix, and $\mathbf{C}^{(m)}$ contains all the constant variables, e.g., timestamps ($t^{(m)}_n$),  extrinsic and intrinsic camera parameters ($\mathcal{E}_{t^{(m)}_n}$ and $\mathcal{I}^{(m)}$), and the standard deviation of the Gaussian point spread function ($\sigma^{(m)}$), etc., which are detailed in the following sections. The composition of functions in $F$ are as follows.

\begin{enumerate}
  \item The propagator function, i.e., $\mathbf{o}_{t^{(m)}_n} = P(\mathbf{o}_{t_{\rm{initial}}}, t^{(m)}_n)$.
  \item The world-to-image projection function, i.e., $\mathbf{u}_{t^{(m)}_n} = W(\mathbf{o}_{t^{(m)}_n}; \mathcal{E}_{t^{(m)}_n}$, $\mathcal{I}^{(m)})$.
  \item The point spread function, i.e., ${g_{xy}}_{t^{(m)}_n} = G(\mathbf{u}_{t^{(m)}_n}; \mathbf{u}_{xy}, \sigma^{(m)})$.
  \item The long-exposure imaging process, i.e., $s^{(m)}_{xy} = E(\{{g_{xy}}_{t^{(m)}_n}\}^N_{n=0})$.
\end{enumerate}

\noindent Fig.~\ref{fig:chip_modelling} depicts our model. The $m$ notations are dropped for compactness in the rest of the modelling subsections. 

\subsubsection{Keplerian propagator}\label{subsubsec:propagator}
The innermost function in $F$ propagates the initial state vector to timestamp $t_n$. The function is illustrated in Fig.~\ref{fig:chip_modelling}, where $\mathbf{p}_{t_{\rm{initial}}}$ and $\mathbf{v}_{t_{\rm{initial}}}$ are propagated to a set of $\mathbf{p}_{t_n}$ and $\mathbf{v}_{t_n}$. We adopt the standard Keplerian propagation model~\cite[Chapter~2]{vallado2001fundamentals} that incorporates gravitational effects only.

\subsubsection{Gnomonic projection}\label{subsubsec:img_projection}
The next function is the projection of RSO positions to the pixel coordinates of the camera given its parameters ($\mathcal{E}_{t_n}$ and $\mathcal{I}$). As illustrated in Fig.~\ref{fig:chip_modelling}, the projected pixel coordinates are labelled with their timestamps, i.e., $\mathbf{u}_{t_n} \in \mathbb{R}^2$. We adopt the standard gnomonic projection model~\citep{calabretta2002representations} (implemented by Astropy~\citep{price2018astropy}) for this task.

\subsubsection{Gaussian point spread function}\label{subsubsec:psf}
We model the spread of the photons on the image plane with the Gaussian point spread function. The intensity at location $x, y$ at timestamp $t_n$ can be computed with the following expression, 

\begin{eqnarray}
    {g_{xy}}_{t_n}(\mathbf{u}_{t_n}; \mathbf{u}_{xy}, \sigma) = \frac{1}{\sigma\sqrt{2\pi}} \exp \bigg( - \frac{\parallel \mathbf{u}_{xy} - \mathbf{u}_{t_n} \parallel_2}{2\sigma^2} \bigg) \: ,
\end{eqnarray}

\noindent where $\sigma$ is the width of the spread that can be determined from the imaged streak. As illustrated in Fig.~\ref{fig:chip_modelling}, the intensity level decreases as the distance of the pixel with the projected pixel $\mathbf{u}_{t_n}$ increases. We highlight that our model assumes constant brightness along the streak. It does not consider the brightness variation in the streak caused by the rotation of RSOs.

\subsubsection{Long-exposure imaging}\label{subsubsec:long_exp}
The intensity of each pixel is accumulated over $N$ discretized timestamps to model the long-exposure imaging process, yielding

\begin{eqnarray}
    s_{xy} = \sum^N_{n=0} {g_{xy}}_{t_n} \: .
\end{eqnarray}

\noindent The summation process of the intensity matrices is illustrated in Fig.~\ref{fig:chip_modelling} as well. 

\subsection{Orbit fitting formulation}\label{subsec:orbit_fitting_formulation}
We are now ready to present the formulation of D-IOD's orbit fitting problem. The optimization problem aims to find the initial state vector ($\mathbf{o}_{t_{\rm{initial}}}$) at $t_{\rm{initial}}$ that minimizes the deviations between the observed streak images $\{\mathbf{D}^{(m)}\}^M_{m=1}$ and the generated streak images $\{\mathbf{S}^{(m)}\}^M_{m=1}$. Formally, it has the following form,

\begin{eqnarray}\label{eqn:img_fitting}
    \underset{\mathbf{o}_{t_{\rm{initial}}} \in \mathbb{R}^6}{minimize} \sum_{m=1}^{M} L^{(m)}(\mathbf{o}_{t_{\rm{initial}}}),
\end{eqnarray}

\noindent where each of the loss terms ($L^{(m)}$) is the mean Frobenium norm ($\parallel . \parallel_F$) of the difference between $\mathbf{S}^{(m)}$ and $\mathbf{D}^{(m)}$, as expressed below,

\begin{eqnarray}\label{eqn:obj_func}
    L^{(m)}(\mathbf{o}_{t_{\rm{initial}}}) \coloneqq \ \frac{1}{\vert \mathbf{D}^{(m)} \vert} \ \parallel \mathbf{S}^{(m)}(\mathbf{o}_{t_{\rm{initial}}}; \mathbf{C}) - \mathbf{D}^{(m)} \parallel_F ,
\end{eqnarray}

\noindent where the cardinality operation ($\vert \mathbf{D}^{(m)} \vert$) returns the number of pixels in $\mathbf{D}^{(m)}$.

In general, $t_{\rm{initial}}$ can be set to any arbitrary timestamp. However, setting it to the middle timestamp between the furthest pair of images, i.e., $t_{\rm{initial}} \coloneqq \frac{t^{(1)}_0 + t^{(M)}_0}{2}$,  has a practical advantage that is detailed in Sec.~\ref{subsec:hyperparams}. The geometrical relationship of D-IOD's orbit fitting problem is visualized in the bottom row of Fig.~\ref{fig:pipeline_comp}.

\begin{figure}[!]
    \centering
    \subfloat{\includegraphics[width=0.5\textwidth]{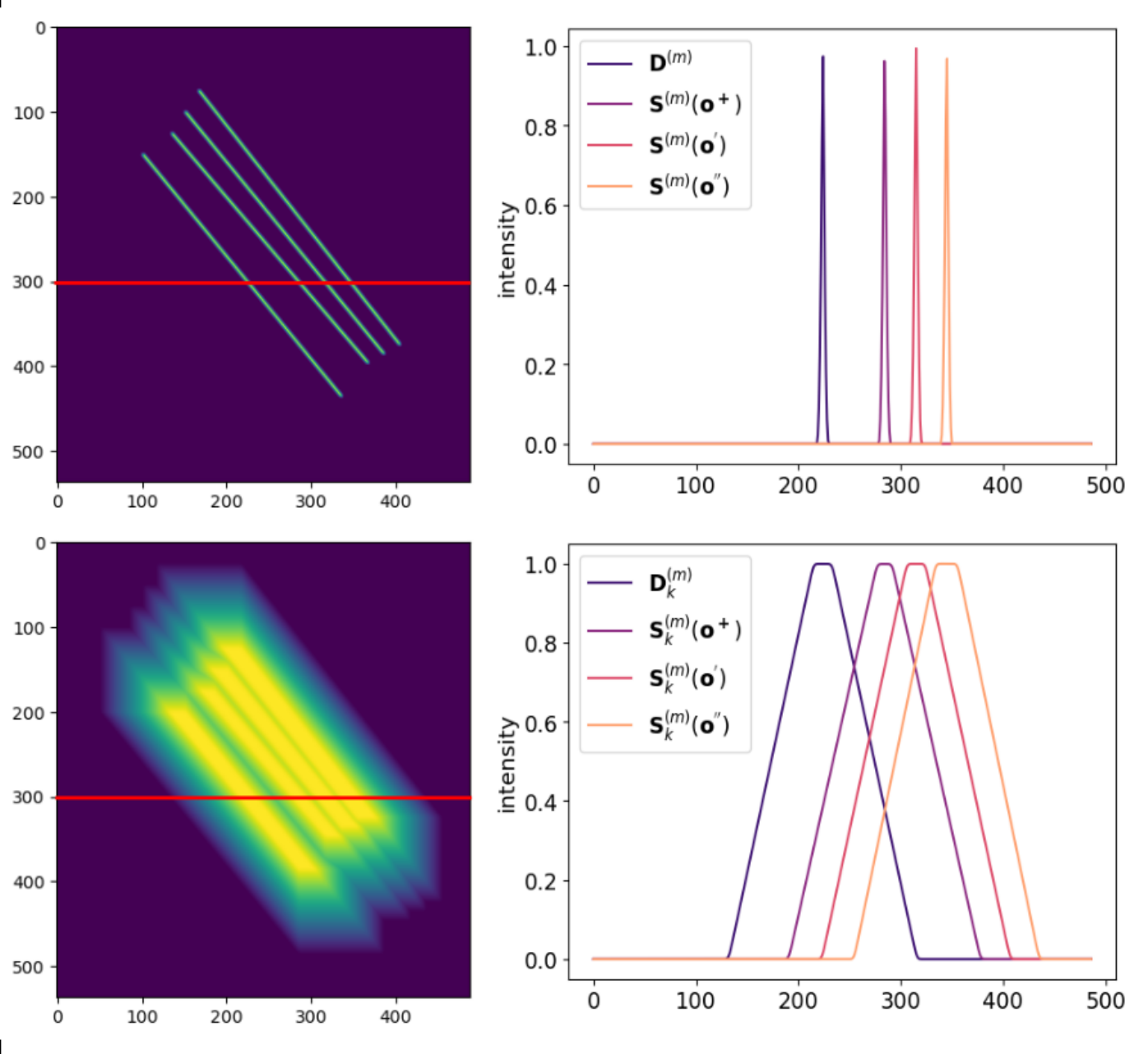}}
    \caption{The blurring operation enlarges the streak region in the image. \textbf{Top row}: The left image shows the stacking of the observed streak image ($\mathbf{D}^{(m)}$) and the streak images generated with three different estimates, $\mathbf{S}^{(m)}(\mathbf{o}^+)$, $\mathbf{S}^{(m)}(\mathbf{o}')$, and $\mathbf{S}^{(m)}(\mathbf{o}'')$. The 1D (intensity) signals on the right plot are extracted from the same (red) row in each image. \textbf{Bottom row}: The blurred correspondences of the top row.}\label{fig:blurring}
\end{figure}

\section{D-IOD}\label{sec:D-IOD}
We present the algorithmic details of D-IOD in this section. We solve the proposed non-linear least squares problem \eqref{eqn:img_fitting} with a gradient descent approach. In Sec.~\ref{subsec:opt_strat}, we highlight several optimization strategies that we employed. Then, we detail our data pre-processing steps in Sec.~\ref{subsec:data_pre} and the gradient descent method in Sec.~\ref{subsec:gd}. All steps of D-IOD are summarized in Alg.~\ref{alg:ctf}.

\subsection{Optimization strategies}\label{subsec:opt_strat}
\subsubsection{Image blurring}\label{subsec:image_blurring}
Intuitively, gradient descent algorithms determine the best direction (also known as the negative gradient vector) in the domain space to travel based on the current estimate's neighbouring loss landscape. However, the \textbf{sparsity} of streak images leads to uninformative gradients. We illustrate this problem in the top row of Fig.~\ref{fig:blurring}. The image on the left is an overlap of four streak images. We extract the red row and plot the intensity against pixel coordinates on the right plot. Following our notation from Sec.~\ref{sec:problem_formulation}, $\mathbf{D}^{(m)}$ denotes one of the observed streak images, and $\mathbf{S}^{(m)}(\mathbf{o}^+)$, $\mathbf{S}^{(m)}(\mathbf{o}')$, and $\mathbf{S}^{(m)}(\mathbf{o}'')$ represent three other generated streak images from different solution candidates. 

Let $\mathbf{o}'$ be the current estimate in this scenario, the goal of gradient descent algorithms is to move towards an optimal orbit that yields the least deviation with $\mathbf{D}^{(m)}$. As alluded to above, the travelling direction (gradient) is determined based on the neighbouring loss changes, or more formally, the first-order derivative of the loss function. As observed in the top row of Fig.~\ref{fig:blurring}, both neighbours yield the same deviation (loss) from $\mathbf{D}^{(m)}$, illustrating our point that the local gradient of $\mathbf{o}'$ provides (almost) no information to improve the fit between two signals.

An effective remedy is to enlarge the streak region with a blurring kernel. It increases the likelihood of overlapping the streaks (or enlarging the overlapping region), which in turn boosting the information provided by the gradient. Formally, given a 2D kernel with $k \times k$ dimensions, the blurring function, $\mathbf{X}_k = B(\mathbf{X}, k)$, can be expressed as 

\begin{eqnarray}
    {{x}_{k}}_{xy} = \sum^{\lfloor \frac{k}{2} \rfloor}_{w=-\lfloor \frac{k}{2} \rfloor}\sum^{\lfloor \frac{k}{2} \rfloor}_{h=-\lfloor \frac{k}{2} \rfloor} \frac{{x}_{x+w,\: y+h}}{k^2} \: ,
\end{eqnarray}

\noindent where ${{x}_{k}}_{xy}$ is the pixel-wise blurred intensity given kernel size of $k$. The bottom row of Fig. \ref{fig:blurring} visualizes the results of applying a blurring kernel to both the observed and the generated streak images. As seen in the 1D example (right plot), $\mathbf{S}^{(m)}_k(\mathbf{o}^+)$ has a smaller deviation from the blurred data $\mathbf{D}^{(m)}_k$. Naturally, the \textit{descending} gradient points to $\mathbf{o}^+$ - the direction to reduce the deviation.

Blurring kernels of different sizes serve different purposes at different fitting stages. We embed the blurring operation in a coarse-to-fine fitting regime as detailed in the next section.

\begin{figure*}[!htb]
    \centering  
    \begin{subfigure}[t]{0.9\textwidth}
        {\centering  \includegraphics[width=\textwidth, keepaspectratio]{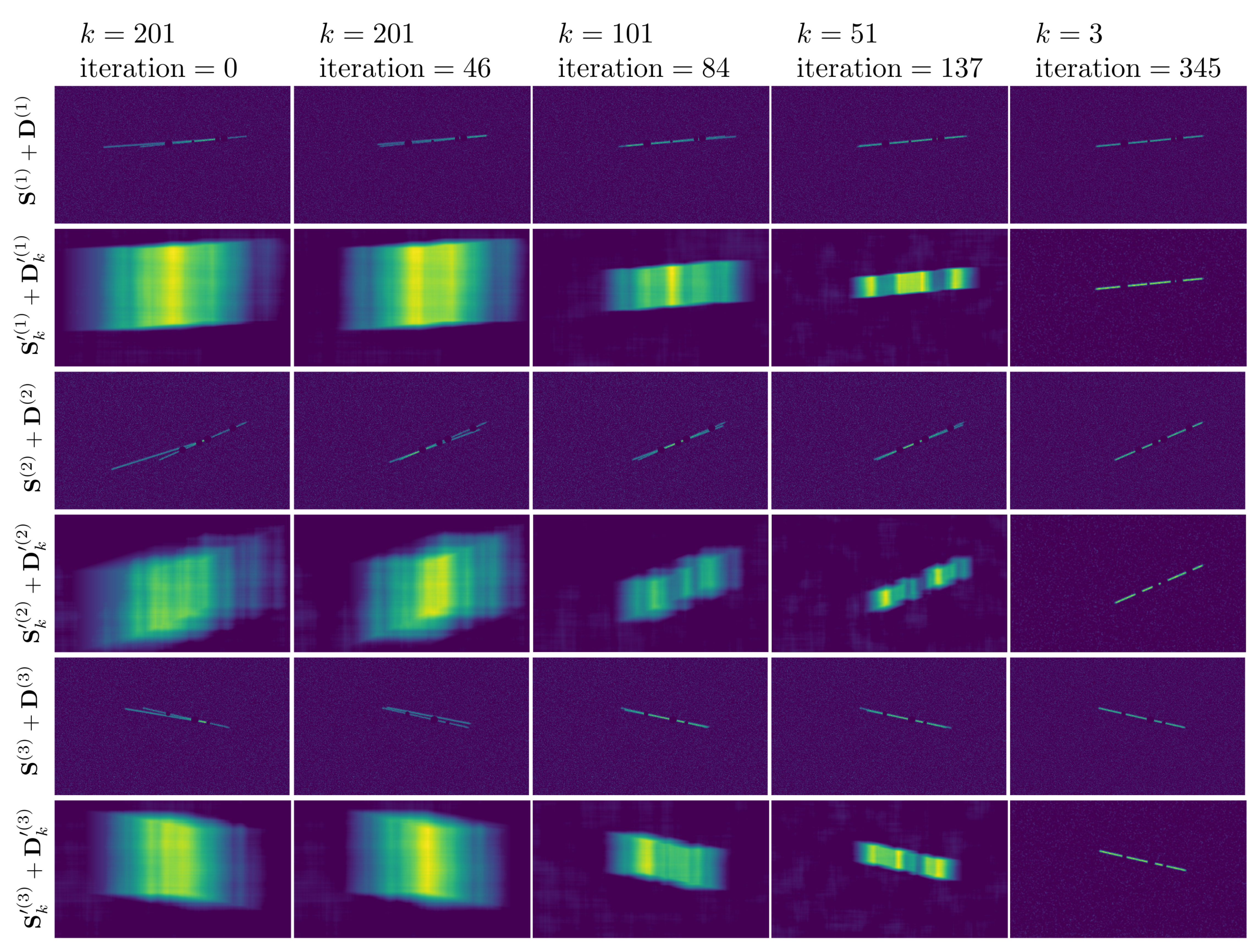}}\caption{}\label{fig:ctf_a}
    \end{subfigure}\\
    \begin{subfigure}[t]{0.3\textwidth}
        {\includegraphics[height=4cm]{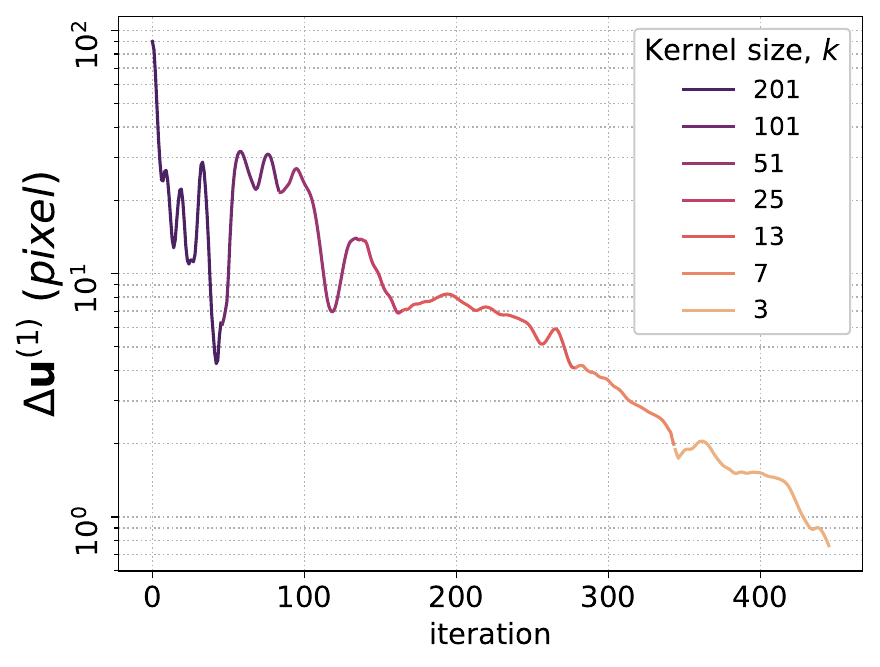}}
    \end{subfigure}
    \begin{subfigure}[t]{0.3\textwidth}
        {\includegraphics[height=4cm]{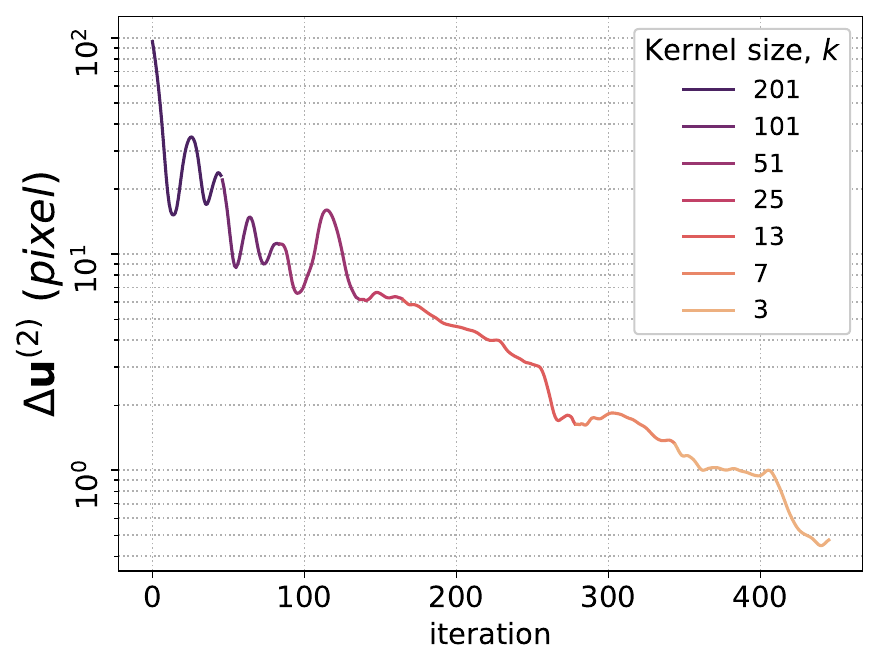}}\caption{}\label{fig:ctf_b}
    \end{subfigure}
    \begin{subfigure}[t]{0.3\textwidth}
        {\includegraphics[height=4cm]{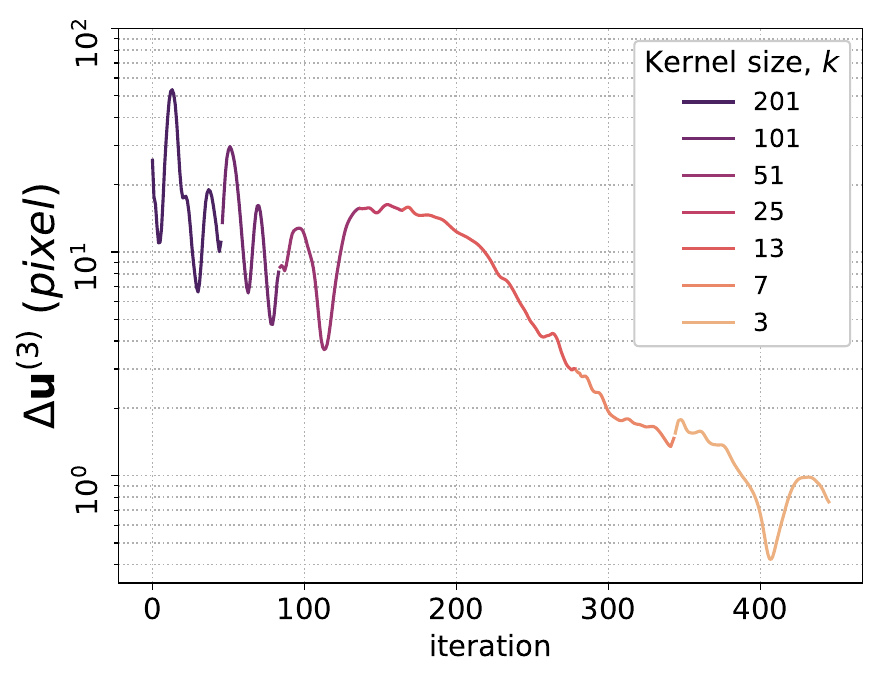}}
    \end{subfigure}\\
     \caption{The progressive improvements throughout D-IOD's coarse-to-fine fitting scheme. \textbf{(a)}: The odd rows in the figure shows the summation of the observed ($\mathbf{D}^{(m)}$) and generated streak images ($\mathbf{S}^{(m)}$). The even rows are the blurred versions of the odd rows, denoted as $\mathbf{D}'^{(m)}_k$ + $\mathbf{S}'^{(m)}_k$. The $'$ superscript denotes other processing steps detailed in Sec.~\ref{subsec:data_pre}. These superimposed images show the deviations between both sets of images. \textbf{(b)}: Endpoints' errors of the first, second, and third image, are denoted as $\Delta \mathbf{u}^{(1)}$, $\Delta \mathbf{u}^{(2)}$, and $\Delta \mathbf{u}^{(3)}$, respectively.}\label{fig:coarse_to_fine_u}
\end{figure*}

\begin{figure}[!htb]
    \centering  
    \begin{subfigure}[t]{0.23\textwidth}
        {\includegraphics[width=\textwidth]{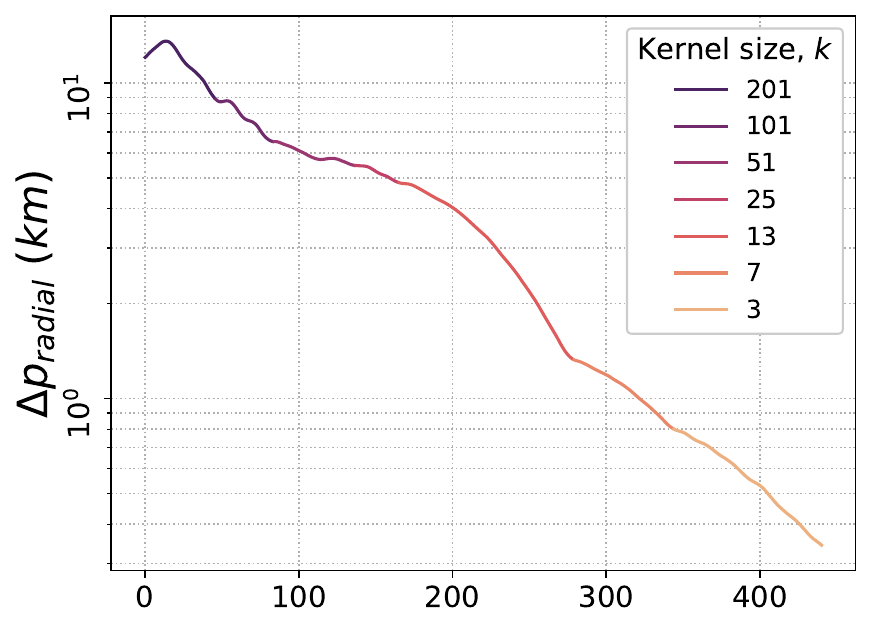}}
    \end{subfigure}
    \begin{subfigure}[t]{0.23\textwidth}
        {\includegraphics[width=\textwidth]{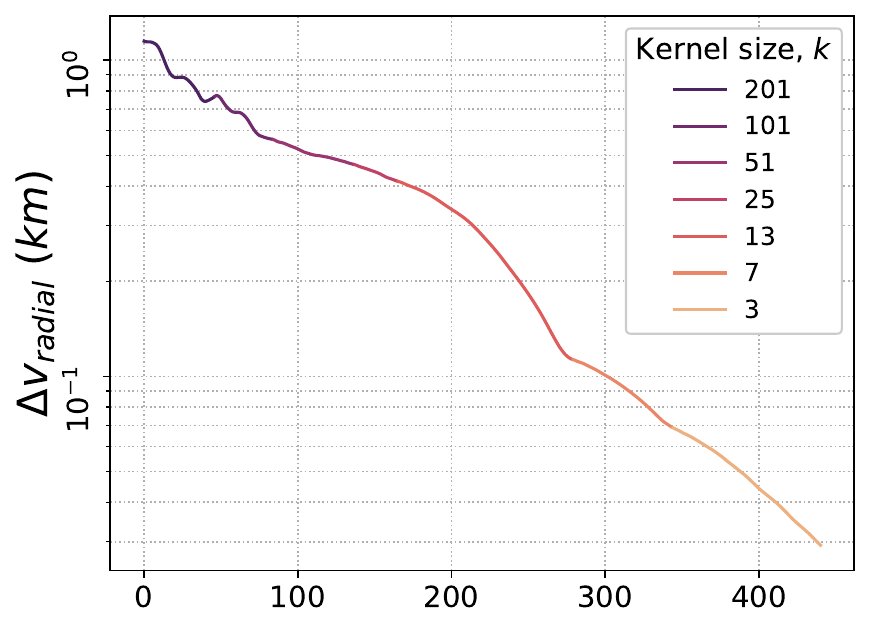}}
    \end{subfigure}\\
    \begin{subfigure}[t]{0.23\textwidth}
        {\includegraphics[width=\textwidth]{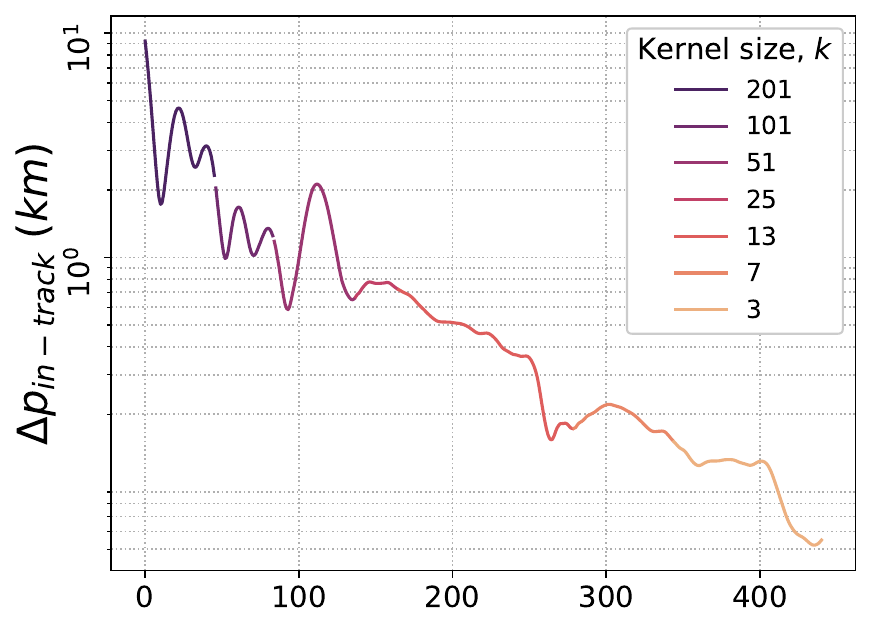}}
    \end{subfigure}
    \begin{subfigure}[t]{0.23\textwidth}
        {\includegraphics[width=\textwidth]{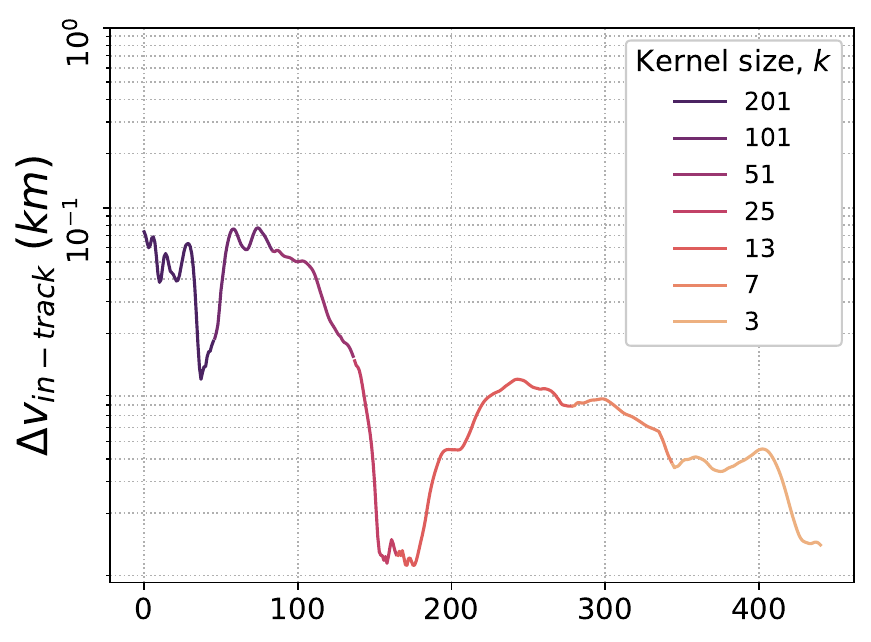}}
    \end{subfigure}\\
    \begin{subfigure}[t]{0.23\textwidth}
        {\includegraphics[width=\textwidth]{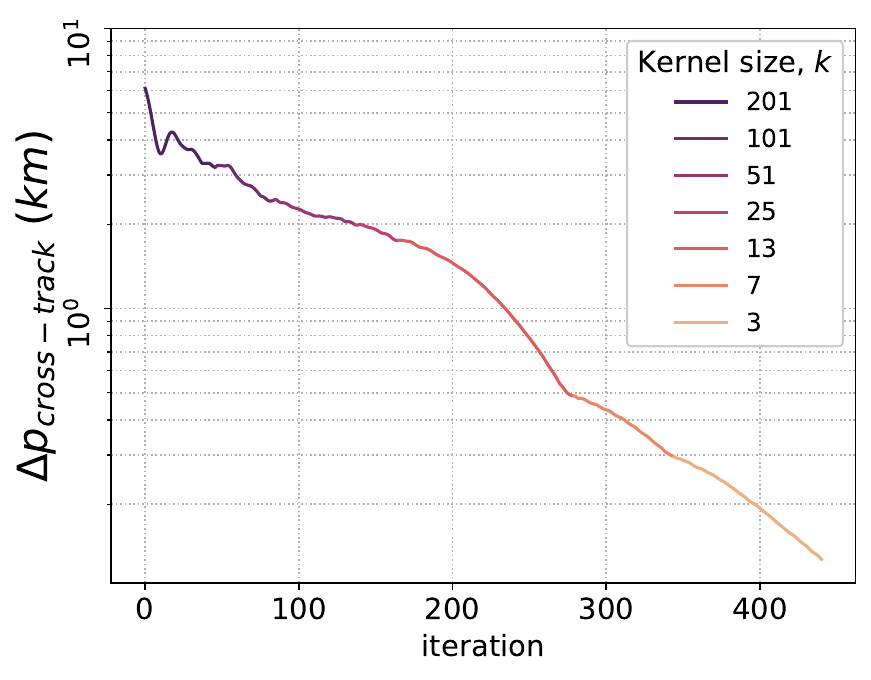}}\caption{}\label{fig:ctf_c}
    \end{subfigure}
    \begin{subfigure}[t]{0.23\textwidth}
        {\includegraphics[width=\textwidth]{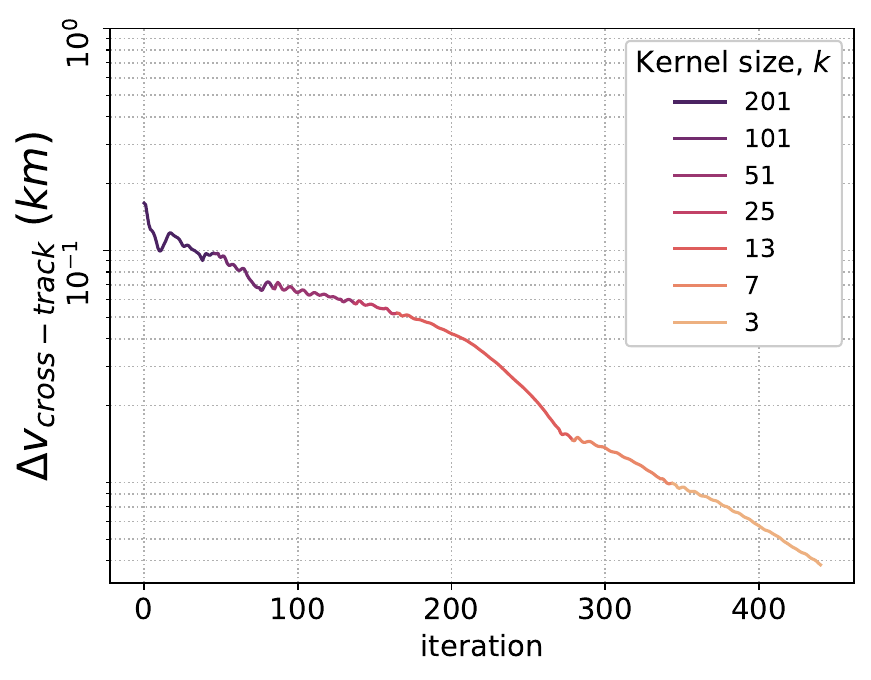}}\caption{}\label{fig:ctf_d}
    \end{subfigure}\\
     \caption{Continuation of Fig.~\ref{fig:coarse_to_fine_u}. \textbf{(a)} and \textbf{(b)}: Position and velocity vectors errors in the radial (left), in-track (middle), and cross-track (right) components.}\label{fig:coarse_to_fine_pv}
\end{figure}

\subsubsection{Coarse-to-fine fitting}\label{subsec:coarse-to-fine}
The granularity refers to the resolution of the streak image, \ie, a larger blurring kernel produces a coarser image. In the early stage, where the overlapping region of the streaks is small, if not non-existent, a larger kernel size ($k$) is required. It is effective in fitting the general location and orientation of the streak. We illustrate this with an example in Fig.~\ref{fig:ctf_a}, which shows the iterative improvement in fitting three streak images associated to an RSO. 

The first two columns of Fig.~\ref{fig:ctf_a} shows a `before-and-after' example of fitting with a large kernel size ($k$ = 201), where the improvement of the fit is visible. Quantitatively, the endpoints' error, denoted as $\Delta \mathbf{u}$, is a metric that precisely measure the fitness of the generated streak images based on the current orbital estimation. It measures the average distance between the endpoints of the streaks projected from the ground-truth\footnote{Available in our simulated settings.} and estimated orbits. As seen in Fig.~\ref{fig:ctf_b}, at iteration 0, the endpoints' errors are 90 pixels, 96 pixels, and 25 pixels for the first to third streak images. When the loss converges\footnote{See Sec.\ref{subsec:gd} for the convergence criteria.} at iteration 46, the errors improve to 6 pixels, 22 pixels, and 13 pixels. The loss converged because the deviations between the (blurred) generated and observed images became insignificant. However, the ill-fitness of the streaks are still apparent in the non-blurred images (finest resolution), as seen in the first, third, and fifth rows of Fig.~\ref{fig:ctf_a}. Recall that the optimization is performed based on the blurred images, we show the non-blurred images to show the actual fitness of the streaks.

As such, we gradually decrease the kernel size ($k$) upon each convergence to progressively refine the details of the fit until the pre-defined smallest kernel size. We found that the blurring operation also help with smoothing out the high frequency Gaussian noise, hence we stop at $k_{\rm{min}} > 1$ to retain its noise suppression ability (more details in Sec.~\ref{subsec:data_pre}). The effectiveness of our coarse-to-fine fitting strategy can be seen in the consistent decrement of the endpoints and state vector errors in Fig.~\ref{fig:ctf_b}, \ref{fig:ctf_c} and \ref{fig:ctf_d}.

The implementation is summarized in Alg.~\ref{alg:ctf}. For each kernel size $k$, we first pre-process the data, which includes applying a $k \times k$ blurring kernel. Then, the orbital parameters (initial state vector) are updated iteratively with the gradient descent method (see Sec.~\ref{subsec:gd}) until convergence (line \ref{alg:ctf_line13} to \ref{alg:ctf_line29} of Alg.~\ref{alg:ctf}). Upon convergence, we restart the whole process to refine the orbital parameters with a smaller (halved) kernel size until $k_{\rm{min}}$.

\begin{figure}[!htb]
    \centering
    \begin{subfigure}[t]{0.3\textwidth}
        {\includegraphics[height=3cm]{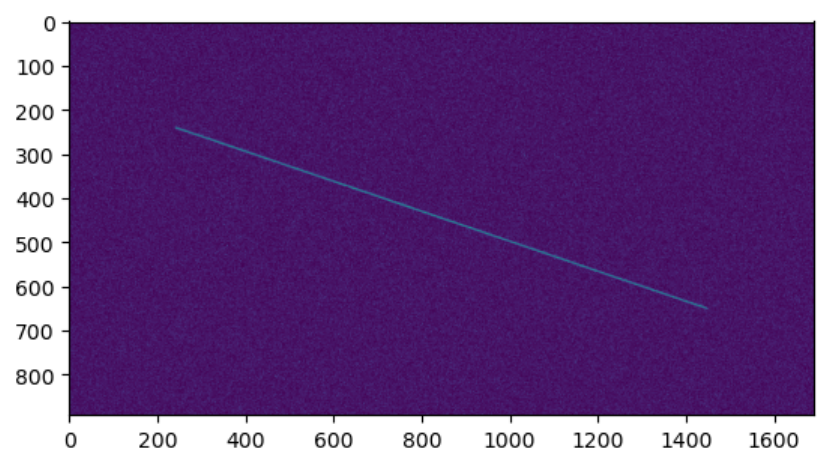}}
    \end{subfigure}
    \begin{subfigure}[t]{0.16\textwidth}
        {\includegraphics[height=3cm]{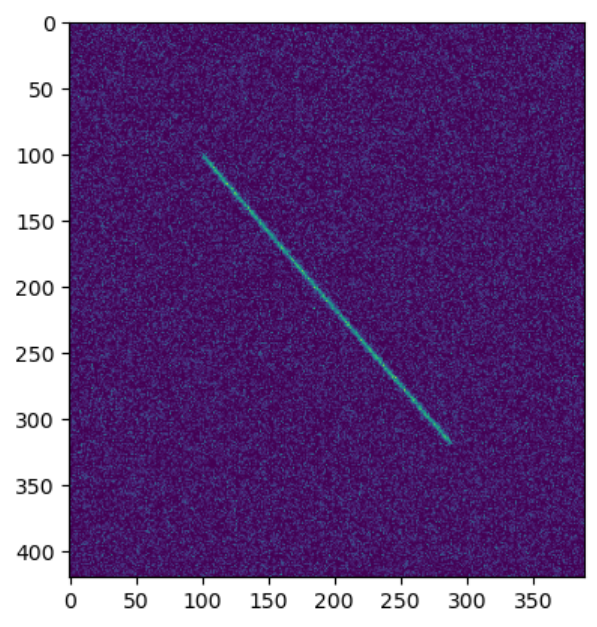}}
    \end{subfigure}
    \begin{subfigure}[t]{0.23\textwidth}
        {\includegraphics[height=3.1cm]{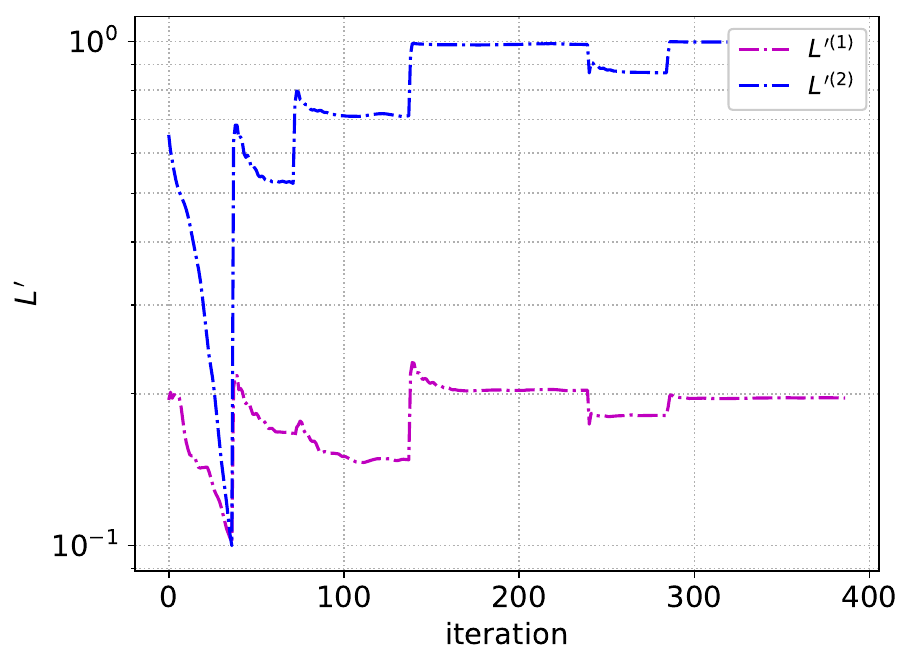}}
    \end{subfigure}
    \begin{subfigure}[t]{0.23\textwidth}
        {\includegraphics[height=3.1cm]{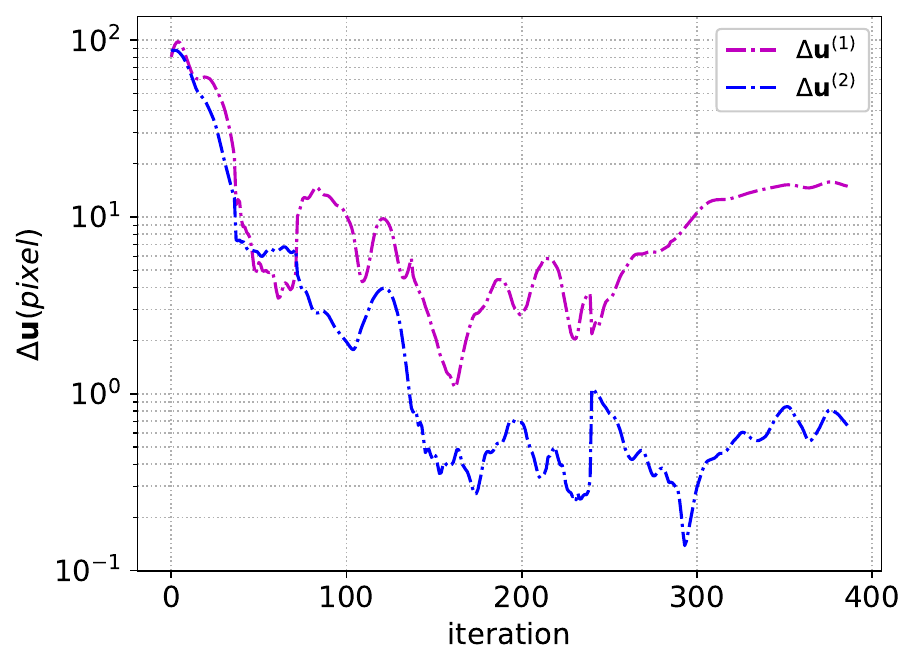}}
    \end{subfigure}\\
    \begin{subfigure}[t]{0.23\textwidth}
        {\includegraphics[height=3.1cm]{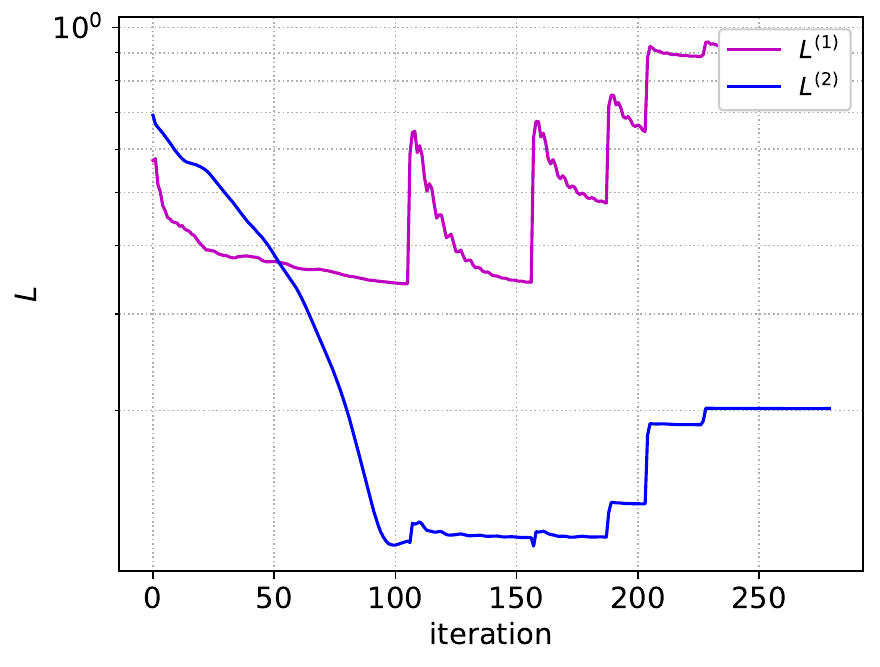}}
    \end{subfigure}
    \begin{subfigure}[t]{0.23\textwidth}
        {\includegraphics[height=3.1cm]{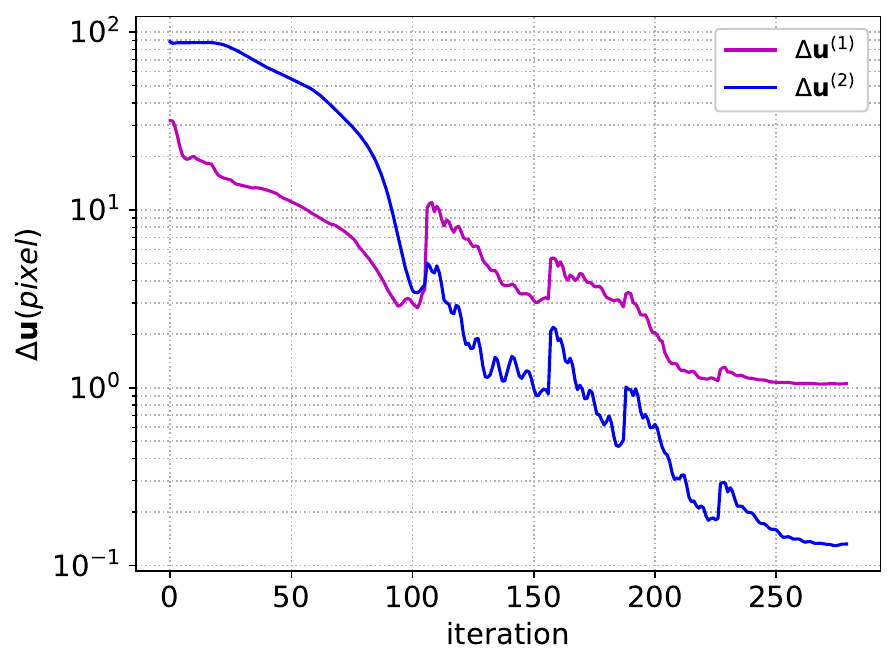}}
    \end{subfigure}
    \caption{Comparison of solving \eqref{eqn:img_fitting} with and without weighting coefficients. \textbf{Top row}: Images with huge SIR differences. Image 1 on the left has an SIR of $1.32\times10^{-6}$, while image 2 on the right has an SIR of $1.22\times10^{-5}$. \textbf{Middle row}: Optimizing \textbf{without} the weighting coefficients. \textbf{Bottom row}: Optimizing \textbf{with} the weighting coefficients. The $(1)$ and $(2)$ superscripts distinguish the loss term ${L}$ and endpoints' error $\Delta \mathbf{u}$ of image 1 and 2. Meanwhile, the loss terms in both cases are differentiated with the $'$ superscript. We highlight that the spikes in the loss terms are caused by the restart of optimization routine with a smaller kernel size. Also, the loss is generally higher for smaller kernel size because the background noise are more significant.}
    \label{fig:weights}
\end{figure}

\begin{algorithm}[!htb]
	\begin{algorithmic}[1]
		\Require Data: $\{{\mathbf{D}}^{(m)}, \mathbf{C}^{(m)}\}^M_{m=1}$
		\Require Initial estimate: $\mathbf{o}'$ 
		\Require For pre-processing: $\eta$
		\Require For coarse-to-fine: $k_{\rm{max}}$, $k_{\rm{min}}$
		\Require For gradient descent: $h, \alpha, c, \gamma, \beta_1, \beta_2$ 
		\State $\mathbf{o}^{0} \leftarrow \mathbf{o}'$\label{alg:ctf_line1}
		\State $i \leftarrow 0$\label{alg:ctf_line2} 
		    \For{$k = k_{\rm{max}}, ... , k_{\rm{min}}$}\label{alg:ctf_line3}
		        \For{$m = 1, ..., M$} \label{alg:ctf_line4} 
		            \State $\mathbf{Z}^{(m)} \leftarrow obtain\_zero\_mask(\mathbf{D}^{(m)})$\label{alg:ctf_line5}
	                \State ${\mathbf{D}}^{(m)}_{k} \leftarrow blur({\mathbf{D}}^{(m)}, k)$ \label{alg:ctf_line6}
	                \State $\beta_{\mathbf{D}_{k}} \leftarrow compute\_background\_noise({\mathbf{D}}^{(m)}_{k})$\label{alg:ctf_line7}
	                \State ${\mathbf{D}}'^{(m)}_{k} \leftarrow noise\_suppression({\mathbf{D}}^{(m)}_{k}, \beta_{\mathbf{D}_{k}})$ \label{alg:ctf_line8}
	                \State $\alpha^{(m)}_{\mathbf{D}'_{k}} \leftarrow compute\_scaling\_factor({\mathbf{D}}'^{(m)}_{k}, \eta)$\label{alg:ctf_line9}
                \EndFor \label{alg:ctf_line10}
                \State $\{\lambda^{(m)} \}^M_{m=1} \leftarrow compute\_weights(\{ {\mathbf{D}}'^{(m)}_{k}, \alpha_D^{(m)}\}^M_{m=1}, )$\label{alg:ctf_line11}
                \State ${L}^{(m)}_{\rm{diff}} \leftarrow \{\},\ {L}^{(m)}_{\rm{prev}} \leftarrow 0, \ \forall m$\label{alg:ctf_line12}
                \While{($\sum^M_{m=1} \lfloor {{L}^{(m)}_{\rm{diff_{\rm{MA}}}}} > \gamma {\overline{{L}}^{(m)}_{\rm{diff_{\rm{MA}}}}} \rfloor) > 0$}\label{alg:ctf_line13}
                        \State ${L} \leftarrow 0$\label{alg:ctf_line14}
                        \For{$m = 1, ..., M$} \label{alg1:ctf_line15} 
                	       \State $\mathbf{S}^{(m)} \leftarrow M(\mathbf{o}^{i}, \mathbf{C}^{(m)})$ \label{alg:ctf_line16} 
                	       \State $\mathbf{S}^{(m)} \leftarrow \mathbf{Z}^{(m)} \odot \mathbf{S}^{(m)}$ \label{alg:ctf_line17}
            		        \State ${\mathbf{S}}^{(m)}_{k} \leftarrow blur({\mathbf{S}}^{(m)}, k)$ \label{alg:ctf_line18}
            		        \State ${\mathbf{S}'}^{(m)}_{k} \leftarrow \alpha_{D'}^{(m)}{\mathbf{S}}^{(m)}_{k}$ \label{alg:ctf_line19}
            		        \State ${L} \leftarrow {L} + \lambda^{(m)} {L}^{(m)}(\mathbf{S}'^{(m)}_{k}, \mathbf{D}'^{(m)}_{k})$ \label{alg:ctf_line20}
            		        \State ${L}^{(m)}_{\rm{diff}} = {L}^{(m)}_{\rm{diff}} \cup ({L}^{(m)} - {L}^{(m)}_{\rm{prev}})$\label{alg:ctf_line21}
            	            \State ${{L}^{(m)}_{\rm{diff_{\rm{MA}}}}} \leftarrow compute\_moving\_average({L}^{(m)}_{\rm{diff}})$\label{alg:ctf_line22}
            	            \State ${\overline{{L}}^{(m)}_{\rm{diff_{\rm{MA}}}}} \leftarrow \max ({\overline{{L}}^{(m)}_{\rm{diff_{\rm{MA}}}}},  {{L}^{(m)}_{\rm{diff_{\rm{MA}}}}})$\label{alg:ctf_line23}
            	            \State ${L}^{(m)}_{\rm{prev}} \leftarrow {L}^{(m)}$ \label{alg:ctf_line24}
            		        \label{alg:ctf_line25}
        		        \EndFor \label{alg:ctf_line26}
    		        \State $\frac{\partial {L}}{\partial \mathbf{o}^{i}} \leftarrow compute\_gradient({L}, h)$  \label{alg:ctf_line27}   
    	            \State $\mathbf{o}^{i+1} = \mathbf{o}^{i} +  \mathcal{A}(\frac{\partial {L}}{\partial\mathbf{o}^{i}}; \beta_1, \beta_2, \alpha)$\label{alg:ctf_line28}
    	            \State $i = i + 1$\label{alg:ctf_line29}
    	        \EndWhile \label{alg:ctf_line30}
    	        \State $\alpha \leftarrow \alpha * c$\label{alg:ctf_line31}
	        \EndFor
        \State \Return $\mathbf{o}^{i}$ \label{alg:ctf_line32}
	\end{algorithmic}
	\caption{D-IOD: Direct Initial Orbit Determination.}
	\label{alg:ctf}
\end{algorithm}

\subsubsection{Weighting coefficients}\label{subsec:weights}
Streak images with different streak-to-image ratios (SIRs) impose a loss imbalance problem on D-IOD's objective function. Formally, the SIR of a pre-processed streak image, denoted as $\mathbf{D}'_k$, can be expressed as $\rm{SIR} \coloneqq \frac{\vert \overline{\mathbf{d}}'_k \vert}{\vert {\mathbf{D}'_k} \vert}$, where $\overline{\mathbf{d}}'_k \coloneqq \{{d'_k}_{xy} \vert {d'_k}_{xy} \geq {\alpha_{\mathbf{D}'_k}}, \forall x, y\}$. The thresholding value, $\alpha_{\mathbf{D}'_k}$, is an intensity value that differentiates if a pixel belongs to parts of the streak in $\mathbf{D}'_k$, which is detailed in Sec.~\ref{subsec:data_pre} alongside other pre-processing steps. 

In a scenario where the observed streak images vary significantly in terms of SIR, the final loss term is dominated by the image with the highest SIR. This stems from the mean operation in \eqref{eqn:obj_func} - a larger background pixel count (lower SIR) dilutes the loss contributed by the streak region. This leads to \textit{over-fitting}, where the gradient descent method fits only to the high SIR image with significantly larger loss term, disregarding the actual fit between the streaks, i.e., endpoints' error. We illustrate this problem in Fig.~\ref{fig:weights}.

The effect of the imbalance losses can be seen in the middle row of Fig. \ref{fig:weights}. Firstly, notice that the gap between both loss terms (denoted as ${L}'^{(1)}$ and ${L}'^{(2)}$) is huge from the beginning despite having similar endpoints' errors (denoted as $\Delta \mathbf{u}^{(1)}$ and $\Delta \mathbf{u}^{(2)}$). Secondly, as alluded to above, the loss terms do not reflect the actual fitness of the streaks - notice from iteration 100 onward, $\Delta \mathbf{u}^{(1)}$ is larger than $\Delta \mathbf{u}^{(2)}$ but the ranking of the loss terms is inverse, i.e., $\mathcal{L}'^{(1)} < \mathcal{L}'^{(2)}$. The much weaker $\mathcal{L}'^{(1)}$ has a negligible influence on the gradient descent algorithm to compute the successive updates, which results in deteriorating $\Delta \mathbf{u}^{(1)}$ after iteration 150.

A simple and effective remedy is to add weighting coefficients $\{\lambda^{(m)}\}^M_{m=1}$ to balance the loss terms. We compute $\lambda^{(m)}$ as a function of the SIR, which can be expressed as follows,

\begin{eqnarray}
    \lambda^{(m)} = \frac{\text{SIR}_{\rm{max}}}{\text{SIR}^{(m)}},
\end{eqnarray}
    
\noindent where $\rm{SIR}_{\rm{max}} \geq \rm{SIR}^{(m)} \ \forall m$. Each weighting term is then multiply to their respective loss terms in \eqref{eqn:img_fitting}.

The effects of having the weighting coefficients are reflected in the bottom row of Fig. \ref{fig:weights}. Notice that the gap is much closer in the beginning, and the loss terms accurately reflect the fitness as the optimization progresses. As a result, the endpoints' errors in both images continuously improved and converged to better accuracy.

\subsubsection{Zero masking}
The discontinuities, i.e., regions with `0' intensity, in the observed streak images are an undesired outcome of the star removal process. These artifacts, `zero holes' henceforth, penalize gradient descent for reaching orbital solutions that go through the holes when projected as streaks. 

To overcome this, we introduce the same `zero holes' structure in the generated streak images. We obtain a binary mask, which can be formally expressed as $\mathbf{Z}^{(m)} \in \mathbb{Z}^{X_m \times Y_m} \coloneqq \{z_{xy} \vert z_{xy} = 0 \ \text{if} \ d_{xy} \in \mathbf{D}^{(m)} == 0, \ \text{and}\ z_{xy} = 1 \ \text{otherwise}\}$. Then, the operation is a simple element-wise multiplication of $\mathbf{Z}^{(m)}$ and $\mathbf{S}^{(m)}$ (see line~\ref{alg:ctf_line16} in Alg.~\ref{alg:ctf}). Masking out these regions prevent these pixels to contribute to the final loss term, hence avoiding the unwanted penalty.

\subsection{Data pre-processing}\label{subsec:data_pre}
Here we summarize the main components in our data pre-processing procedure: blurring, background noise subtraction and streak scale determination. 

\begin{figure}[!htb]
    \centering
    \begin{subfigure}[t]{0.23\textwidth}
        {\includegraphics[height=3cm]{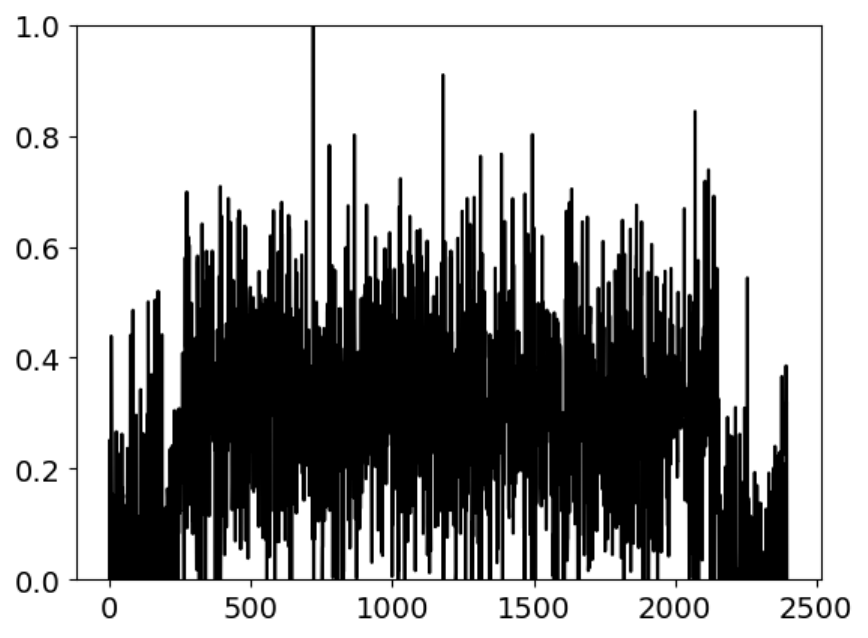}}\caption{Normalized streak}\label{fig:pre_a}
    \end{subfigure}
    \begin{subfigure}[t]{0.23\textwidth}
        {\includegraphics[height=3cm]{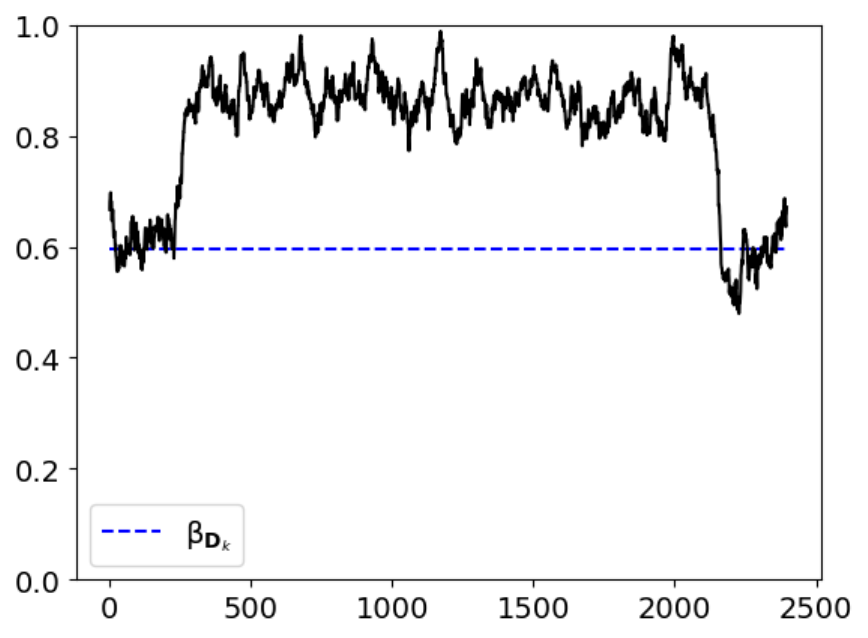}}\caption{Blurred streak}\label{fig:pre_b}
    \end{subfigure}
    \begin{subfigure}[t]{0.23\textwidth}
        {\includegraphics[height=3cm]{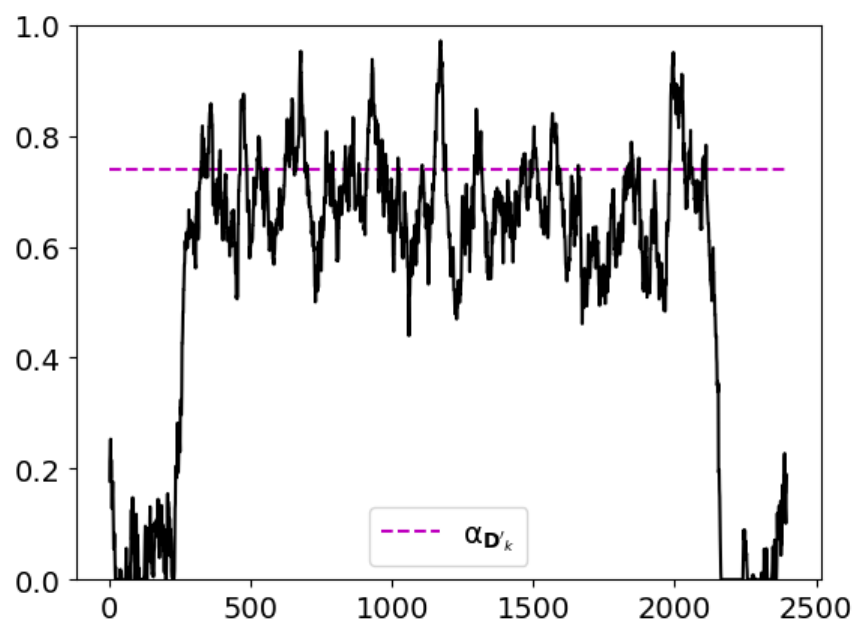}}\caption{Noise-reduced streak}\label{fig:pre_c}
    \end{subfigure}%
    \begin{subfigure}[t]{0.23\textwidth}
        {\includegraphics[height=3cm]{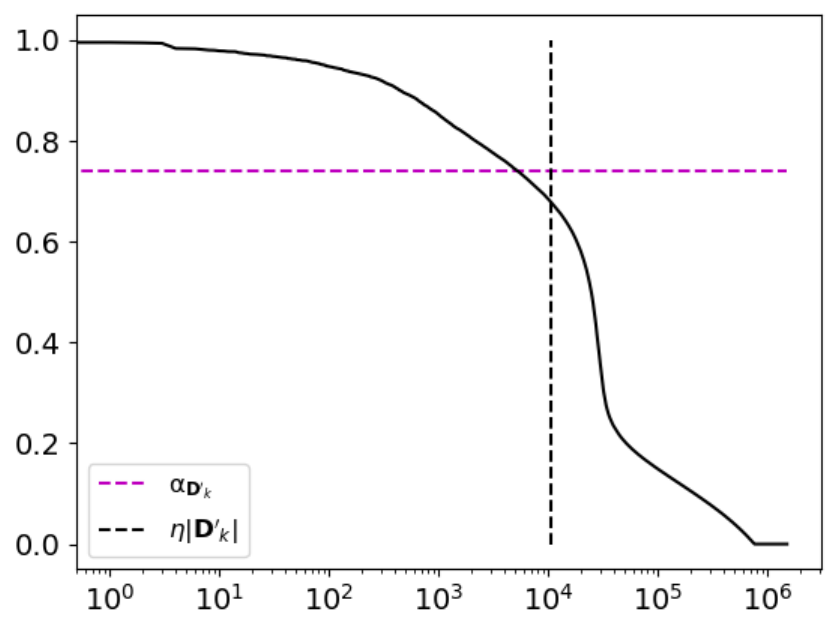}}\caption{Sorted (vectorized) data}\label{fig:pre_d}
    \end{subfigure}%
    \caption{D-IOD's data pre-processing steps; see text for details. The y-axes represent the normalized intensity values. Plots \ref{fig:pre_a}, \ref{fig:pre_b}, and \ref{fig:pre_c} shows the intensity values extracted from the streak region of image (including some nearby background pixels) at different stages of the pre-processing pipeline. The blue dashed line in \ref{fig:pre_b} indicates the background noise level obtained with a median operation over the blurred observed image $\mathbf{D}_k$. The log scale plot (x-axis) in \ref{fig:pre_d} visualizes the intensity-sorted (processed) streak image, $\mathbf{D}'_k$. The median intensity of the streak is highlighted with magenta dashed lines. The hyperparameter $\eta$ is the percentage of the observed image that is assumed to be part of the streak.}
    \label{fig:pre-processing}
\end{figure}

The blurring operation was described in Sec.~\ref{subsec:image_blurring}, and the blurred streak is shown in Fig.~\ref{fig:pre_b}. Upon blurring, the separation between the signal and background noise became distinctive. 

In order to further increase the signal-to-noise ratio, we subtract the blurred image with the background noise, $\beta_{\mathbf{D}_k}$, which is obtained with a median operation over the blurred image (${\mathbf{D}_k}$). Formally, the noise-reduced pixel value ${{d}'_k}_{xy} \in {\mathbf{D}'_k}$ can be expressed as 
\begin{equation}
    {{d}'_k}_{xy} = max({d_k}_{xy} - \beta_{\mathbf{D}_k}, 0) \ .
\end{equation}

We then scale the noise-reduced image to have a unity scale. As a result, the SNR of the pre-processed streak, as shown in Fig.~\ref{fig:pre_c}, has increased significantly, e.g. from approximately 2 to 4. Note that the SNR increment is not a constant; it depends on the noise in the image and the blurring kernel size.

In order to deal with the intensity variability in the streak region (as seen in Fig.~\ref{fig:pre_c}), we compute its median ($\alpha_{\mathbf{D}'_k}$) to scale the amplitude of our generated image $\mathbf{S}$ (Alg.~\ref{alg:ctf}, line~\ref{alg:ctf_line19}). The scaling factor ($\alpha_{\mathbf{D}'_k}$) can be easily determined with a median operation over the top-$\eta$ percentage of an intensity-sorted (vectorized) image. The sorted vector is illustrated in Fig.~\ref{fig:pre_d}, where the $\eta \vert \mathbf{D}'_k \vert$ black dashed line separates the top-$\eta$ percentage and the rest of the pixels (note the log scale x-axis). Recall that $\alpha_{\mathbf{D}'_k}$ is also used in computing SIR in Sec.~\ref{subsec:weights}. The selection of $\eta$ is discussed in Sec.~\ref{subsec:hyperparams}.

\subsection{Gradient descent}\label{subsec:gd}
The main workhorse of D-IOD, a gradient descent optimizer, is detailed in this section. The main steps are as follows.
\begin{enumerate}
    \item Generate streak images with the current estimate (Alg.~\ref{alg:ctf}, line~\ref{alg:ctf_line16} to line~\ref{alg:ctf_line19}).
    \item Loss computation (Alg.~\ref{alg:ctf}, line~\ref{alg:ctf_line20}).
    \item Gradient computation (Alg.~\ref{alg:ctf}, line~\ref{alg:ctf_line26}).
    \item Parameters update (Alg.~\ref{alg:ctf}, line~\ref{alg:ctf_line27}).
    \item Convergence detection (Alg.~\ref{alg:ctf}, line~\ref{alg:ctf_line12}).
\end{enumerate}

\noindent Step 1 has been covered in Sec.~\ref{subsec:chip_modelling} (generation) and Sec.~\ref{subsec:data_pre} (blurring and scaling operation). We elaborate only step 3 and 5 below since step 2 and 4 are straightforward operations.\\

\subsubsection{Gradient computation}
\paragraph{Numerical differentiation}
We opt for the numerical differentiation instead of the analytical differentiation since the transcendental Keplerian propagator\footnote{It is solved with numerical methods.} (Sec.~\ref{subsubsec:propagator}) has no analytical gradient. We approximate the gradient via the central finite difference method \cite[Chapter~8]{wright1999numerical}. Specifically, each partial derivative of the loss function $L$, $\frac{\partial L}{\partial o_q} \in \frac{\partial L}{\partial \mathbf{o}} \in \mathbb{R}^Q$ is expressed as

\begin{equation}\label{eqn:gradient}
    \frac{\partial L}{\partial o_q} \approx \lim_{h \to 0} \frac{L(\mathbf{o} + h\mathbf{e}_q) - L(\mathbf{o} - h\mathbf{e}_q)}{2h} \, ,
\end{equation}

\noindent where $\mathbf{e}_q$ is the $q$-$th$ column of the identity matrix $\mathbf{E} \in \mathbb{R}^{Q \times Q}$, and $h$ is the step size to perturb the current estimate. Note that the $t_{\rm{initial}}$ subscript of the initial state vector ($\mathbf{o}$) is dropped here for compactness.\\

\paragraph{ADAM optimizer} To speed up convergence, we use the ADAM optimizer \citep{kingma2014adam} as part of our gradient descent algorithm. Detailing the ADAM optimizer is out of the scope of this paper, hence we summarize only the essential components below. 

% Interested readers are encouraged to read \cite{kingma2014adam} for more details. 

In contrast to normal gradient descent, it leverages past gradient information compute a better update. We denote the ADAM optimizer as $\mathcal{A}$ in line \ref{alg:ctf_line26} of Alg.~\ref{alg:ctf}, where the hyperparameters are 1) the step size of the gradient, $\alpha$, 2) exponential decay rates for the moment estimates, $\beta_1$ and $\beta_2$. The decay rates are set to the recommended $\beta_1 = 0.9$ and $\beta_2 = 0.999$ for all the experiments in this paper. 

For D-IOD, the crucial hyperparameters in the gradient descent method are $h$ (for gradient computation) and $\alpha$ (for parameter updates). The standard practice of gradient descent methods is to reduce the step size as the solution converges. We implement that by down-scaling $\alpha$ with a \textit{cooldown} hyperparameter denoted as $c$ (see line~\ref{alg:ctf_line29} in Alg.~\ref{alg:ctf}). The selections of $\alpha$, $c$, and $h$ are detailed in Sec. \ref{subsec:hyperparams}.\\

\subsubsection{Convergence detection}
We compute the moving average (MA) of the absolute loss differences $L_{\rm{diff}}$ (see line \ref{alg:ctf_line20} in Alg.~\ref{alg:ctf}) to determine if the optimization has reached a plateau. The moving average of $L_{\rm{diff}}$ at iteration $i$ can be expressed as

\begin{equation}\label{eqn:ma}
    {L_{\rm{diff}_{MA}}} = \frac{1}{v} \sum^i_{j=i-v+1} {L_{\rm{diff}}}_j \ .
\end{equation}

\noindent D-IOD breaks out of the \textit{while} loop when the current ${L_{\rm{diff}_{MA}}}$ is smaller than $\gamma  {\overline{L}_{\rm{diff}_{MA}}}$ (Alg.~\ref{alg:ctf}, line \ref{alg:ctf_line12}), where ${\overline{L}_{\rm{diff}_{MA}}}$ is the current maximum absolute loss difference (Alg.~\ref{alg:ctf}, line \ref{alg:ctf_line22}). We set $\gamma$ to 0.3 and the number of $L_{\rm{diff}}$ to be averaged ($v$) to 10 in all of our experiments.

\section{Experiments}\label{sec:experiments}
We detail our experiments in this section. First, we provide the simulated settings and the chosen hyperparameters in Sec.~\ref{sec:sim_setting} and Sec.~\ref{subsec:hyperparams}, respectively. Additionally, the evaluation metrics used in our experiments are outlined in Sec.~\ref{subsec:metrics}.

We evaluated both modes of D-IOD - \textit{refine} and \textit{end-to-end}, under a variety of simulated scenarios, including different orbit types, variations in the quality of the initialization (Sec.~\ref{sec:init_exp}), different time intervals between images (Sec.~\ref{sec:time_exp}), and varying signal-to-noise ratios (Sec.~\ref{sec:snr_exp}). These simulated experiments allow us to measure the accuracy of D-IOD in predicting the orbital state due to the availability of ground truths. 
As a proof-of-concept, we generate only three streak images for an RSO from a consistent orbit (recall assumption A1 in Sec.~\ref{sec:intro}) in all the simulated experiments in this section. In practice, three streak images are used for the LOS-based IOD methods, with each streak image contributing one LOS vector. Additionally, we also demonstrate the robustness of D-IOD against real streak images (with no orbital information) with manual endpoints annotations in Sec.~\ref{sec:real_data_exp}.

D-IOD was implemented using Python version 3.7. All experiments were conducted on an Intel i5-8400 2.8 GHz CPU machine with 32GB of RAM and running Ubuntu 18.04 as the operating system.

\subsection{Simulated settings}\label{sec:sim_setting}
The simulated camera produces images of dimension $4930 \times 7382$ (equivalent to a camera with 36 Megapixels). It has an effective field-of-view of $12.5^\circ$ by $20^\circ$ degrees, with 10-arcsecond per pixel. Each image is captured from a randomly generated location on Earth's surface, and its pointing direction is randomly tilted (with positive elevation). Additionally, the exposure of each image is set to 5 seconds, and they are cropped with extra border regions to simulate the real streak images as shown in Fig.~\ref{fig:real_chip}.

In our experiments, we used the proposed model (described in Sec.~\ref{subsec:chip_modelling}) to generate streak images from four different orbit types in order to test the generality of D-IOD. The periapsis and eccentricity of these simulated orbits were sampled uniformly and are listed in Table~\ref{tab:test_cases}. Orbit type A represents nearly circular orbits in Low Earth Orbit (LEO), while B, C, and D simulate orbits with a range of eccentricities in Medium Earth Orbit (MEO). The rest of the Keplerian orbital elements are uniformly sampled from their full angular ranges, i.e., inclination $i \sim U(0^{^\circ}, 180^{^\circ})$, the longitude of the ascending node $\omega \sim U(0^{^\circ}, 360^{^\circ})$, the argument of periapsis $\Omega \sim U(0^{^\circ}, 360^{^\circ})$, and true anomaly $f \sim U(0^{^\circ}, 360^{^\circ})$. 

The differences in average image size (diagonal length, $d$) are tabulated in Table \ref{tab:test_cases} as well. Images of test case A are larger due to the closer range of the simulated orbits. The upper limit of the periapsis is restricted to ensure that each orbit projection forms a proper streak instead of a light blob.

\begin{figure}[!]
    \centering
    \subfloat{\includegraphics[width=0.15\textwidth]{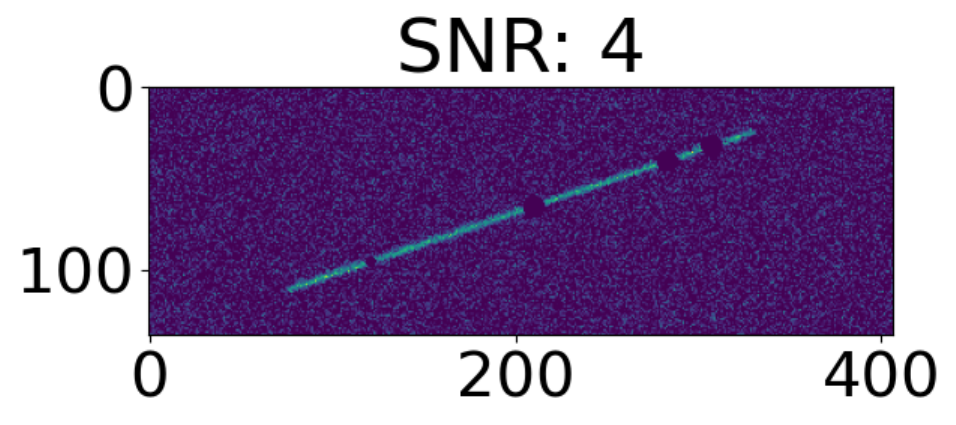}}
    \subfloat{\includegraphics[width=0.15\textwidth]{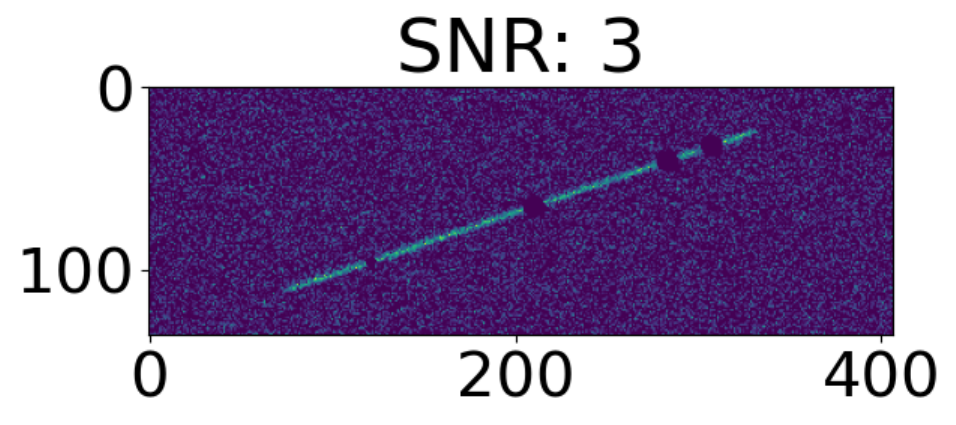}}
    \subfloat{\includegraphics[width=0.15\textwidth]{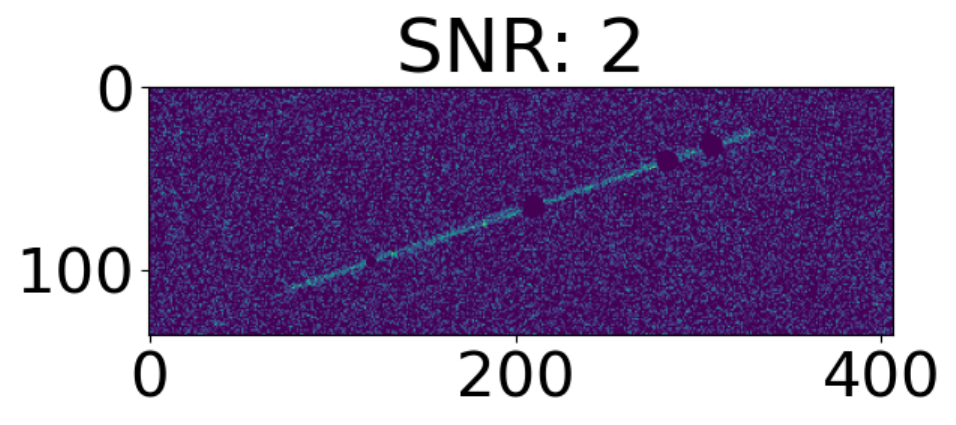}}\\
    \subfloat{\includegraphics[width=0.15\textwidth]{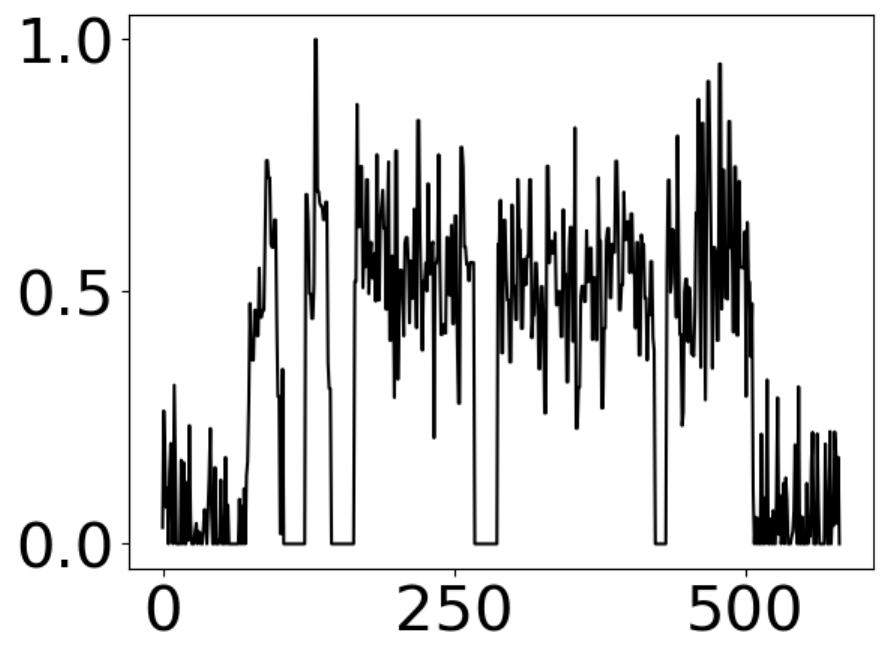}}
    \subfloat{\includegraphics[width=0.15\textwidth]{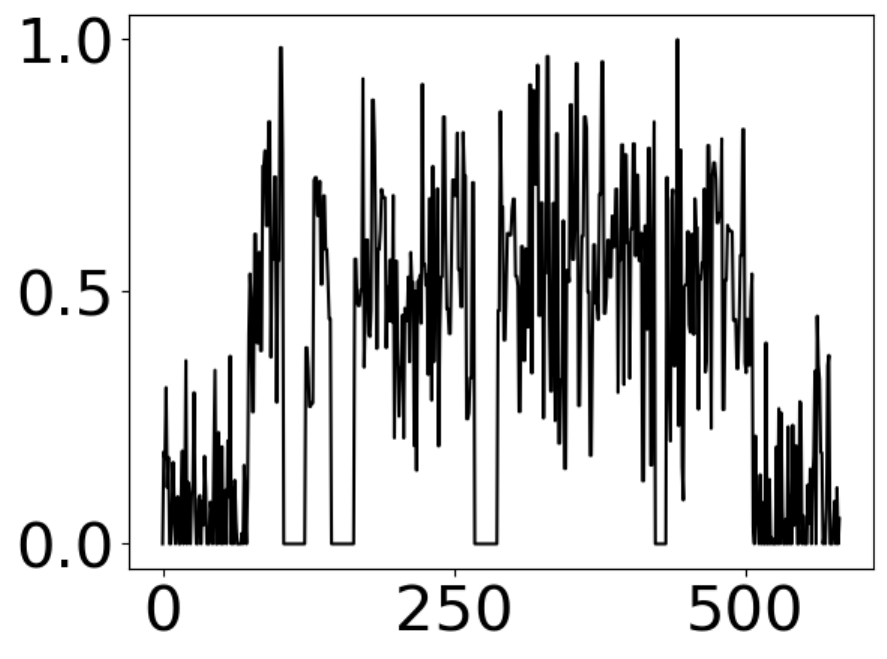}}
    \subfloat{\includegraphics[width=0.15\textwidth]{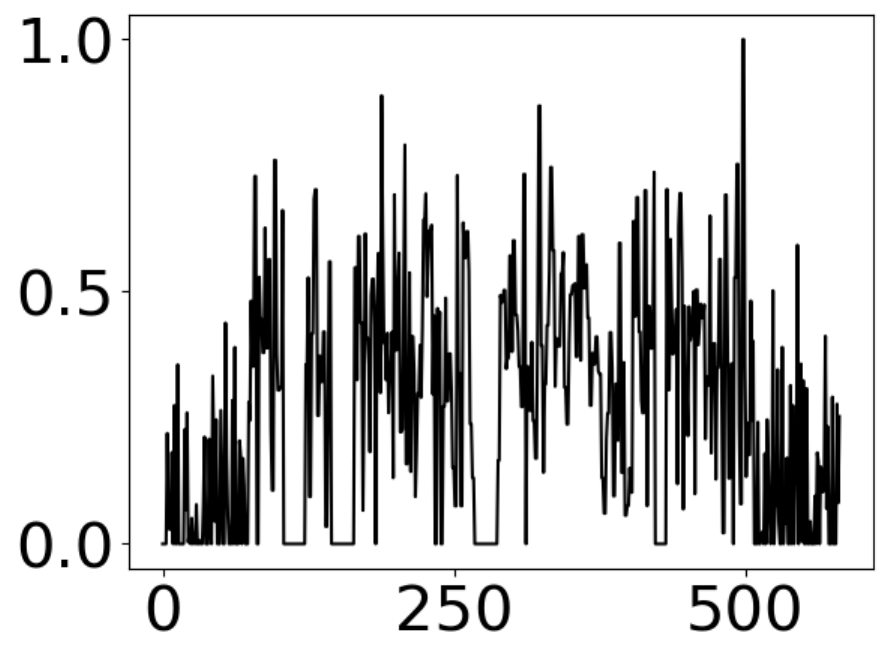}}\\
    \caption{\textbf{Top row}: Simulated streak images with different SNRs. \textbf{Bottom row}: Intensity values extracted from the streak region of the image.}
    \label{fig:diff_SNR}
\end{figure}

In order to simulate the holes that we observe in real streak images, we added four uniformly distributed holes on the streak as seen in Fig.~\ref{fig:diff_SNR}. The diameter of these holes is uniformly sampled between 5 to 20 pixels.

\begin{table}[!htb]
	\begin{center}
		\caption{Properties of the simulated orbits. The periapsis ($r_p$), eccentricity ($e$) of the simulated orbits, and diagonal length ($d$) of the streak images are presented here.}
		\label{tab:test_cases}
		\begin{tabular}{cccc}
			\toprule
			 Orbit types & $r_p (km)$ &  $e$ & $d$ (pixels)\\
			 \cmidrule(lr){1-4}
			 A & $U(6880, 8380)$ & $U(0, 0.01)$ & 538  \\
			 B & $U(8380, 9380)$ & $U(0.01, 0.2)$ & 333  \\
			 C & $U(8380, 9380)$ & $U(0.2, 0.4)$ & 343 \\
			 D & $U(8380, 9380)$ & $U(0.4, 0.6)$ & 353 \\
			\bottomrule
		\end{tabular}
	\end{center}
\end{table}

\begin{table}[!htb]
	\begin{center}
        \centering
		\caption{Hyperparameters of D-IOD for different ${\Delta t}_{\rm{max}}$ (see text).}\label{tab:hyperparams}
		\begin{tabular}{cccc}
			\toprule
          &  \multicolumn{3}{c}{ ${\Delta t}_{\rm{max}}$(s)} \\
          \cmidrule(lr){2-4}
			 Hyperparameters & 30s & 60s & 120s \\
			 \cmidrule(lr){1-4}
            $h$ & $2\times10^{-3}$ & $4\times10^{-4}$ & $2\times10^{-4}$ \\
			  $\alpha$ & 0.1 & 0.02& 0.01 \\
            $c$ & \multicolumn{3}{c}{0.5}  \\
			  $k_{\rm{max}}$ & \multicolumn{3}{c}{101} \\
			  $k_{\rm{min}}$ & \multicolumn{3}{c}{3}\\
			  $\eta$ & \multicolumn{3}{c}{0.1}\\
			\bottomrule
		\end{tabular}
	\end{center}
\end{table}

\subsection{Hyperparameters}\label{subsec:hyperparams}
In this section, we present the hyperparameters of D-IOD. A summary of the hyperparameters can be found in Table ~\ref{tab:hyperparams}. 

Firstly, both step size parameters, i.e., $\alpha$ and $h$, play crucial roles in the convergence of D-IOD. The initial step size ($\alpha$) determines the magnitude of each iterative update, and it is adjusted during the optimization process (See Alg.~\ref{alg:ctf}, line~\ref{alg:ctf_line30}). Meanwhile, $h$ is used to approximate the gradient vector in the finite difference method (see \eqref{eqn:gradient}), and it is fixed throughout the optimization process. Setting $\alpha$ and $h$ too large leads to convergence failure, while too small causes slow convergence.

In D-IOD, we found that the magnitude of ${\Delta t}_{\rm{max}}$ impacts the appropriate values for these hyperparameters. The notation ${\Delta t}_{\rm{max}}$ represents the maximum time interval between the timestamp of the initial state vector to be optimized ($t_{\rm{initial}}$) and the timestamps ($\{\tau^{m}\}^M_{m=1}$) of the observed streak images. Here we provide three sets of $\alpha$ and $h$ that cover all the experiments performed in this section. In general, a larger ${\Delta t}_{\rm{max}}$ requires smaller values of $\alpha$ and $h$. The reason behind that is the propagated state vector is sensitive to both the initial state vector and the time interval. Increments in both factors lead to larger deviations in the propagated state vectors. As such, when the time interval increases, the perturbation (affected by $\alpha$ and $h$) should be decreased to compensate for the sensitivity. Propagated state vectors with too large of a deviation might fall out of the field-of-view of the observed streak images, where the gradient is not informative due to the non-overlapping streaks (recall the uninformative gradient problem in Sec.~\ref{subsec:image_blurring}), which in turn leads to convergence failure. 

As such, it is of interest to decrease $\Delta t_{\rm{max}}$ which allows the usage of larger $\alpha$ and $h$. As described in Sec.~\ref{sec:problem_formulation}, we achieve this by setting $t_{\rm{initial}}$ to the middle timestamp between the furthest timestamps in $\{\tau^{m}\}^M_{m=1}$.

The maximum kernel size $k_{\rm{max}}$ of 101 deems to be an appropriate starting size in general. The mentioned kernel occupies approximately $20\%$ of the average image diagonal length from orbit type A, and approximately $33\%$ for the smaller images in orbit types B, C, and D. We found that this coverage has a high chance of overlapping the streaks from the observed and generated images (from the initial estimates). For images larger than average (Table~\ref{tab:test_cases}), the maximum kernel size is increased automatically before the optimization begins. The minimum kernel size $k_{\rm{min}}$ is set to 3 instead of 1 to retain the blurring effect that is part of the noise reduction as detailed in Sec.~\ref{subsec:data_pre}.

The $\eta$ hyperparameter is needed for the streak's scale determination (Sec.~\ref{subsec:data_pre}). We found that a median operation over the top-$0.1\%$ of the brightest pixels is optimal for the SIR computation (Sec.~\ref{subsec:weights}) and the scaling of our generated streak (Sec.~\ref{subsec:data_pre}).

\subsection{Metrics}\label{subsec:metrics}
In our experiments, we report two main metrics: the endpoints' error and the orbital errors. The endpoints' error, denoted as $\Delta \mathbf{u}$, is the average Euclidean distance between the predicted and simulated streak's endpoints.
% \footnote{Starting and ending pixel coordinates of the streak.}. 

The orbital error, on the other hand, is the absolute deviation between predicted and simulated Keplerian orbital elements, which are denoted as $\Delta r_p$, $\Delta e$, $\Delta i$, $\Delta \Omega$, $\Delta \omega$, and $\Delta f$. The conversions between the initial state vector (domain of D-IOD) and Keplerian orbital elements can be referred to in the textbook by \cite{vallado2001fundamentals}.

\subsection{Results}\label{sec:results}
\subsubsection{Initialization experiment}\label{sec:init_exp}
D-IOD$_{\rm{refine}}$ assumes given an initial orbit estimate as input. As described in the introduction, one practical usage of D-IOD$_{\rm{refine}}$ is to improve the potentially sub-optimal orbital estimate from the two-stage IOD method. As such, it is important to evaluate the robustness of D-IOD against initialization of different qualities.\\

\paragraph{Setup} As mentioned earlier, the orbital solution of the IOD solver in the two-stage method highly depends on the accuracy of the given set of LOS vectors, which are projected from the estimated endpoints. The higher the endpoints' error, the worse the orbital solution is. As such, we simulated the difficulty levels based on the accuracy of the estimated endpoints. We present the median of the endpoints' errors ($\Delta \mathbf{u}$) and the periapsis errors ($\Delta r_p$) of each level in Table~\ref{Tab:init_exp}. Specifically, the orbital estimate from level III is obtained by feeding the Gauss IOD solver with a set of LOS vectors that are back-projected from the streaks' endpoints with an error of approximately 70 pixels. The rest of the orbital-elements errors follow the same pattern, which can be seen in 
Appendix \ref{app:init} (Table~\ref{Tab:init_exp_1} to~\ref{Tab:init_exp_5}). The constant variables in this experiments are the time interval and SNR (see below for their respective experiments). The time interval between the two furthest images (\ie first and third) is fixed at 60s, and the SNR is set to 4.

\begin{table}[!htb]
	\begin{center}
		\caption{Results of D-IOD against different initialization and orbit types. Reported numbers are the median (Q2) of the endpoints' errors ($\Delta \mathbf{u}$) and the periapsis errors ($\Delta r_p$). Best numbers in \textbf{bold}. See Table~\ref{Tab:init_exp_1} to Table~\ref{Tab:init_exp_5} for the first (Q1) and third (Q3) quartiles' numbers. Legends: `M' for metrics, `S' for the stage of optimization, `Init.' for initial estimates, and `Conv.' for converged solutions. The unit for $\Delta \mathbf{u}$ is pixel, and \textit{km} for $\Delta r_p$.}\label{Tab:init_exp}
		\begin{tabular}{ccccccc}
			\toprule
			  \multirow{2}{*}{M}& \multirow{2}{*}{S} & \multicolumn{5}{c}{Quality of initial estimates}\\
			  \cmidrule(lr){3-7}
			     &  & I & II & III & IV & V \\
			\cmidrule(lr){1-7}
            \multicolumn{7}{c}{Orbit type A}\\
            \cmidrule(lr){1-7}
			  $\Delta \mathbf{u}$ &Init. &\textbf{0.72} &34.48 &69.12 &172.69 &172.69 \\
                &Conv. &0.73 &\textbf{0.81} &\textbf{0.78} &\textbf{0.77} &\textbf{0.77} \\
                \cmidrule{2-7}
                $\Delta r_p$ &Init. &\textbf{5.07} &129.74 &248.77 &248.77 &350.25 \\
                &Conv. &7.98 &\textbf{9} &\textbf{9.44} &\textbf{9.44} &\textbf{9.15} \\
            \cmidrule(lr){1-7}
            \multicolumn{7}{c}{Orbit type B}\\
            \cmidrule(lr){1-7}
                $\Delta \mathbf{u}$ &Init. &0.7 &24.45 &50.12 &124.99 &124.99 \\
                &Conv. &\textbf{0.66} &\textbf{1.06} &\textbf{1.25} &\textbf{1.18} &\textbf{1.18} \\
                \cmidrule(lr){2-7}
                $\Delta r_p$ &Init. &\textbf{9.72} &177.81 &302.82 &302.82 &577.02 \\
                &Conv. &16.17 &\textbf{32.21} &\textbf{27.22} &\textbf{27.22} &\textbf{31.77} \\
                \cmidrule(lr){1-7}
                \multicolumn{7}{c}{Orbit type C}\\
                \cmidrule(lr){1-7}
                $\Delta \mathbf{u}$ &Init. &0.82 &25.55 &51.81 &133.18 &133.18 \\
                &Conv. &\textbf{0.67} &\textbf{1.1} &\textbf{1.33} &\textbf{1.14} &\textbf{1.14} \\
                \cmidrule(lr){2-7}
                $\Delta r_p$ &Init. &\textbf{6.6} &78.98 &204.12 &204.12 &273.26 \\
                &Conv. &8.68 &\textbf{17.62} &\textbf{18.3} &\textbf{18.3} &\textbf{15.96} \\
                \cmidrule(lr){1-7}
                \multicolumn{7}{c}{Orbit type D}\\
                \cmidrule(lr){1-7}
                $\Delta \mathbf{u}$ &Init. &0.9 &26.97 &53.68 &134.08 &134.08 \\
                &Conv. &\textbf{0.65} &\textbf{0.96} &\textbf{1.14} &\textbf{0.92} &\textbf{0.92} \\
                \cmidrule(lr){2-7}
                $\Delta r_p$ &Init. &\textbf{4.25} &75.94 &142.55 &142.55 &218.5 \\
                &Conv. &6.16 &\textbf{14.63} &\textbf{10.01} &\textbf{10.01} &\textbf{9.18} \\
			  \bottomrule
		\end{tabular}
	\end{center}
\end{table}
\paragraph{Results} The large differences in the median endpoints' error and the median periapsis error between the initial and converged solutions can be observed in Table~\ref{Tab:init_exp}. The improvement in $\Delta \mathbf{u}$ is significant (order of magnitudes) and consistent across all levels and orbit types, particularly the challenging levels II, III, IV, and V. For level I, the converged solutions are slightly worse than the initial solutions. We associate this to the sub-optimal hyperparameters settings. The step size $\alpha$ that we used is ideal in the scenario where the initial solution is far from the optimal solution. It encourages the gradient descent algorithm to take larger steps towards the optimal solution. However, in the scenario where the initial solution is close to the optimal solution, the step size is too large, and the gradient descent algorithm "escapes" the region around the optimal solution. This is a classical \textit{overshooting} problem in optimization \citep{dixon1972choice}. We used the same set of hyperparameters for all the levels here to show that D-IOD requires no intensive tuning.

Overall, the endpoints' errors of D-IOD are consistently low, with a median range of 0.67 to 1.33 pixels in all experiments, demonstrating its robustness against initialization. The improvement in terms of periapsis error is also consistent, albeit the different error ranges across the orbit types. Such differences stem from the variations in arc spanned by the orbit types - images that capture a larger arc tend to better constrain the solution space. The order of the orbit types based on average arc-span in this experiment is A($3.3^{\circ}$) > D($3.07^{\circ}$) > C($2.88^{\circ}$) > B($2.67^{\circ}$).

\begin{figure*}
    \centering
    \subfloat{\includegraphics[height=5.1cm]{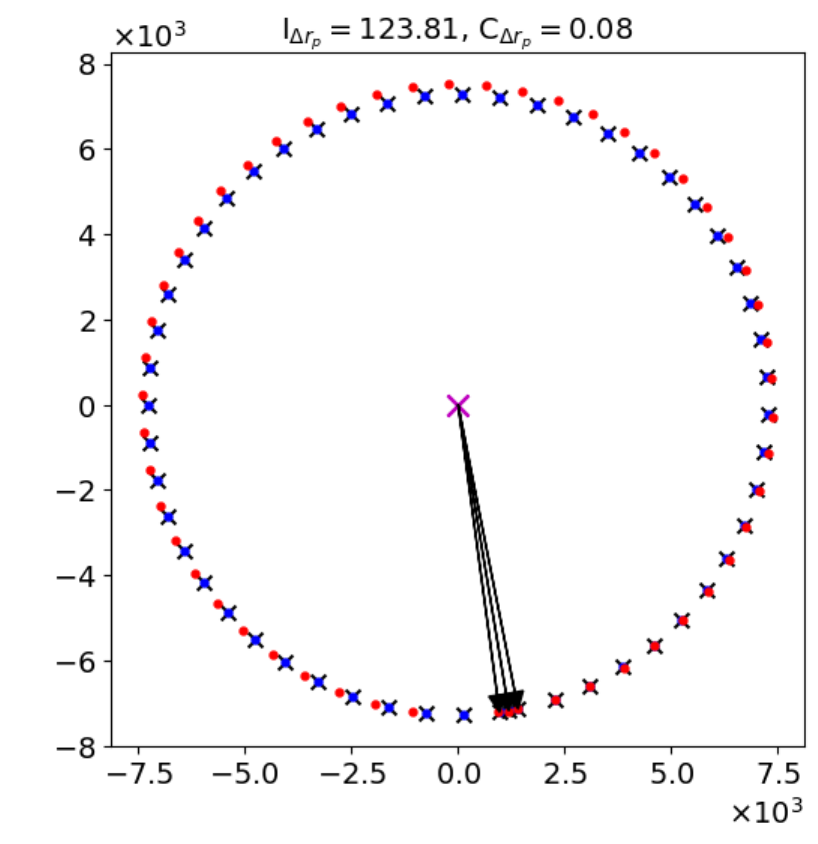}}
    \subfloat{\includegraphics[height=5.1cm]{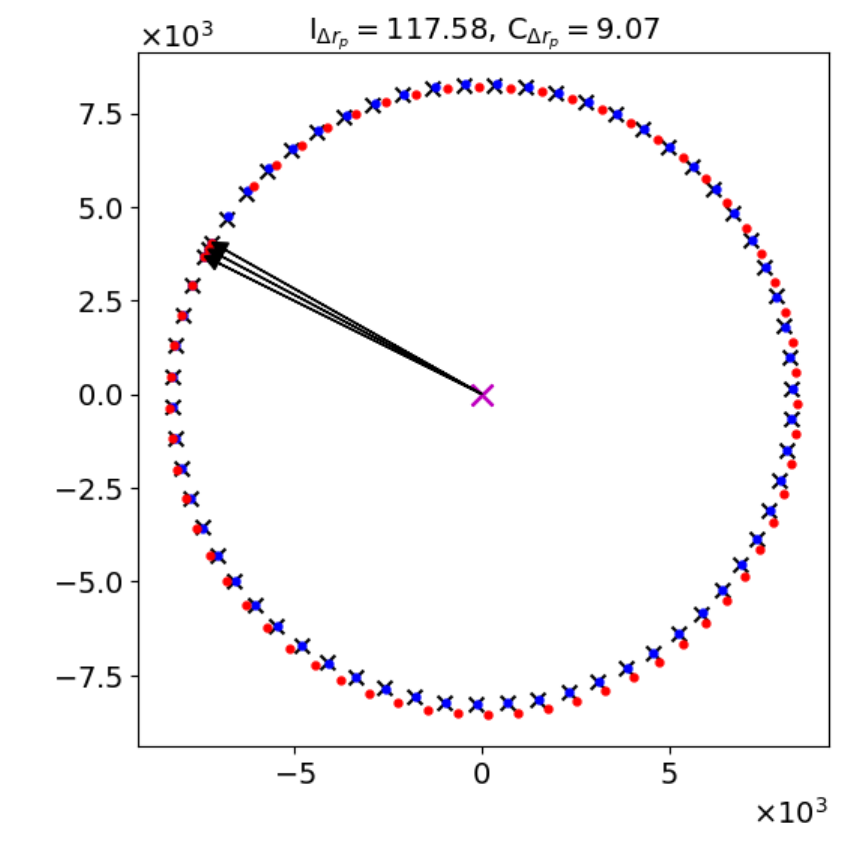}}
    \subfloat{\includegraphics[height=5.1cm]{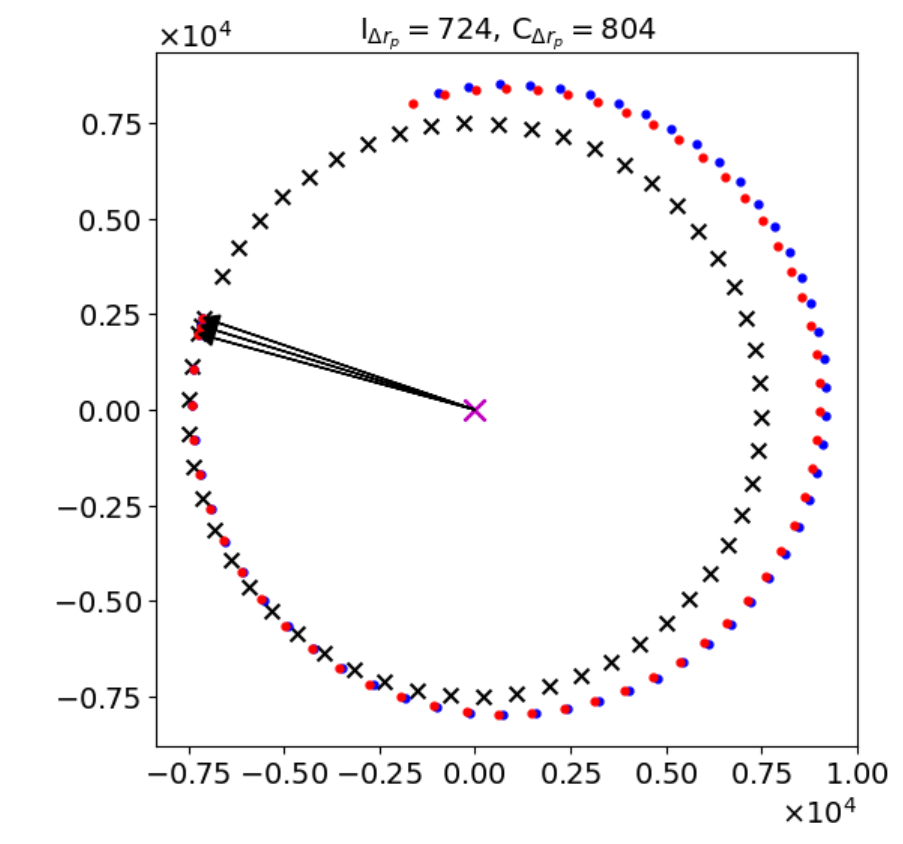}}\\
    \subfloat{\includegraphics[height=5cm]{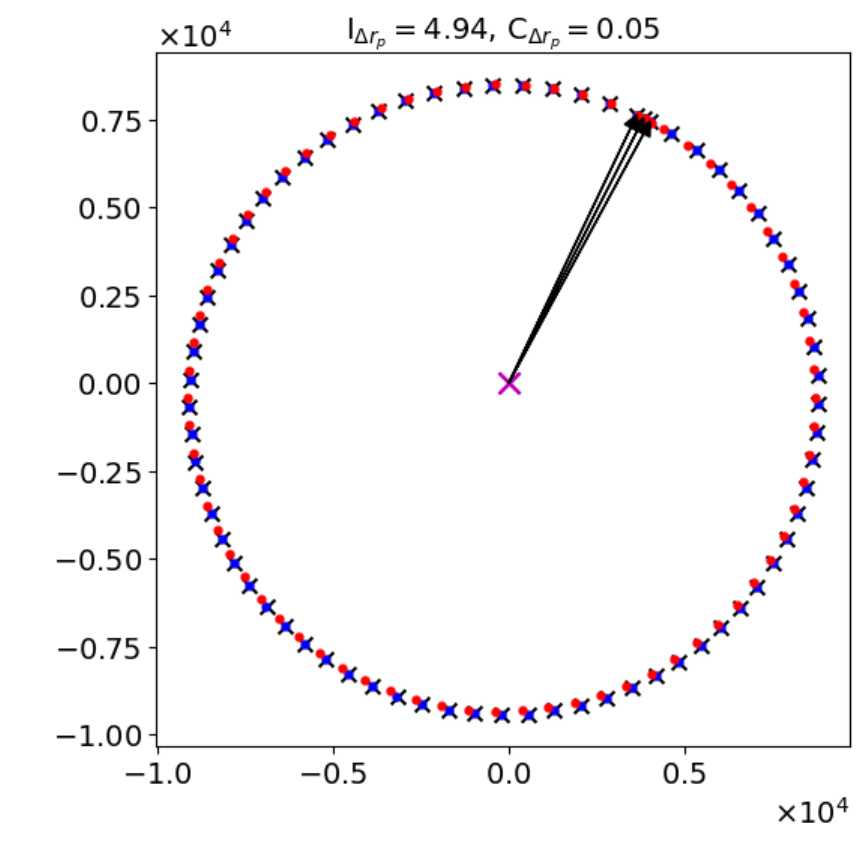}}
    \subfloat{\includegraphics[height=5cm]{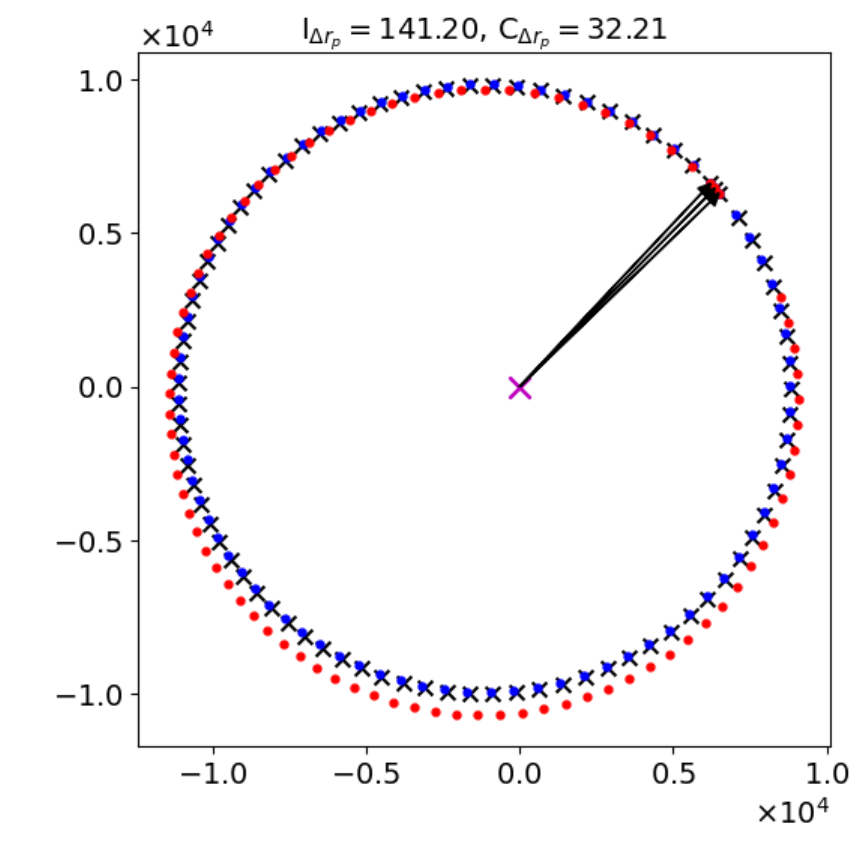}}
    \subfloat{\includegraphics[height=5cm]{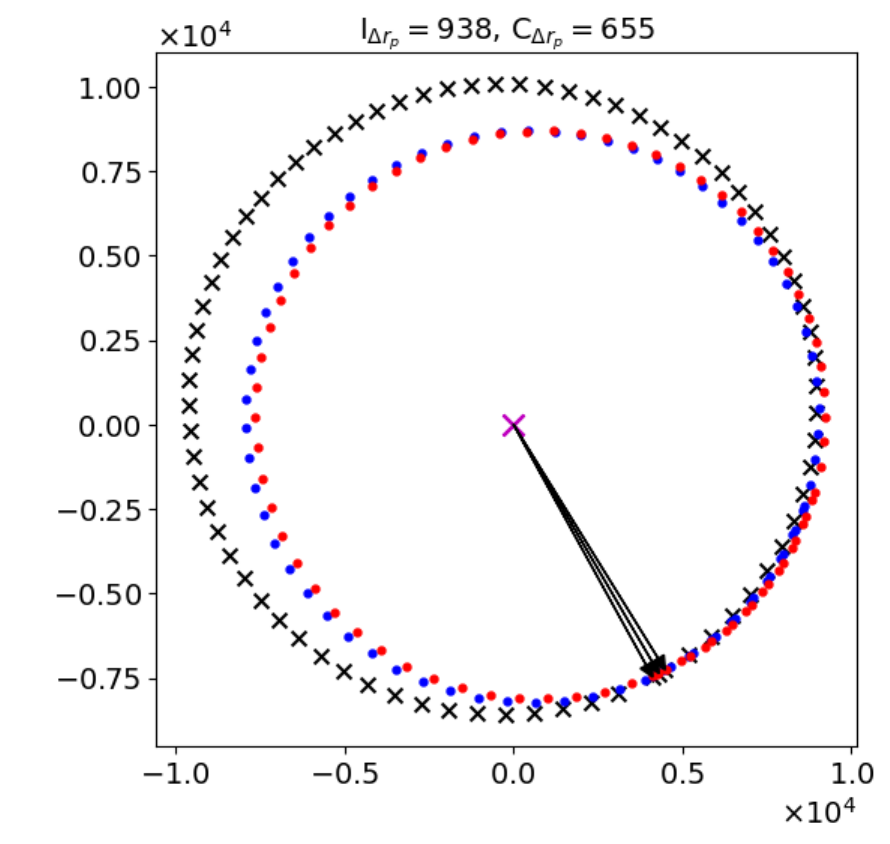}}\\
    \subfloat{\includegraphics[height=5cm]{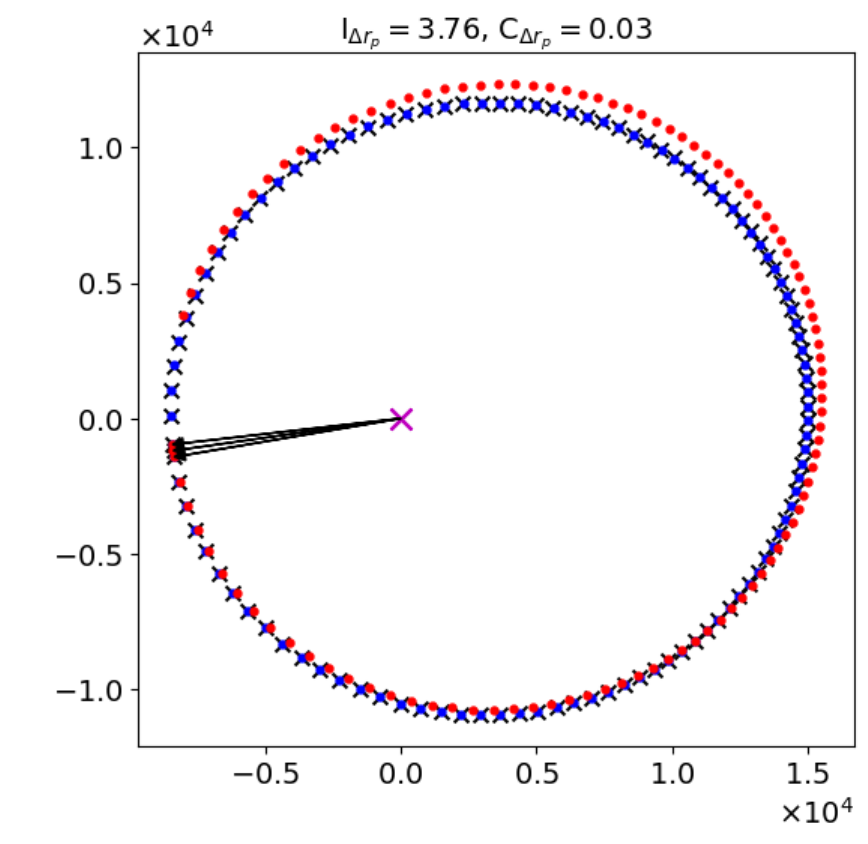}}
    \subfloat{\includegraphics[height=5cm]{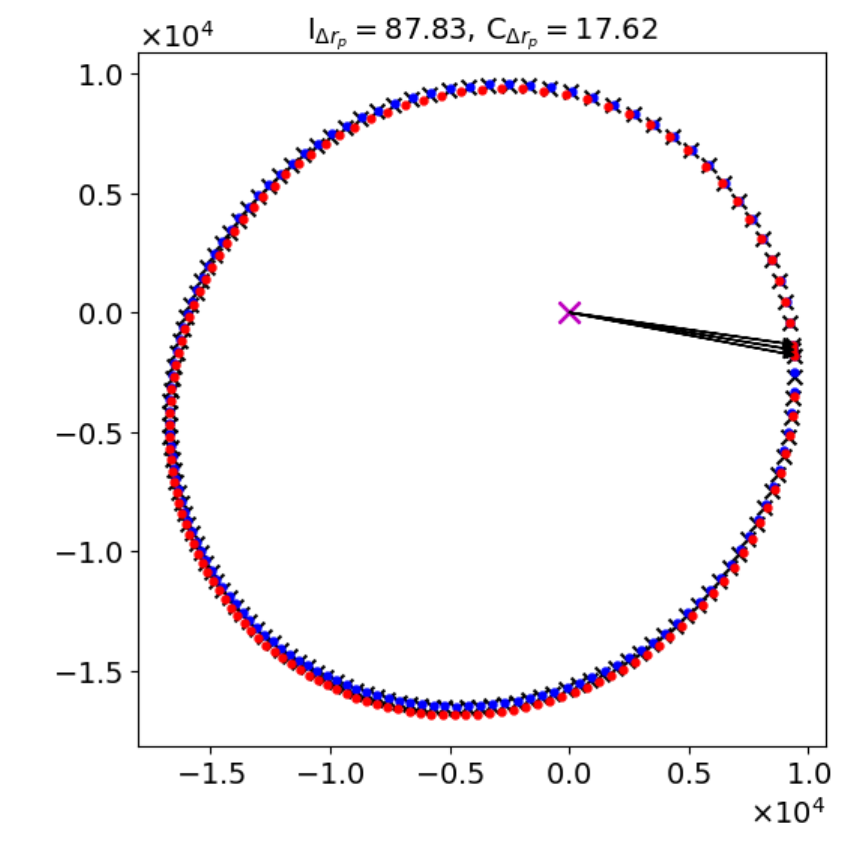}}
    \subfloat{\includegraphics[height=5cm]{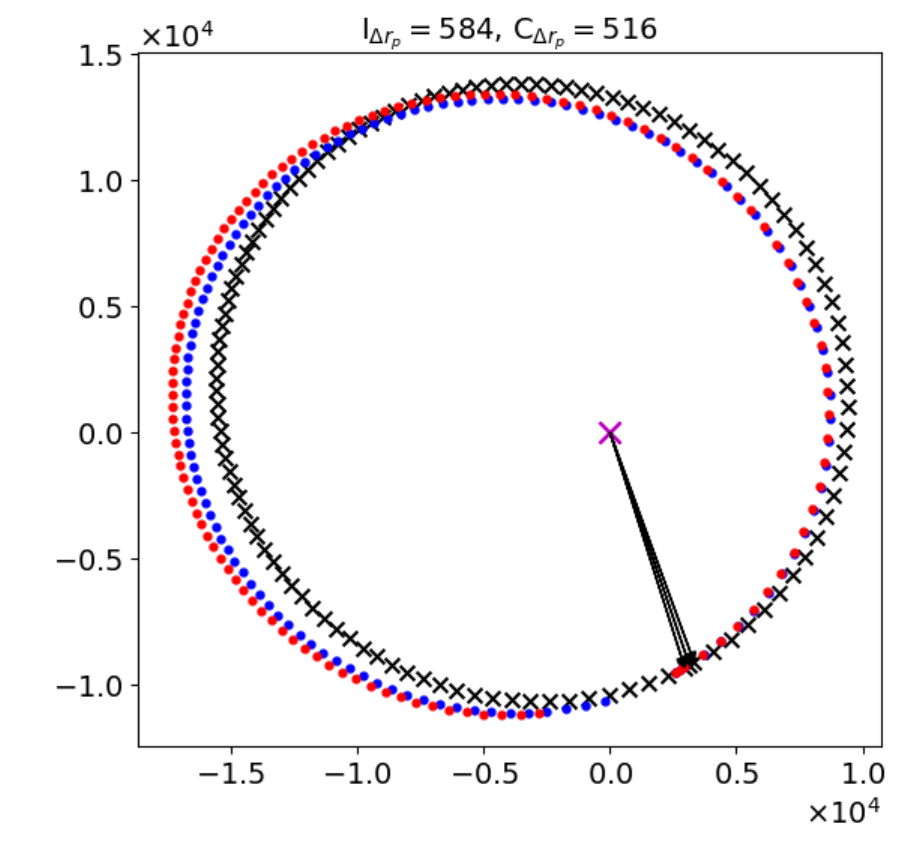}}\\
    \subfloat{\includegraphics[height=5.3cm]{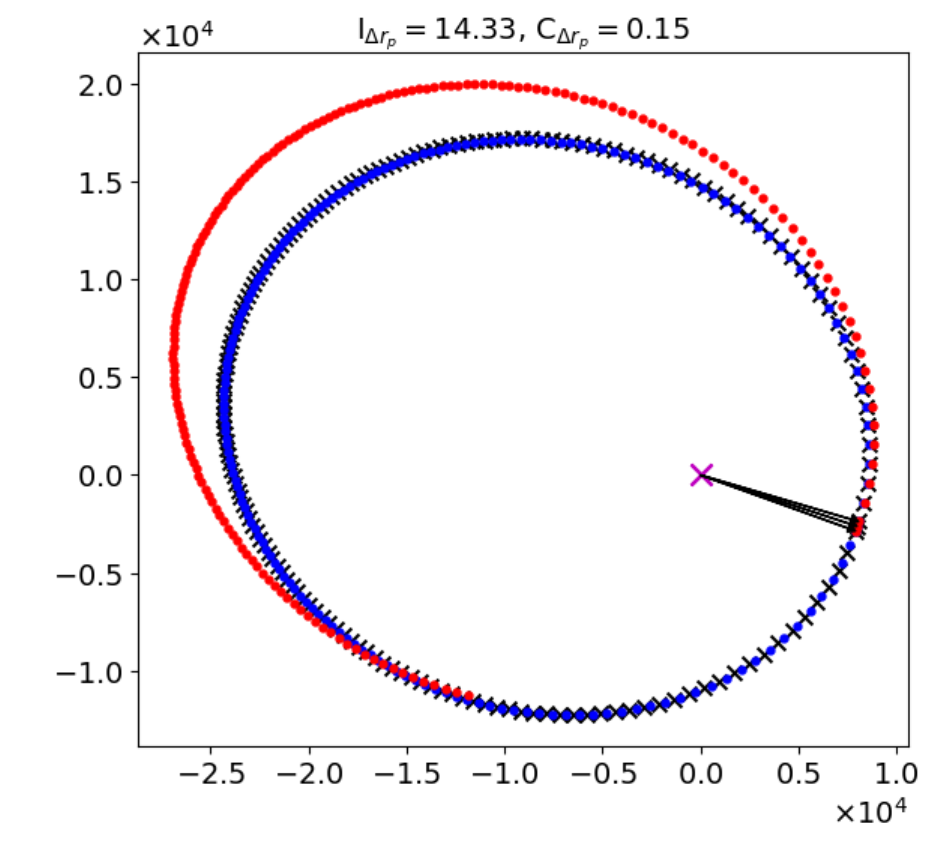}}
    \subfloat{\includegraphics[height=5.3cm]{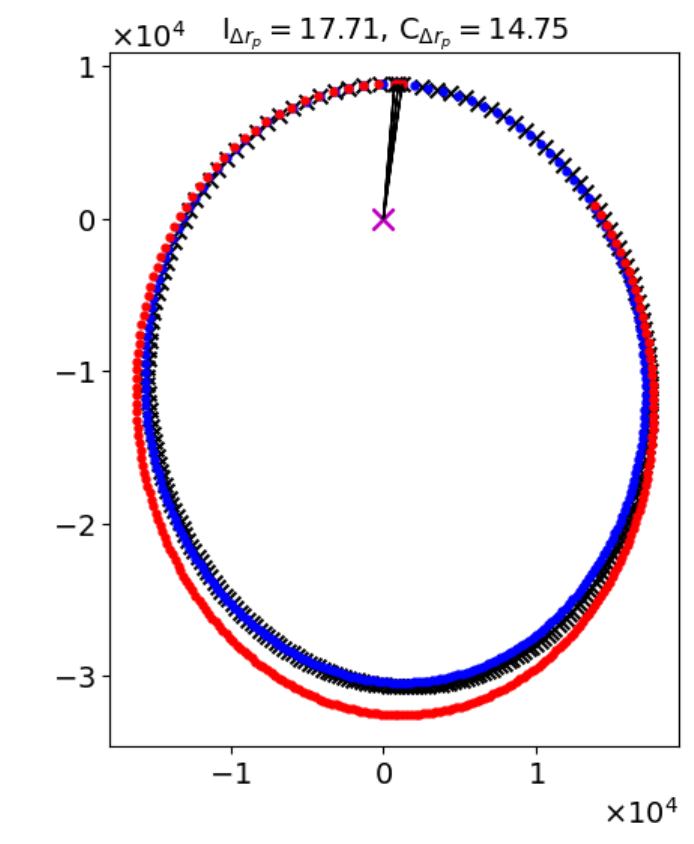}}
    \subfloat{\includegraphics[height=5.3cm]{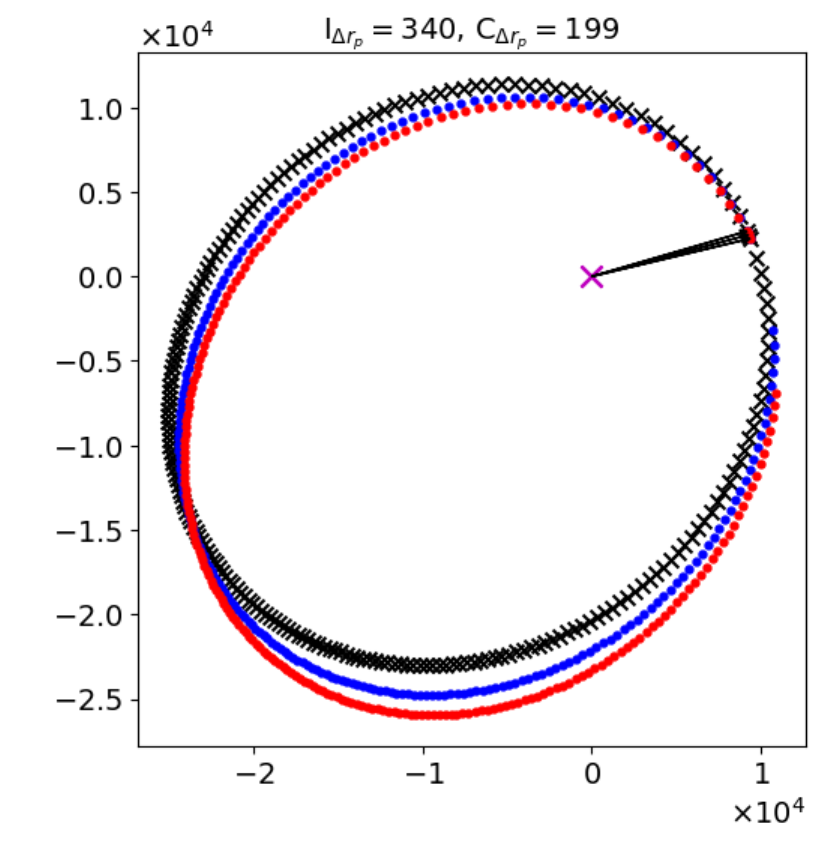}}
    \caption{Visualizations of the initial (red dots) and converged orbits (blue dots) of D-IOD. The simulated orbits (ground-truth) are represented by black 'x', and the magenta `x' denotes the Earth's center. The three black arrows point to the three captured orbital positions. Each plot is labelled with the corresponding periapsis errors (km) I$_{\Delta_{r_p}}$ (for the initial estimate) and C$_{\Delta_{r_p}}$ (for the converged estimate). From \textbf{top} to \textbf{bottom} rows: orbits from A, B, C, and D. The orbits are transformed to perifocal coordinate (PQW) systems, and viewed from +W axis for better visualizations.}
    \label{fig:orbit_vis}
\end{figure*}

We also provide a visual comparison of the initial and final orbital solutions in Fig.~\ref{fig:orbit_vis}. The examples were sampled from different periapsis error ranges - left and right columns were sampled from the low to median error ranges, while right column show failure examples. The qualitative improvements can be seen in these examples, where the converged (blue) orbital solutions fit the ground-truth (in black) much better than the initial estimates (in red). 

\begin{figure}[!tb]
    \centering
    \subfloat{\includegraphics[width=0.32\textwidth]{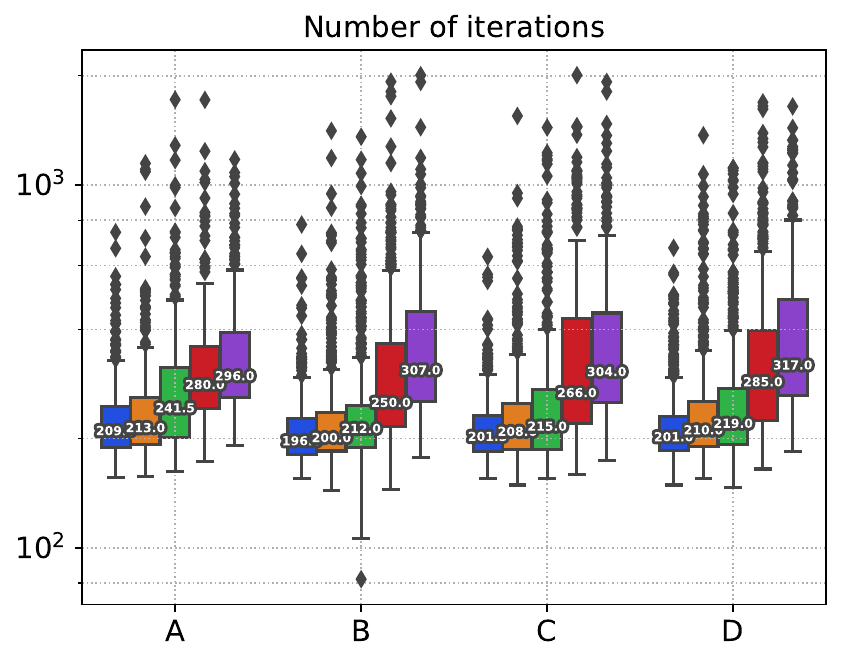}}\\
    \subfloat{\includegraphics[width=0.32\textwidth]{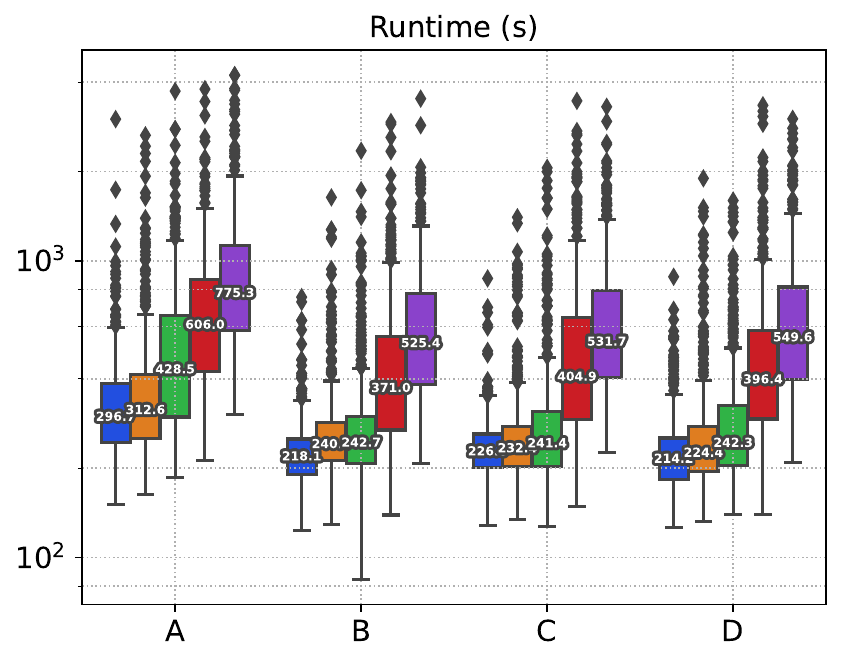}}
    \caption{Comparison of the convergence speed of D-IOD in term of number of iterations and runtime for four distinct orbit types across five initialization levels. The X-axes represent the orbit types. The Y-axes of the top and bottom plot represent the iteration count and seconds, respectively. The box plots are color coded to represent the levels - blue for I, orange for II, green for III, red for IV, and purple for V. The Q2 (median) of the runtimes are labeled on the boxplots, and the Q1 (first quartile) and Q3 (first quartile) of the runtimes are represented by the bottom and top of each box plot.}
    \label{fig:init_exp_time}\vspace{-3mm}
\end{figure}
\paragraph{Runtime} The number of iterations and runtime of D-IOD are plotted in Fig.~\ref{fig:init_exp_time}. The speed of convergence of D-IOD is affected by two factors: the image size and the quality of initialization. The first factor can be observed in the figure where the runtime for orbit type A (in blue) is consistently slower due to its larger image size (see Table~\ref{tab:test_cases}). The second factor is also obvious in the increasing pattern across the levels.

\subsubsection{Time interval experiments}\label{sec:time_exp}
In real application settings, the time interval between the streak images from an RSO varies due to factors such as observing geometry and the speed of the RSO. We generated streak images that were captured over different time intervals and evaluated the performance of D-IOD (both \textit{refine} and \textit{end-to-end} modes) against them in this experiment. 

\paragraph{Setup} The experiment includes three time intervals: 60s, 120s, and 240s between the first image and the third image. Meanwhile, the corresponding time intervals between the first image and the second image are randomly sampled with a mean (and a standard deviation of) of 30s (10s), 60s (15s), and 120s (20s), respectively. The SNR of the images is fixed to 4 in this experiment.

\paragraph{Initialization} For the \textit{refine} mode, we provided D-IOD with level III initialization (detailed in Sec.~\ref{sec:init_exp}). D-IOD$_{\rm{end-to-end}}$ initializes by running the two-stage method without the LOS extraction module. D-IOD backprojects the corner of the observed streak images to obtain the LOS vectors. Specifically, the corner that is close to the beginning of the streak, which we assume is given as part of the meta-data. We highlight that the two-stage method shares a similar assumption.\\

\begin{figure}[!tb]
    \centering
    \subfloat{\includegraphics[width=0.23\textwidth]{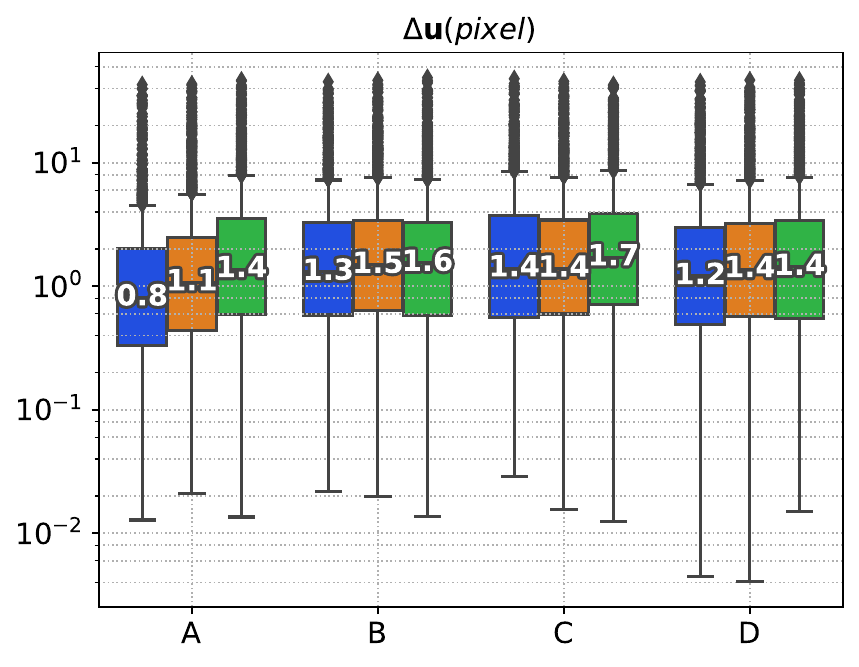}}
    \subfloat{\includegraphics[width=0.23\textwidth]{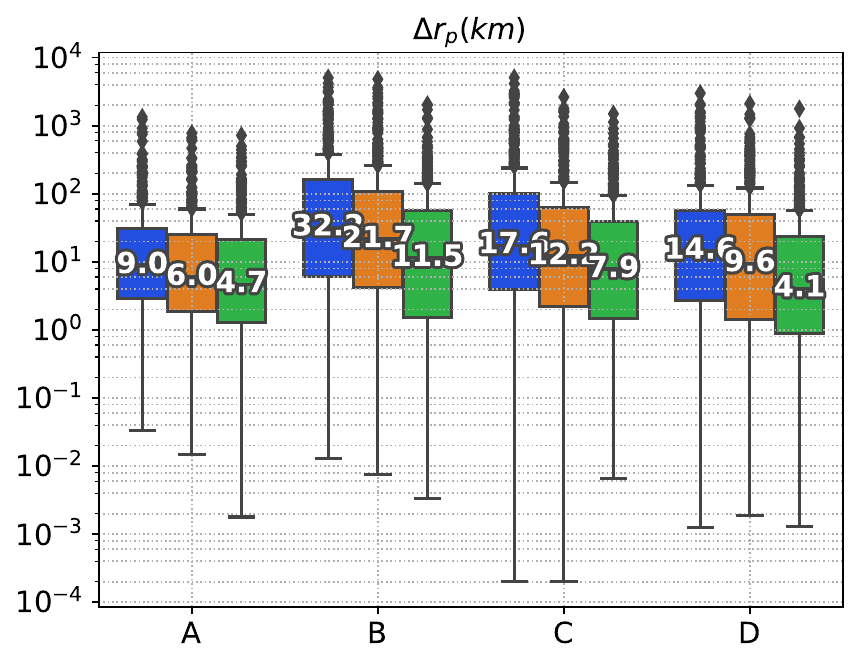}}\\
    \subfloat{\includegraphics[width=0.23\textwidth]{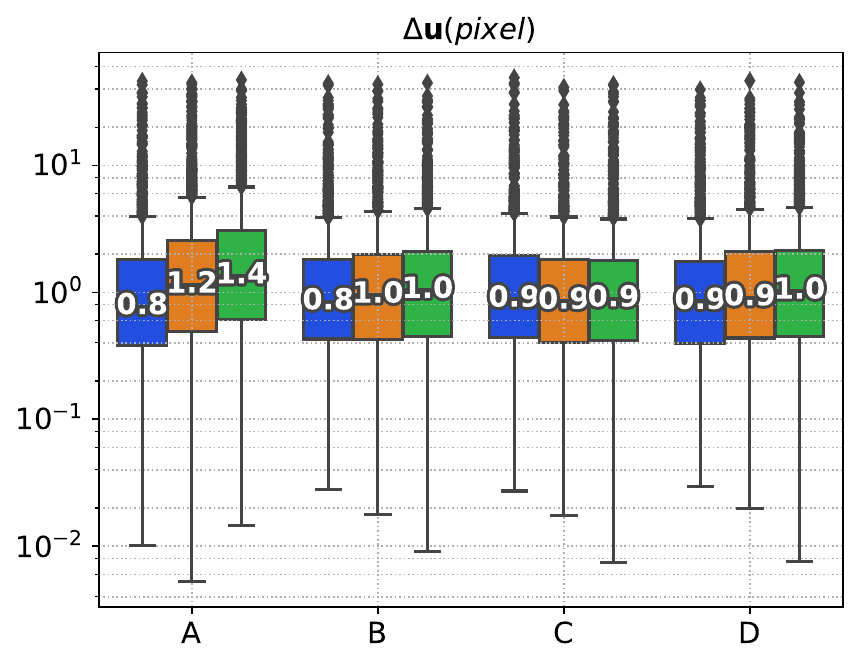}}
    \subfloat{\includegraphics[width=0.23\textwidth]{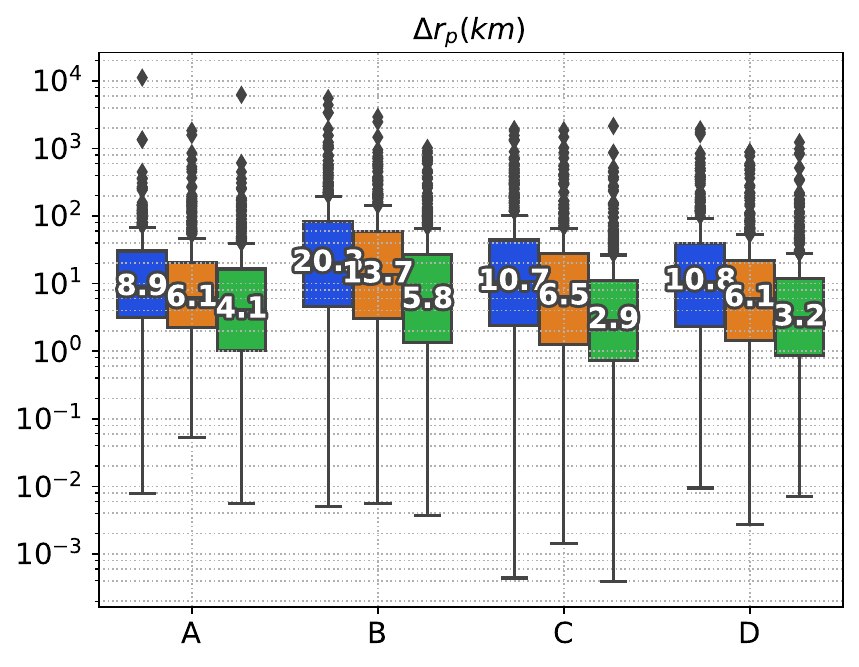}}\\
    \caption{Comparison of endpoints' error $\Delta \mathbf{u}$ and periapsis error $\Delta r_p$ for four orbit types (A, B, C, and D) across three time intervals. The X-axes represent the orbit types. The Y-axes represent pixel distance and $km$ for $\Delta \mathbf{u}$ and $\Delta r_p$ plots, respectively. The box plots are color coded to represent the time interval: blue for 60s, orange for 120s, and green for 240s. \textbf{Top row}: D-IOD$_{\rm{refine}}$ mode, \textbf{bottom row}: D-IOD$_{\rm{end-to-end}}$ mode.}
    \label{fig:time_int_exp}
\end{figure}

\paragraph{Results} As shown in Fig.~\ref{fig:time_int_exp}, the endpoints' errors of the converged solutions are consistently low. The median of the endpoints' errors for the \textit{refine} mode and the \textit{end-to-end} mode range from 0.8 to 1.7 pixels and 0.8 to 1.4 pixels, respectively. The generally lower errors of the \textit{end-to-end }mode, as observed in the bottom row of  Fig.~\ref{fig:time_int_exp}, stem from its better initial estimate quality. Specifically, the endpoints' error of the initial estimate for the \textit{refine} mode is approximately 60 pixels, which is roughly two times larger than the \textit{end-to-end} mode. This is because the corner of the cropped image is usually closer to the streak than the simulated level III, which is selected to display the robustness of the \textit{refine} mode in the remaining experiments.

Despite having similar ranges of $\Delta \mathbf{u}$ for all three testing time intervals, the periapsis error (and other orbital-element errors) is lower for test cases with a larger time interval. This aligns with the results in the initialization experiment, i.e., images that capture a larger orbital arc (due to a larger time interval in this case) better constrain the orbital solution space. Likewise, the full result table and figures can be found in Appendix~\ref{app:time} (Table~\ref{Tab:time_exp_1} to ~\ref{Tab:time_exp_3}).

\begin{figure}[!htb]
    \centering
    \subfloat{\includegraphics[width=0.23\textwidth]{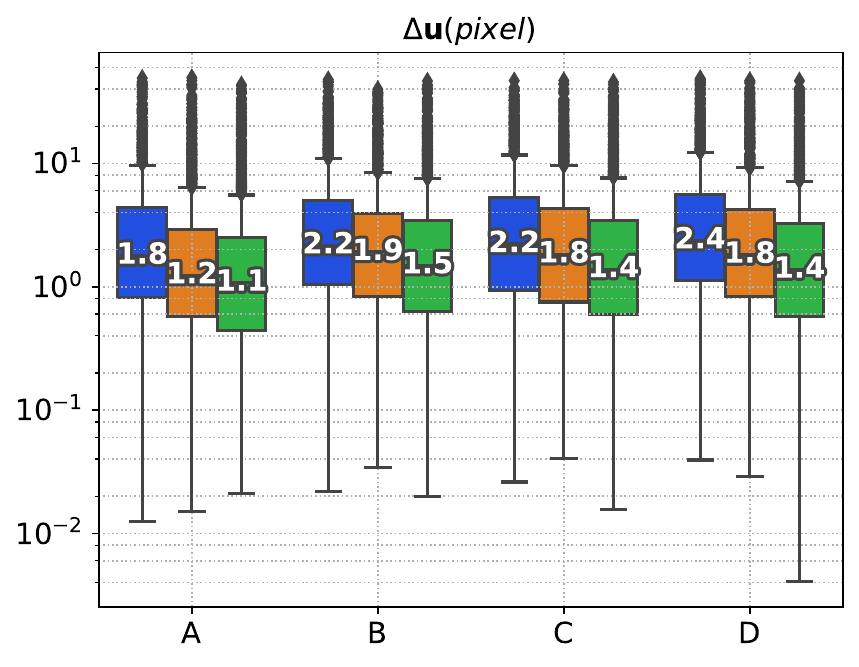}}
    \subfloat{\includegraphics[width=0.23\textwidth]{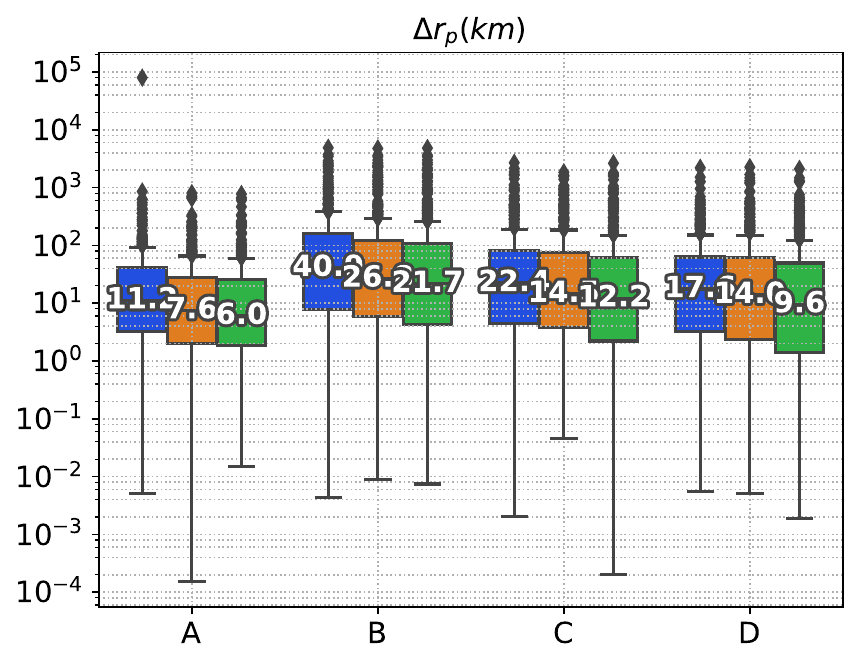}}\\
    \subfloat{\includegraphics[width=0.23\textwidth]{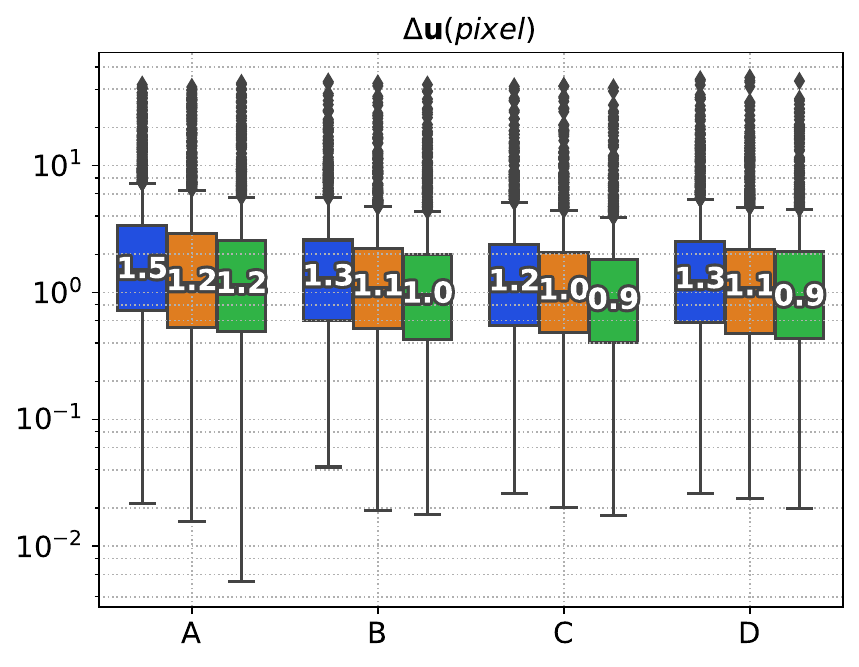}}
    \subfloat{\includegraphics[width=0.23\textwidth]{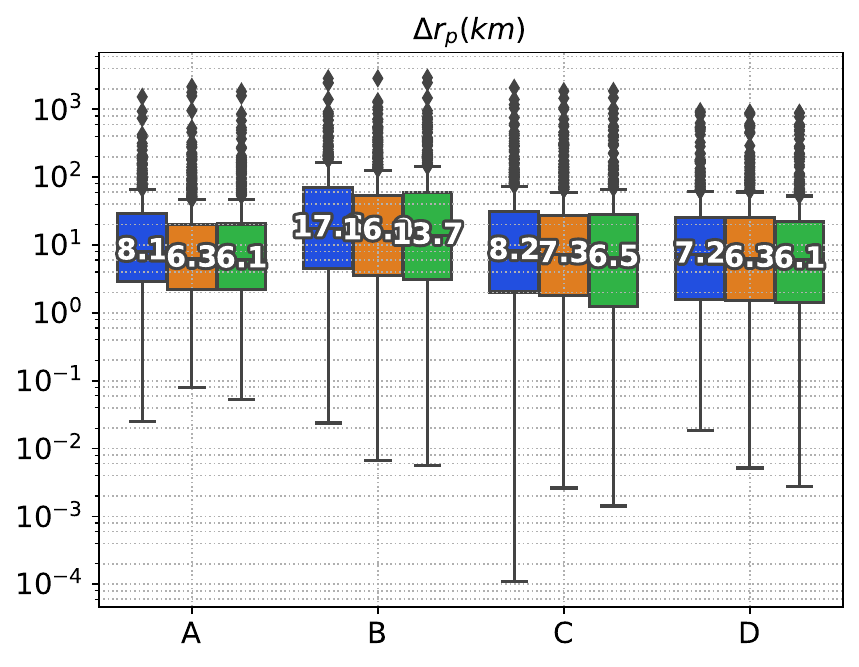}}\\
    \caption{Comparison of endpoints' error $\Delta \mathbf{u}$ and periapsis error $\Delta r_p$ for four orbit types (A, B, C, and D) across three SNRs. The X-axes represent the orbit types. The Y-axes represent pixel distance and $km$ for $\Delta \mathbf{u}$ and $\Delta r_p$ plots, respectively. The box plots are color coded to differentiate the SNRs: blue for 2, orange for 3, and green for 4. \textbf{Top row}: D-IOD$_{\rm{refine}}$ mode, \textbf{bottom row}: D-IOD$_{\rm{end-to-end}}$ mode.}
    \label{fig:snr_exp}
\end{figure}

\subsubsection{Signal-to-noise ratio experiment}\label{sec:snr_exp}
The SNR of an streak image varies depending on factors such as the imaging condition and the RSO size. As such, we simulated streak images of three different noise levels to evaluate the robustness of D-IOD against them.

\paragraph{Setup} We added zero-mean Gaussian intensity noise to the testing images to simulate different SNRs. The chosen sigmas are 0.25, 0.33, and 0.5, corresponding to the SNR of 4, 3, and 2. Fig. \ref{fig:diff_SNR} provides imagery examples. The fixed variable in this experiment is the time interval between the furthest images, which we set to 120s (detailed in Sec.~\ref{sec:time_exp}).

\paragraph{Initialization} The initialization scheme is the same as the time interval experiment above.

\paragraph{Results} As expected, the endpoints' errors, as seen in Fig.~\ref{fig:snr_exp}, are lower for higher SNR. As the SNR increases from 2 to 4, the average median $\Delta \mathbf{u}$ for \textit{refine} mode decreases from approximately 2.15 pixels to 1.35 pixels. Consistent with the observation in the time interval experiment above, the average median $\Delta \mathbf{u}$ for the \textit{end-to-end} mode is generally lower, which decreases from approximately 1.4 pixels to 1 pixel. A similar declining periapsis error can be observed in Fig.~\ref{fig:snr_exp} as well. See Appendix~\ref{app:snr} (Table~\ref{Tab:SNR_1} to Table~\ref{Tab:SNR_3}) for the full results.

\subsubsection{Real data experiments}\label{sec:real_data_exp}
We evaluated the robustness of D-IOD against real streak images in this experiment. Note that the real streak images we used are not associated. So we performed only single-image fitting with D-IOD. Besides, they were not provided with orbital information. As such, we evaluated only the ability of D-IOD in fitting the streak here by evaluating its endpoints' error. However, recall that the endpoints' error is strongly correlated to the orbital accuracy, as observed in our simulated experiments above.

\paragraph{Setup} We manually annotated the endpoints of fifty real streaks images for this experiment.

\paragraph{Initialization} We ran D-IOD with the \textit{end-to-end} mode since the initial estimates are not available. Contrast to the setting where we have multiple streak images, D-IOD backprojects three pixels, i.e., two from the image corners that are closed to the endpoints of the streak and one from the middle, to obtain the LOS vectors for its initialization scheme.

\begin{figure}[!htb]
    \centering
    % \subfloat{\includegraphics[width=0.5\textwidth, keepaspectratio]{39055.pdf}}\\
    \subfloat{\includegraphics[width=0.5\textwidth, keepaspectratio]{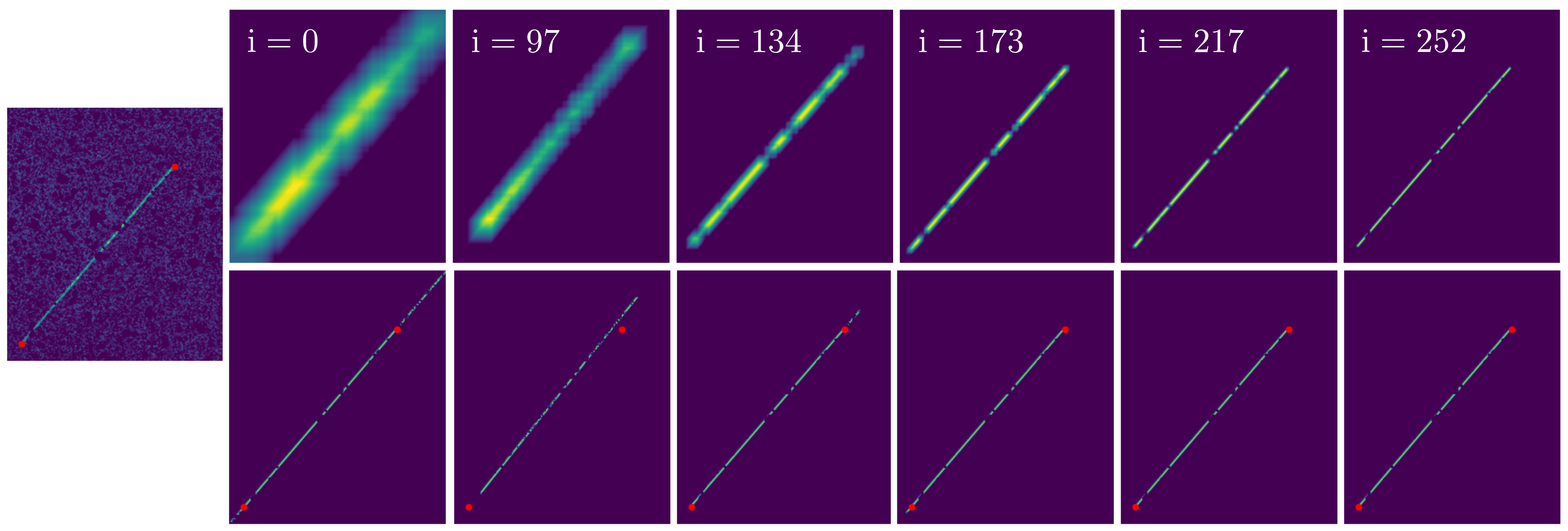}}\\
    \subfloat{\includegraphics[width=0.5\textwidth]{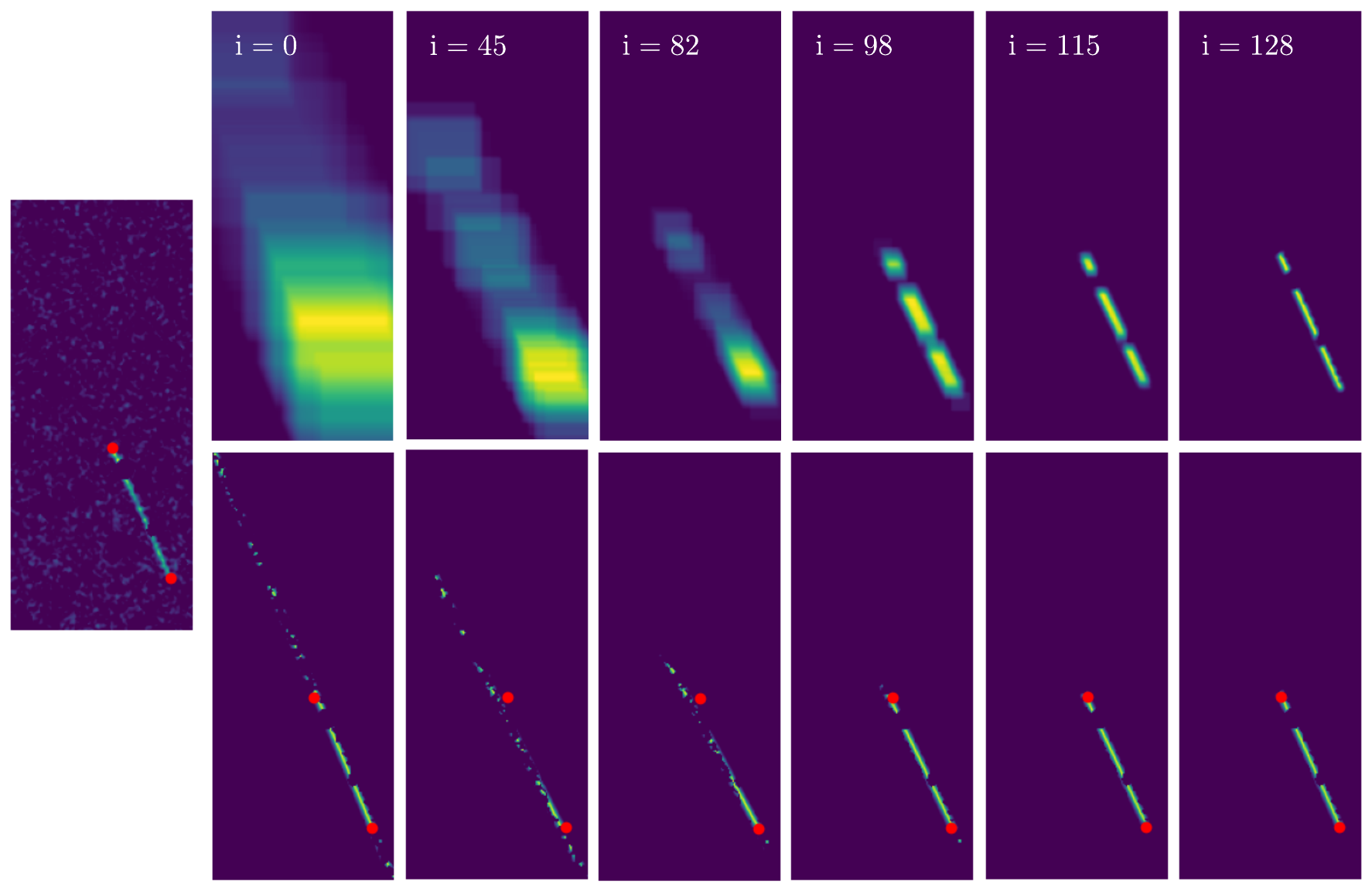}}\\
    \subfloat{\includegraphics[width=0.5\textwidth]{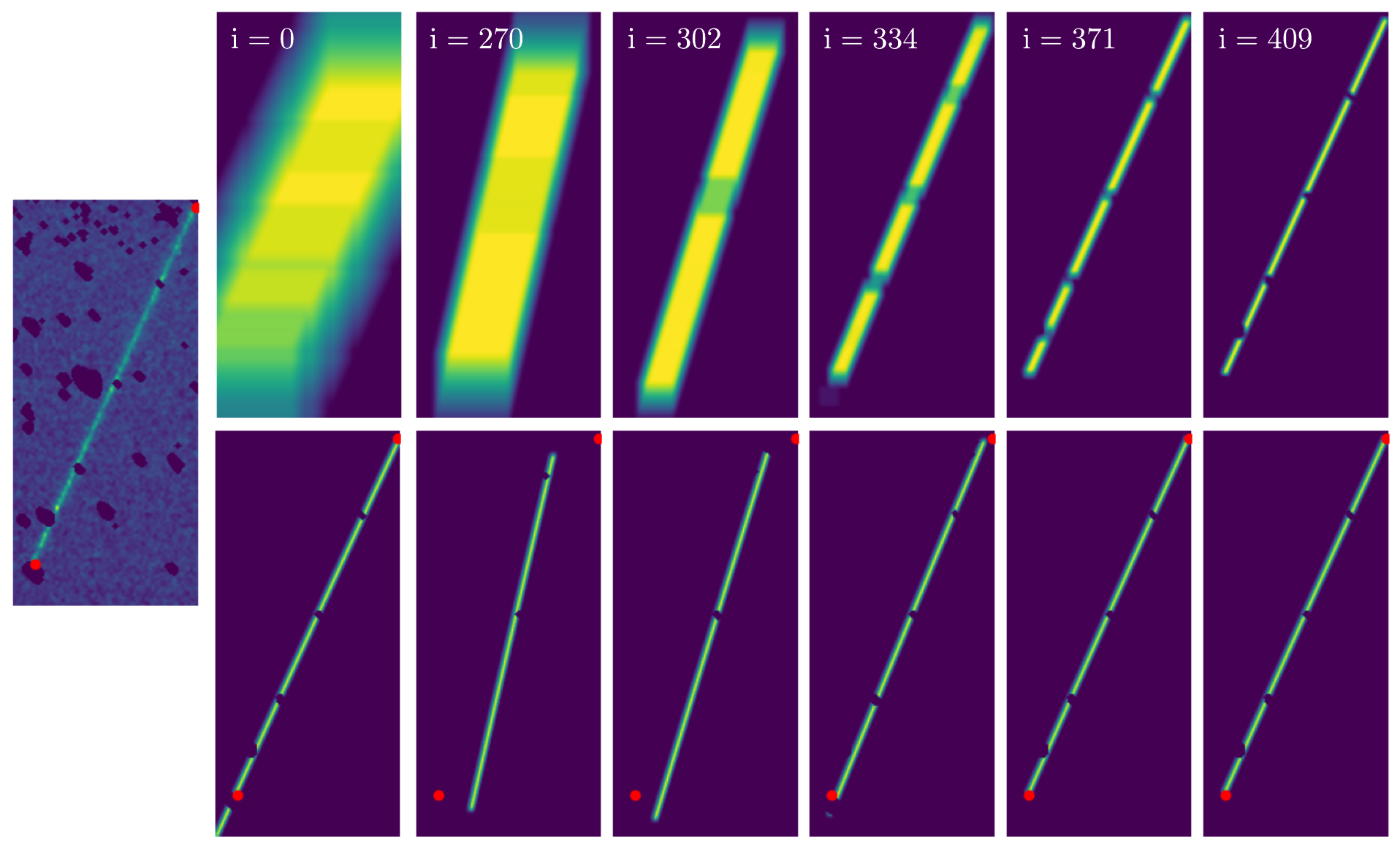}}\\
    \subfloat{\includegraphics[width=0.5\textwidth]{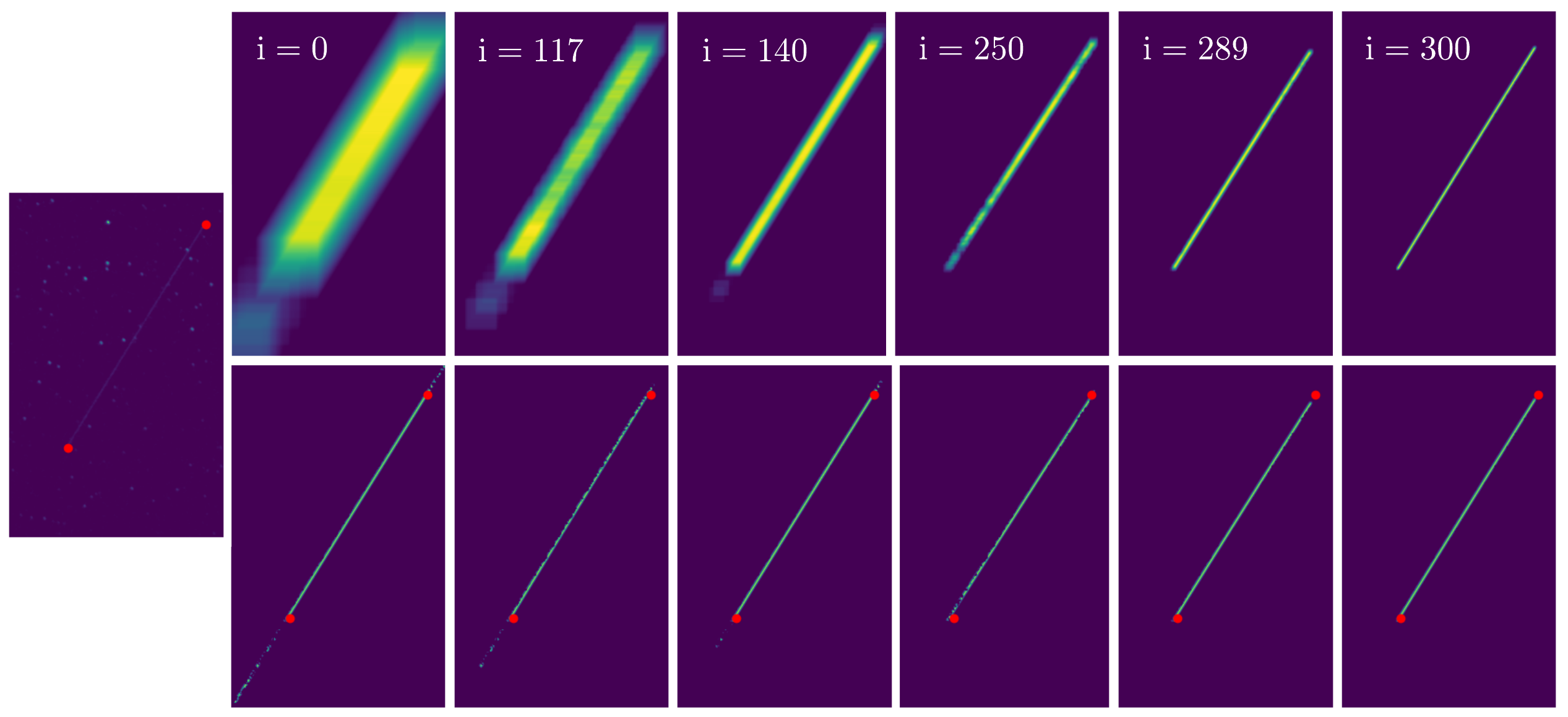}}\\
    \caption{The iterative improvement of D-IOD on real streak images. The real streak images in each example are displayed on the left column. The annotation of the endpoints are highlighted by the red circles. The bottom row of each example shows the generated streak images and the top row depicts the corresponding blurred version. The current iteration count $i$ and the kernel size $k$ are written on the blurred streak images.}\label{fig:real_data_exp_success_1}\vspace{-3mm}
\end{figure}

\paragraph{Results} The median endpoints' error of D-IOD is 1.59 pixels, which is similar to the range of our simulated experiments. We show several examples of the real streak image fitting process in Fig.~\ref{fig:real_data_exp_success_1}. These examples highlight the robustness of D-IOD against artifacts caused by the star removal process and poor imaging conditions. 

\begin{figure}[!htb]
    \centering
    \subfloat{\includegraphics[height=3.8cm]{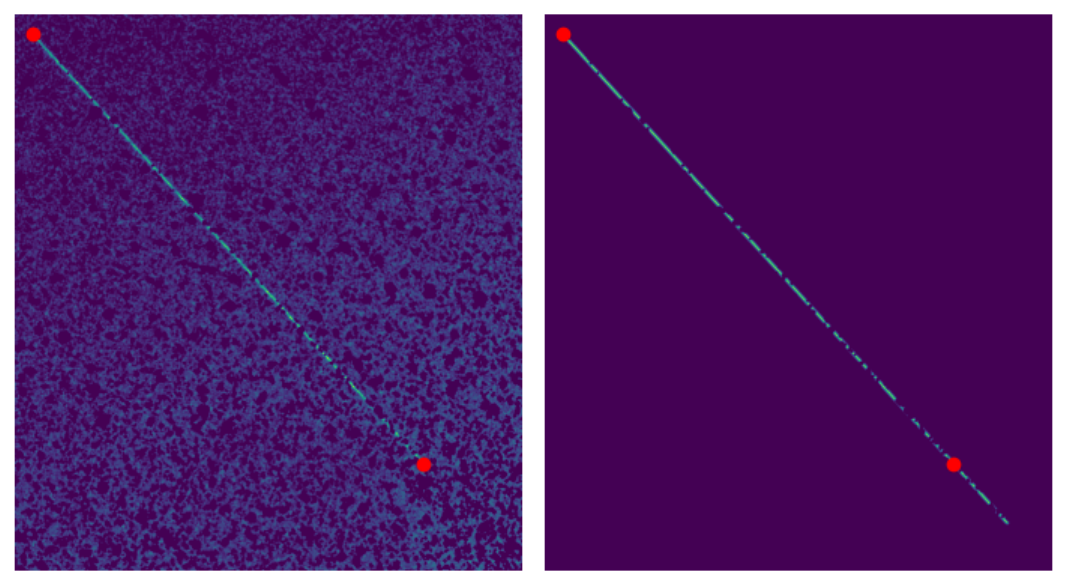}}\\
    \subfloat{\includegraphics[height=2.5cm]{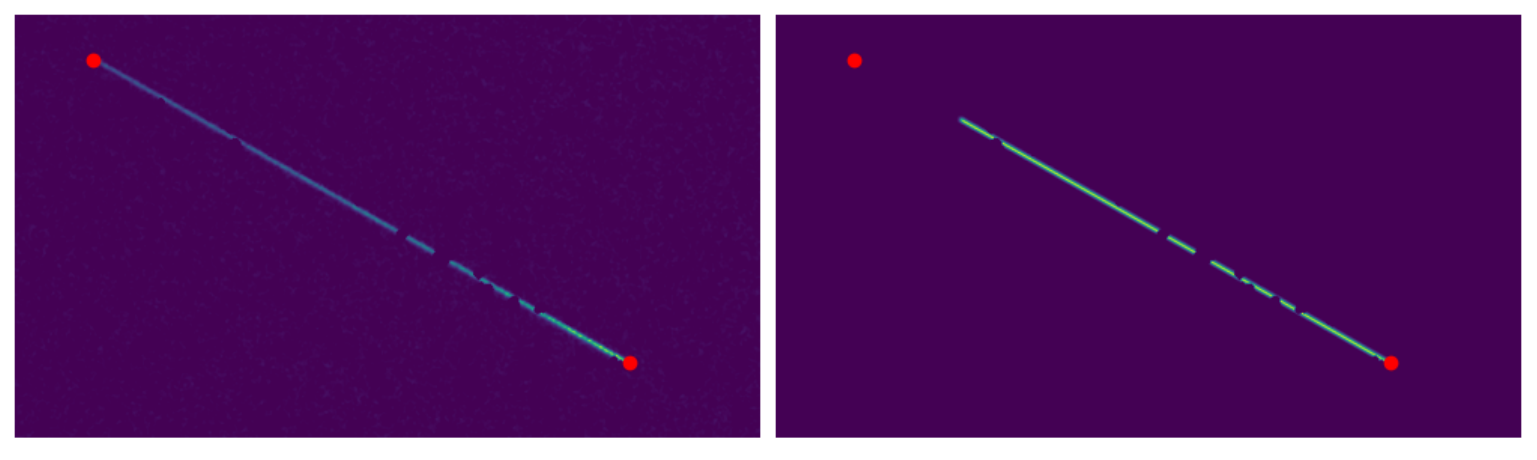}}\\
    \subfloat{\includegraphics[height=1.5cm]{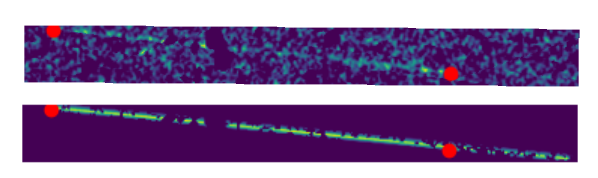}}\\
    \caption{Failure examples of D-IOD on real streak images.}\label{fig:real_data_exp_failure}\vspace{-3mm}
\end{figure}

We show some failure examples in Fig.~\ref{fig:real_data_exp_failure}. These examples showcase some challenging conditions. In the first example, D-IOD fails to fit the lower right corner endpoint that is drowned by the brighter background region. Meanwhile, the second example shows an example where the streak has uneven intensity, where the right end is visibly much brighter than the left end. Similarly, D-IOD fails to fit to the fainter (left) end in this example. We highlight that the challenge here is the uneven intensity and not low intensity which D-IOD has shown to be robust against in the last example of Fig.~\ref{fig:real_data_exp_success_1}. Lastly, the background noise in the third example is as bright as the segmented streak, which also causes problems for D-IOD.

\section{Limitations}\label{sec:limitation}
We discuss several limitations of D-IOD here. Firstly, our proposed model is simplistic by design since this is a proof-of-concept paper to put forward a new IOD paradigm. In the application where the streak images are days or months apart, a more robust propagator such as the SGP4 \cite[Chapter 9]{vallado2001fundamentals} is needed. Besides, the Gaussian PSF could also be replaced with a more sophisticated model such as the Airy disk \citep{airy1835diffraction}. 

Secondly, the current implementation of D-IOD is slow - it takes approximately 250 seconds to converge. Although, we highlight that more than 80\% of the runtime was occupied by the image formation process which can be sped up with parallel computing. Specifically, given $N$ timestamps, a simple parallelization strategy is to generate $N$ copies of images before the long-exposure operation that sums all of them to form $\mathbf{S}$.

Lastly, we tested D-IOD only on images with visible streaks. Under the 5-second time exposure setting, RSOs in most of the MEO and GEO regions (9400 km and above) would produce very short streaks or point sources. We suspect that convergence could be an issue since these point sources a share similar pattern with background noises.

\section{Conclusion}\label{sec:conclusion}
We presented D-IOD, a direct approach to solve the IOD problem in this paper. The proposed method is driven by the principle of \textit{making full use of the available data}. D-IOD iteratively refines the orbital estimate by minimizing our proposed objective function, i.e., the deviations between the generated and observed streak images. Apart from the optimization formulation, we introduced a series of optimization strategies that were inspired by the computer vision literature. D-IOD showcases its robustness against various testing scenarios in both simulated and real data experiments. The significant improvement of D-IOD given poor initial orbital estimates demonstrates its practicality in enhancing the existing two-stage IOD pipeline. Last but not least, we also discussed several future developments for the direct orbit fitting regime.

\section{Acknowledgements}
Chee-Kheng Chng was funded by Lockheed Martin Australia. Tat-Jun Chin is SmartSat CRC Professorial Chair of Sentient Satellites. The imaging data in this paper were provided by a network of widefield optical staring sensors called FireOPAL. FireOPAL is a research project funded by a partnership between Lockheed Martin Australia and the Space Science Technology Centre at Curtin University.

\section{Appendices}
\subsection{Appendix A - Full results for the initialization experiments}\label{app:init}
\setcounter{table}{0}
\renewcommand{\thetable}{A\arabic{table}}
Table~\ref{Tab:init_exp_1} to~\ref{Tab:init_exp_5} contains all three quartiles (i.e., Q1, Q2, and Q3) for all metrics (i.e., $\Delta \mathbf{u}$, $\Delta r_p$, $\Delta e$, $\Delta i$, $\Delta \Omega$, $\Delta \omega$, and $\Delta f$) in the initialization experiment (Sec.\ref{sec:init_exp}). For circular orbits (A), $\Delta \omega$, and $\Delta f$ are not reported because they are undefined. 

\subsection{Appendix B - Full results for the time interval experiments}\label{app:time}
\setcounter{table}{0}
\renewcommand{\thetable}{B\arabic{table}}
The same information (as above) for the time interval experiments (Sec.~\ref{sec:time_exp}) are tabulated in Table~\ref{Tab:time_exp_1},~\ref{Tab:time_exp_2} and~\ref{Tab:time_exp_3}. These tables supplement Fig.~\ref{fig:time_int_exp}.

\subsection{Appendix C - Full results for the SNR experiments}\label{app:snr}
\setcounter{table}{0}
\renewcommand{\thetable}{C\arabic{table}}
The same information (as above) for the SNR experiments (Sec.~\ref{sec:snr_exp}) are tabulated in Table~\ref{Tab:SNR_1},~\ref{Tab:SNR_2} and~\ref{Tab:SNR_3}. These tables supplement Fig.~\ref{fig:snr_exp}.

\bibliographystyle{jasr-model5-names}
\biboptions{authoryear}
\bibliography{sample.bib}

\begin{table*}[!htb]
	\begin{center}
		\caption{Initialization experiment - Initial estimate with Level I quality. Legends: `Q' stands for quartiles, `Init.' for initial estimates, and `Conv.' for converged solutions.}\label{Tab:init_exp_1}
		\begin{tabular}{cccccccccc}
			\toprule
			  \multirow{4}{*}{Metrics}& \multirow{4}{*}{Q} & \multicolumn{8}{c}{Orbit types}\\
			  \cmidrule(lr){3-10}
			    & &   \multicolumn{2}{c}{A} & \multicolumn{2}{c}{B} & \multicolumn{2}{c}{C} & \multicolumn{2}{c}{D} \\
        \cmidrule(lr){3-4}\cmidrule(lr){5-6}\cmidrule(lr){7-8}\cmidrule(lr){9-10}
            &&  Init. & Conv. & Init. & Conv. & Init. & Conv.  & Init. & Conv. \\
			 \cmidrule(lr){1-10}
			 $\Delta \mathbf{u}$ (pixel) &Q1 &0.46 &0.35 &0.43 &0.33 &0.5 &0.35 &0.53 &0.32 \\
            &Q2 &0.72 &0.73 &0.7 &0.66 &0.82 &0.67 &0.9 &0.65 \\
            &Q3 &1.05 &1.61 &1.05 &1.33 &1.3 &1.32 &1.54 &1.27 \\
            \cmidrule(lr){1-10}
            $\Delta r_p$ (km) &Q1 &2.33 &2.89 &2.7 &3.95 &1.9 &1.76 &1.26 &1.89 \\
            &Q2 &5.07 &7.98 &9.72 &16.17 &6.6 &8.68 &4.25 &6.16 \\
            &Q3 &10.1 &26.18 &28.02 &54.86 &19.46 &35.37 &13.27 &24.04 \\
            \cmidrule(lr){1-10}
            $\Delta e$ &Q1 &2.74e-4 &2.91e-4 &6.09e-4 &6.90e-4 &8.24e-4 &7.10e-4 &1.07e-3 &6.61e-4 \\
            &Q2 &7.73e-4 &9.92e-4 &1.59e-3 &2.71e-3 &1.91e-3 &2.45e-3 &2.35e-3 &2.11e-3 \\
            &Q3 &1.62e-3 &4.00e-3 &4.05e-3 &6.84e-3 &4.70e-3 &7.45e-3 &5.27e-3 &5.55e-3 \\
            \cmidrule(lr){1-10}
            $\Delta i \:(\circ)$ &Q1 &4.29e-3 &3.00e-3 &5.54e-3 &5.13e-3 &5.66e-3 &5.43e-3 &5.01e-3 &5.63e-3 \\
            &Q2 &1.17e-2 &1.48e-2 &1.48e-2 &2.24e-2 &1.57e-2 &1.88e-2 &1.49e-2 &2.12e-2 \\
            &Q3 &3.09e-2 &4.54e-2 &3.66e-2 &7.09e-2 &4.05e-2 &7.41e-2 &4.09e-2 &6.21e-2 \\
            \cmidrule(lr){1-10}
            $\Delta \Omega\:(\circ)$ &Q1 &6.25e-3 &6.50e-3 &8.46e-3 &9.68e-3 &8.44e-3 &8.27e-3 &7.52e-3 &9.41e-3 \\
            &Q2 &2.13e-2 &3.02e-2 &2.36e-2 &3.78e-2 &2.51e-2 &3.13e-2 &2.76e-2 &3.62e-2 \\
            &Q3 &6.80e-2 &1.16e-1 &8.88e-2 &1.51e-1 &8.09e-2 &1.13e-1 &6.99e-2 &1.26e-1 \\
            \cmidrule(lr){1-10}
            $\Delta \omega\:(\circ)$ &Q1 &- &- &0.67 &0.69 &0.24 &0.25 &0.13 &0.14 \\
            &Q2 &- &- &1.67 &2.66 &0.59 &0.88 &0.3 &0.49 \\
            &Q3 &- &- &5.04 &7.32 &1.5 &2.51 &0.81 &1.4 \\
            \cmidrule(lr){1-10}
            $\Delta f\:(\circ)$ &Q1 &- &- &0.69 &0.69 &0.24 &0.27 &0.13 &0.16 \\
            &Q2 &- &- &1.72 &2.71 &0.59 &0.87 &0.3 &0.48 \\
            &Q3 &- &- &5.5 &7.5 &1.49 &2.56 &0.81 &1.41 \\
			  \bottomrule
		\end{tabular}
	\end{center}
\end{table*}

\begin{table*}[!htb]
	\begin{center}
		\caption{Initialization experiment - Initial estimate with Level II quality. Legends: `Q' stands for quartiles, `Init.' for initial estimates, and `Conv.' for converged solutions.}\label{Tab:init_exp_2}
		\begin{tabular}{cccccccccc}
			\toprule
			  \multirow{4}{*}{Metrics}& \multirow{4}{*}{Q} & \multicolumn{8}{c}{Orbit types}\\
			  \cmidrule(lr){3-10}
			    & &   \multicolumn{2}{c}{A} & \multicolumn{2}{c}{B} & \multicolumn{2}{c}{C} & \multicolumn{2}{c}{D} \\
        \cmidrule(lr){3-4}\cmidrule(lr){5-6}\cmidrule(lr){7-8}\cmidrule(lr){9-10}
            &&  Init. & Conv. & Init. & Conv. & Init. & Conv.  & Init. & Conv. \\
			 \cmidrule(lr){1-10}
			 $\Delta \mathbf{u}$ (pixel) &Q1 &25.31 &0.38 &19.86 &0.47 &20.52 &0.49 &21.07 &0.41 \\
            &Q2 &34.48 &0.81 &24.45 &1.06 &25.55 &1.1 &26.97 &0.96 \\
            &Q3 &49.13 &2.1 &30.33 &2.25 &32.08 &2.55 &34.02 &2.37 \\
            \cmidrule(lr){1-10}
            $\Delta r_p$ (km) &Q1 &28.38 &2.95 &19.23 &6.23 &11.06 &3.16 &8.03 &1.92 \\
            &Q2 &61.92 &9.97 &85.57 &22.93 &45.19 &17.13 &38.9 &11.04 \\
            &Q3 &127.69 &36.22 &265.91 &113.42 &155.26 &72.63 &144.66 &44.01 \\
            \cmidrule(lr){1-10}
            $\Delta e$ &Q1 &3.86e-3 &3.13e-4 &5.28e-3 &1.04e-3 &6.30e-3 &1.14e-3 &6.04e-3 &8.57e-4 \\
            &Q2 &9.74e-3 &1.33e-3 &1.26e-2 &3.57e-3 &1.43e-2 &4.68e-3 &1.26e-2 &3.11e-3 \\
            &Q3 &2.39e-2 &4.98e-3 &3.67e-2 &1.49e-2 &3.66e-2 &1.64e-2 &3.43e-2 &1.19e-2 \\
            \cmidrule(lr){1-10}
            $\Delta i \:(\circ)$ &Q1 &2.63e-2 &4.02e-3 &2.74e-2 &8.08e-3 &2.56e-2 &8.44e-3 &3.10e-2 &7.52e-3 \\
            &Q2 &9.00e-2 &1.49e-2 &1.03e-1 &3.30e-2 &9.74e-2 &3.54e-2 &1.19e-1 &3.57e-2 \\
            &Q3 &3.42e-1 &6.16e-2 &3.10e-1 &1.18e-1 &3.45e-1 &1.40e-1 &3.70e-1 &1.24e-1 \\
            \cmidrule(lr){1-10}
            $\Delta \Omega\:(\circ)$ &Q1 &5.02e-2 &7.02e-3 &4.66e-2 &1.62e-2 &4.33e-2 &1.29e-2 &4.61e-2 &1.31e-2 \\
            &Q2 &1.72e-1 &3.11e-2 &1.99e-1 &6.58e-2 &1.54e-1 &7.36e-2 &1.96e-1 &5.31e-2 \\
            &Q3 &6.56e-1 &1.27e-1 &7.01e-1 &2.53e-1 &6.76e-1 &2.40e-1 &6.44e-1 &2.16e-1 \\
            \cmidrule(lr){1-10}
            $\Delta \omega\:(\circ)$ &Q1 &- &- &3.87 &0.99 &1.49 &0.39 &0.83 &0.21 \\
            &Q2 &- &- &13.01 &3.94 &4.22 &1.71 &2.64 &0.83 \\
            &Q3 &- &- &38.4 &16.59 &11.97 &5.18 &8.11 &2.55 \\
            \cmidrule(lr){1-10}
            $\Delta f\:(\circ)$ &Q1 &- &- &3.9 &1.01 &1.49 &0.41 &0.87 &0.2 \\
            &Q2 &- &- &14.2 &3.97 &4.75 &1.71 &2.6 &0.82 \\
            &Q3 &- &- &52.99 &18.08 &15.81 &5.66 &8.66 &2.61 \\
			  \bottomrule
		\end{tabular}
	\end{center}
\end{table*}

\begin{table*}[!htb]
	\begin{center}
		\caption{Initialization experiment - Initial estimate with Level III quality. Legends: `Q' stands for quartiles, `Init.' for initial estimates, and `Conv.' for converged solutions.}\label{Tab:init_exp_3}
		\begin{tabular}{cccccccccc}
			\toprule
			  \multirow{4}{*}{Metrics}& \multirow{4}{*}{Q} & \multicolumn{8}{c}{Orbit types}\\
			  \cmidrule(lr){3-10}
			    & &   \multicolumn{2}{c}{A} & \multicolumn{2}{c}{B} & \multicolumn{2}{c}{C} & \multicolumn{2}{c}{D} \\
        \cmidrule(lr){3-4}\cmidrule(lr){5-6}\cmidrule(lr){7-8}\cmidrule(lr){9-10}
            &&  Init. & Conv. & Init. & Conv. & Init. & Conv.  & Init. & Conv. \\
			 \cmidrule(lr){1-10}
			  $\Delta \mathbf{u}$ (pixel) &Q1 &52.03 &0.33 &40.15 &0.54 &41.84 &0.53 &41.91 &0.48 \\
                &Q2 &69.12 &0.78 &50.12 &1.25 &51.81 &1.33 &53.68 &1.14 \\
                &Q3 &99.46 &1.9 &61.24 &3.05 &64.19 &3.56 &67.2 &2.89 \\
                \cmidrule(lr){1-10}
                $\Delta r_p$ (km) &Q1 &55.75 &2.9 &39.99 &6.03 &17.6 &3.93 &20.69 &2.71 \\
                &Q2 &129.74 &9 &177.81 &32.21 &78.98 &17.62 &75.94 &14.63 \\
                &Q3 &273.55 &30.71 &654.91 &160.21 &314.61 &100.9 &276.27 &56.66 \\
                \cmidrule(lr){1-10}
                $\Delta e$ &Q1 &8.43e-3 &3.12e-4 &9.97e-3 &1.22e-3 &9.98e-3 &1.38e-3 &1.29e-2 &1.08e-3 \\
                &Q2 &2.31e-2 &1.04e-3 &2.44e-2 &5.18e-3 &2.83e-2 &5.01e-3 &2.81e-2 &3.79e-3 \\
                &Q3 &5.58e-2 &4.88e-3 &9.17e-2 &2.09e-2 &7.10e-2 &2.28e-2 &6.80e-2 &1.42e-2 \\
                \cmidrule(lr){1-10}
                $\Delta i \:(\circ)$ &Q1 &5.68e-2 &3.73e-3 &4.47e-2 &1.06e-2 &4.40e-2 &8.76e-3 &5.74e-2 &9.93e-3 \\
                &Q2 &2.11e-1 &1.59e-2 &2.11e-1 &4.40e-2 &1.94e-1 &4.39e-2 &2.84e-1 &3.81e-2 \\
                &Q3 &6.93e-1 &7.16e-2 &6.00e-1 &2.02e-1 &6.40e-1 &1.97e-1 &8.00e-1 &1.64e-1 \\
                \cmidrule(lr){1-10}
                $\Delta \Omega\:(\circ)$ &Q1 &9.34e-2 &6.17e-3 &8.29e-2 &1.80e-2 &7.65e-2 &1.38e-2 &9.89e-2 &1.77e-2 \\
                &Q2 &3.49e-1 &2.98e-2 &3.59e-1 &8.86e-2 &2.92e-1 &6.61e-2 &4.32e-1 &7.35e-2 \\
                &Q3 &1.42  &1.26e-1 &1.40  &3.62e-1 &1.29  &3.79e-1 &1.56  &2.99e-1 \\
                \cmidrule(lr){1-10}
                $\Delta \omega\:(\circ)$ &Q1 &- &- &7.69 &1.36 &2.75 &0.44 &1.94 &0.25 \\
                &Q2 &- &- &24.95 &5.53 &7.81 &1.79 &6.34 &0.98 \\
                &Q3 &- &- &67.38 &23.53 &23.55 &7.07 &18.2 &3.76 \\
                \cmidrule(lr){1-10}
                $\Delta f\:(\circ)$ &Q1 &- &- &8.93 &1.4 &2.76 &0.45 &2 &0.26 \\
                &Q2 &- &- &25.62 &5.73 &9.06 &1.78 &6.19 &0.98 \\
                &Q3 &- &- &150.33 &27.17 &34.25 &7.61 &24.23 &3.77 \\
			  \bottomrule
		\end{tabular}
	\end{center}
\end{table*}

\begin{table*}[!htb]
	\begin{center}
		\caption{Initialization experiment - Initial estimate with Level IV quality. Legends: `Q' stands for quartiles, `Init.' for initial estimates, and `Conv.' for converged solutions.}\label{Tab:init_exp_4}
		\begin{tabular}{cccccccccc}
			\toprule
			  \multirow{4}{*}{Metrics}& \multirow{4}{*}{Q} & \multicolumn{8}{c}{Orbit types}\\
			  \cmidrule(lr){3-10}
			    & &   \multicolumn{2}{c}{A} & \multicolumn{2}{c}{B} & \multicolumn{2}{c}{C} & \multicolumn{2}{c}{D} \\
        \cmidrule(lr){3-4}\cmidrule(lr){5-6}\cmidrule(lr){7-8}\cmidrule(lr){9-10}
            &&  Init. & Conv. & Init. & Conv. & Init. & Conv.  & Init. & Conv. \\
			 \cmidrule(lr){1-10}
			  $\Delta \mathbf{u}$ (pixel) &Q1 &81.42 &0.36 &65.05 &0.49 &66.12 &0.46 &67.61 &0.42 \\
            &Q2 &115.25 &0.86 &84.1 &1.24 &86.47 &1.17 &87.29 &0.99 \\
            &Q3 &167.08 &2.1 &106.21 &3.25 &108.87 &3.78 &111.62 &2.65 \\
            \cmidrule(lr){1-10}
            $\Delta r_p$ (km) &Q1 &102.02 &3.22 &76.67 &5.65 &51.37 &2.94 &36.37 &1.97 \\
            &Q2 &248.77 &9.44 &302.82 &27.22 &204.12 &18.3 &142.55 &10.01 \\
            &Q3 &509.38 &35.44 &1218.79 &152.59 &796.54 &120.53 &504.45 &59.87 \\
            \cmidrule(lr){1-10}
            $\Delta e$ &Q1 &2.08e-2 &2.83e-4 &2.13e-2 &1.25e-3 &2.34e-2 &9.52e-4 &2.32e-2 &9.21e-4 \\
            &Q2 &4.71e-2 &1.38e-3 &5.22e-2 &4.93e-3 &5.77e-2 &4.11e-3 &5.31e-2 &2.82e-3 \\
            &Q3 &1.06e-1 &5.63e-3 &1.89e-1 &1.95e-2 &1.53e-1 &2.79e-2 &1.44e-1 &1.27e-2 \\
            \cmidrule(lr){1-10}
            $\Delta i \:(\circ)$ &Q1 &1.00e-1 &3.30e-3 &9.83e-2 &8.71e-3 &1.10e-1 &6.83e-3 &1.19e-1 &7.61e-3 \\
            &Q2 &3.99e-1 &1.45e-2 &3.77e-1 &4.22e-2 &4.08e-1 &3.88e-2 &4.33e-1 &3.70e-2 \\
            &Q3 &1.22  &5.75e-2 &1.34  &1.75e-1 &1.35  &1.90e-1 &1.45  &1.19e-1 \\
            \cmidrule(lr){1-10}
            $\Delta \Omega\:(\circ)$ &Q1 &1.91e-1 &6.14e-3 &1.94e-1 &1.71e-2 &1.69e-1 &1.19e-2 &1.74e-1 &1.53e-2 \\
            &Q2 &6.14e-1 &2.81e-2 &7.12e-1 &9.27e-2 &7.26e-1 &6.81e-2 &7.70e-1 &6.58e-2 \\
            &Q3 &2.46  &1.22e-1 &2.58  &3.86e-1 &2.61  &3.94e-1 &2.75  &2.51e-1 \\
            \cmidrule(lr){1-10}
            $\Delta \omega\:(\circ)$ &Q1 &- &- &13.73 &1.17 &6.02 &0.36 &3.2 &0.17 \\
            &Q2 &- &- &37.89 &5.32 &18.56 &1.58 &10.03 &0.77 \\
            &Q3 &- &- &93.06 &23.39 &53.31 &8.95 &33.24 &3.57 \\
            \cmidrule(lr){1-10}
            $\Delta f\:(\circ)$ &Q1 &- &- &14.34 &1.15 &6.28 &0.36 &3.44 &0.18 \\
            &Q2 &- &- &43.68 &5.48 &21.6 &1.56 &11.69 &0.77 \\
            &Q3 &- &- &229.31 &30.74 &252.34 &10.57 &49.94 &3.64 \\
			  \bottomrule
		\end{tabular}
	\end{center}
\end{table*}

\begin{table*}
	\begin{center}
		\caption{Initialization experiment - Initial estimate with Level V quality. Legends: `Q' stands for quartiles, `Init.' for initial estimates, and `Conv.' for converged solutions.}\label{Tab:init_exp_5}
		\begin{tabular}{cccccccccc}
			\toprule
			  \multirow{4}{*}{Metrics}& \multirow{4}{*}{Q} & \multicolumn{8}{c}{Orbit types}\\
			  \cmidrule(lr){3-10}
			    & &   \multicolumn{2}{c}{A} & \multicolumn{2}{c}{B} & \multicolumn{2}{c}{C} & \multicolumn{2}{c}{D} \\
        \cmidrule(lr){3-4}\cmidrule(lr){5-6}\cmidrule(lr){7-8}\cmidrule(lr){9-10}
            &&  Init. & Conv. & Init. & Conv. & Init. & Conv.  & Init. & Conv. \\
			 \cmidrule(lr){1-10}
			 $\Delta \mathbf{u}$ (pixel) &Q1 &124.13 &0.36 &98.18 &0.47 &103.36 &0.44 &102.08 &0.39 \\
                &Q2 &172.69 &0.77 &124.99 &1.18 &133.18 &1.14 &134.08 &0.92 \\
                &Q3 &257.24 &2 &154.78 &3.61 &166.2 &2.99 &167.36 &2.47 \\
                \cmidrule(lr){1-10}
                $\Delta r_p$ (km) &Q1 &161.36 &3.04 &128.72 &5.21 &71.5 &2.45 &46.9 &1.88 \\
                &Q2 &350.25 &9.15 &577.02 &31.77 &273.26 &15.96 &218.5 &9.18 \\
                &Q3 &727.53 &31.36 &1851.32 &193.16 &1147.43 &67.63 &912.3 &49.26 \\
                \cmidrule(lr){1-10}
                $\Delta e$ &Q1 &3.42e-2 &3.18e-4 &3.30e-2 &1.22e-3 &3.36e-2 &9.94e-4 &4.04e-2 &6.76e-4 \\
                &Q2 &6.36e-2 &1.18e-3 &9.62e-2 &5.64e-3 &7.93e-2 &3.81e-3 &9.25e-2 &3.17e-3 \\
                &Q3 &1.54e-1 &3.94e-3 &2.82e-1 &2.44e-2 &2.73e-1 &1.71e-2 &2.74e-1 &1.70e-2 \\
                \cmidrule(lr){1-10}
                $\Delta i \:(\circ)$ &Q1 &1.60e-1 &3.32e-3 &1.54e-1 &8.26e-3 &1.49e-1 &6.92e-3 &1.65e-1 &6.61e-3 \\
                &Q2 &6.08e-1 &1.49e-2 &5.82e-1 &4.35e-2 &5.69e-1 &3.71e-2 &6.28e-1 &2.81e-2 \\
                &Q3 &2.05  &5.03e-2 &1.84  &2.26e-1 &1.70  &1.53e-1 &2.20  &1.43e-1 \\
                \cmidrule(lr){1-10}
                $\Delta \Omega\:(\circ)$ &Q1 &2.36e-1 &6.55e-3 &2.48e-1 &1.94e-2 &2.72e-1 &1.06e-2 &2.75e-1 &1.28e-2 \\
                &Q2 &1.00  &2.65e-2 &9.28e-1 &9.28e-2 &8.85e-1 &5.48e-2 &1.21  &6.07e-2 \\
                &Q3 &3.97  &9.64e-2 &3.64  &3.86e-1 &3.54  &2.86e-1 &4.15  &2.33e-1 \\
                \cmidrule(lr){1-10}
                $\Delta \omega\:(\circ)$ &Q1 &- &- &17.31 &1.1 &7.33 &0.31 &4.75 &0.16 \\
                &Q2 &- &- &53.32 &4.67 &22.35 &1.46 &15.47 &0.7 \\
                &Q3 &- &- &117.37 &30.38 &59.37 &5.24 &40.37 &2.96 \\
                \cmidrule(lr){1-10}
                $\Delta f\:(\circ)$ &Q1 &- &- &19.61 &1.08 &7.95 &0.31 &4.93 &0.16 \\
                &Q2 &- &- &62.45 &4.68 &28.49 &1.45 &18.54 &0.64 \\
                &Q3 &- &- &230.05 &33.33 &273.73 &5.75 &234.92 &3.05 \\
			  \bottomrule
		\end{tabular}
	\end{center}
\end{table*}

\begin{table*}
	\begin{center}
		\caption{Time interval experiment - time interval = 60 seconds.}\label{Tab:time_exp_1}
		\begin{tabular}{cccccccccc}
			\toprule
			  \multirow{4}{*}{Metrics}& \multirow{4}{*}{Q} & \multicolumn{8}{c}{Orbit types}\\
			  \cmidrule(lr){3-10}
                & &   \multicolumn{4}{c}{D-IOD$_{\rm{refine}}$} &   \multicolumn{4}{c}{D-IOD$_{\rm{end-to-end}}$} \\
                \cmidrule(lr){3-6}\cmidrule(lr){7-10}
			    & &   A & B & C & D &   A & B & C & D \\
			 \cmidrule(lr){1-10}
			  $\Delta \mathbf{u}$ (pixel) &Q1 &0.33 &0.54 &0.53 &0.48 &0.36 &0.4 &0.41 &0.37 \\
            &Q2 &0.78 &1.25 &1.33 &1.14 &0.76 &0.83 &0.87 &0.82 \\
            &Q3 &1.9 &3.05 &3.56 &2.89 &1.75 &1.73 &1.89 &1.7 \\
            \cmidrule(lr){1-10}
            $\Delta r_p$ (km) &Q1 &2.9 &6.03 &3.93 &2.71 &3.11 &4.64 &2.43 &2.35 \\
            &Q2 &9 &32.21 &17.62 &14.63 &8.91 &20.3 &10.67 &10.84 \\
            &Q3 &30.71 &160.21 &100.9 &56.66 &30.47 &81.88 &43.78 &39.55 \\
            \cmidrule(lr){1-10}
            $\Delta e$ &Q1 &3.12e-4 &1.22e-3 &1.38e-3 &1.08e-3 &3.01e-4 &9.95e-4 &8.17e-4 &9.10e-4 \\
            &Q2 &1.04e-3 &5.18e-3 &5.01e-3 &3.79e-3 &9.63e-4 &3.21e-3 &3.01e-3 &2.93e-3 \\
            &Q3 &4.88e-3 &2.09e-2 &2.28e-2 &1.42e-2 &4.45e-3 &1.05e-2 &1.03e-2 &1.07e-2 \\
            \cmidrule(lr){1-10}
            $\Delta i \:(\circ)$ &Q1 &3.73e-3 &1.06e-2 &8.76e-3 &9.93e-3 &3.86e-3 &5.81e-3 &6.76e-3 &4.71e-3 \\
            &Q2 &1.59e-2 &4.40e-2 &4.39e-2 &3.81e-2 &1.62e-2 &2.32e-2 &2.77e-2 &2.33e-2 \\
            &Q3 &7.16e-2 &2.02e-1 &1.97e-1 &1.64e-1 &6.40e-2 &7.41e-2 &8.44e-2 &9.19e-2 \\
            \cmidrule(lr){1-10}
            $\Delta \Omega\:(\circ)$ &Q1 &6.17e-3 &1.80e-2 &1.38e-2 &1.77e-2 &7.43e-3 &1.22e-2 &1.49e-2 &1.12e-2 \\
            &Q2 &2.98e-2 &8.86e-2 &6.61e-2 &7.35e-2 &2.54e-2 &4.63e-2 &5.32e-2 &4.62e-2 \\
            &Q3 &1.26e-1 &3.62e-1 &3.79e-1 &2.99e-1 &1.61e-1 &2.10e-1 &2.26e-1 &1.68e-1 \\
            \cmidrule(lr){1-10}
            $\Delta \omega\:(\circ)$ &Q1 &- &1.36 &0.44 &0.25 &- &0.88 &0.27 &0.21 \\
            &Q2 &- &5.53 &1.79 &0.98 &- &3.03 &1.17 &0.77 \\
            &Q3 &- &23.53 &7.07 &3.76 &- &10.45 &3.73 &2.04 \\
            \cmidrule(lr){1-10}
            $\Delta f\:(\circ)$ &Q1 &- &1.4 &0.45 &0.26 &- &0.85 &0.31 &0.2 \\
            &Q2 &- &5.73 &1.78 &0.98 &- &3.08 &1.2 &0.74 \\
            &Q3 &- &27.17 &7.61 &3.77 &- &11.03 &3.99 &2 \\
			  \bottomrule
		\end{tabular}
	\end{center}
\end{table*}

\begin{table*}
	\begin{center}
		\caption{Time interval experiment - time interval = 120 seconds.}\label{Tab:time_exp_2}
		\begin{tabular}{cccccccccc}
			\toprule
			  \multirow{4}{*}{Metrics}& \multirow{4}{*}{Q} & \multicolumn{8}{c}{Orbit types}\\
			  \cmidrule(lr){3-10}
                & &   \multicolumn{4}{c}{D-IOD$_{\rm{refine}}$} &   \multicolumn{4}{c}{D-IOD$_{\rm{end-to-end}}$} \\
                \cmidrule(lr){3-6}\cmidrule(lr){7-10}
			    & &   A & B & C & D &   A & B & C & D \\
			 \cmidrule(lr){1-10}
			  $\Delta \mathbf{u}$ (pixel) &Q1 &0.43 &0.58 &0.55 &0.54 &0.36 &0.4 &0.37 &0.41 \\
                &Q2 &1.04 &1.4 &1.32 &1.29 &0.76 &0.83 &0.82 &0.91 \\
                &Q3 &2.39 &3.44 &3.26 &3.14 &1.75 &1.73 &1.81 &2.04 \\
                \cmidrule(lr){1-10}
                $\Delta r_p$ (km) &Q1 &1.86 &4.24 &2.21 &1.41 &3.11 &4.64 &1.26 &1.42 \\
                &Q2 &6.04 &21.71 &12.2 &9.6 &8.91 &20.3 &6.47 &6.12 \\
                &Q3 &25.41 &107.32 &62.35 &49.66 &30.47 &81.88 &27.78 &22.08 \\
                \cmidrule(lr){1-10}
                $\Delta e$ &Q1 &2.00e-4 &6.87e-4 &6.44e-4 &6.03e-4 &3.01e-4 &9.95e-4 &4.46e-4 &5.74e-4 \\
                &Q2 &7.90e-4 &3.77e-3 &3.03e-3 &2.84e-3 &9.63e-4 &3.21e-3 &1.45e-3 &1.96e-3 \\
                &Q3 &3.55e-3 &1.43e-2 &1.24e-2 &1.42e-2 &4.45e-3 &1.05e-2 &4.91e-3 &5.88e-3 \\
                \cmidrule(lr){1-10}
                $\Delta i \:(\circ)$ &Q1 &3.04e-3 &4.41e-3 &4.19e-3 &4.85e-3 &3.86e-3 &5.81e-3 &2.67e-3 &2.93e-3 \\
                &Q2 &1.49e-2 &2.64e-2 &2.26e-2 &2.73e-2 &1.62e-2 &2.32e-2 &1.31e-2 &1.46e-2 \\
                &Q3 &5.47e-2 &1.40e-1 &9.11e-2 &1.18e-1 &6.40e-2 &7.41e-2 &4.99e-2 &6.91e-2 \\
                \cmidrule(lr){1-10}
                $\Delta \Omega\:(\circ)$ &Q1 &4.53e-3 &8.86e-3 &8.73e-3 &8.09e-3 &7.43e-3 &1.22e-2 &4.38e-3 &7.11e-3 \\
                &Q2 &2.07e-2 &5.14e-2 &4.95e-2 &4.68e-2 &2.54e-2 &4.63e-2 &1.85e-2 &3.36e-2 \\
                &Q3 &8.43e-2 &2.13e-1 &2.26e-1 &2.17e-1 &1.61e-1 &2.10e-1 &8.48e-2 &1.36e-1 \\
                \cmidrule(lr){1-10}
                $\Delta \omega\:(\circ)$ &Q1 &- &0.66 &0.21 &0.12 &- &0.88 &0.16 &0.11 \\
                &Q2 &- &3.69 &0.9 &0.75 &- &3.03 &0.64 &0.48 \\
                &Q3 &- &16.24 &4.75 &3.22 &- &10.45 &2 &1.42 \\
                \cmidrule(lr){1-10}
                $\Delta f\:(\circ)$ &Q1 &- &0.66 &0.21 &0.11 &- &0.85 &0.15 &0.12 \\
                &Q2 &- &3.69 &0.95 &0.84 &- &3.08 &0.64 &0.49 \\
                &Q3 &- &16.98 &4.72 &3.26 &- &11.03 &2.04 &1.42 \\
			  \bottomrule
		\end{tabular}
	\end{center}
\end{table*}

\begin{table*}
	\begin{center}
		\caption{Time interval experiment - time interval = 240 seconds.}\label{Tab:time_exp_3}
		\begin{tabular}{cccccccccc}
			\toprule
			  \multirow{4}{*}{Metrics}& \multirow{4}{*}{Q} & \multicolumn{8}{c}{Orbit types}\\
			  \cmidrule(lr){3-10}
                & &   \multicolumn{4}{c}{D-IOD$_{\rm{refine}}$} &   \multicolumn{4}{c}{D-IOD$_{\rm{end-to-end}}$} \\
                \cmidrule(lr){3-6}\cmidrule(lr){7-10}
			    & &   A & B & C & D &   A & B & C & D \\
			 \cmidrule(lr){1-10}
			$\Delta \mathbf{u}$ (pixel) &Q1 &0.56 &0.55 &0.67 &0.52 &0.49 &0.4 &0.4 &0.42 \\
            &Q2 &1.32 &1.44 &1.64 &1.32 &1.14 &0.93 &0.88 &0.99 \\
            &Q3 &3.45 &3.21 &3.83 &3.35 &2.56 &1.96 &1.79 &2.15 \\
            \cmidrule(lr){1-10}
            $\Delta r_p$ (km) &Q1 &1.27 &1.51 &1.48 &0.89 &2.24 &3.08 &0.73 &0.86 \\
            &Q2 &4.71 &11.52 &7.86 &4.08 &6.14 &13.7 &2.89 &3.17 \\
            &Q3 &21.84 &57.67 &38.82 &23.51 &20.38 &58.64 &11.19 &11.87 \\
            \cmidrule(lr){1-10}
            $\Delta e$ &Q1 &1.35e-4 &3.05e-4 &4.00e-4 &3.14e-4 &2.31e-4 &5.79e-4 &2.11e-4 &2.70e-4 \\
            &Q2 &6.82e-4 &1.66e-3 &2.00e-3 &1.28e-3 &9.29e-4 &2.36e-3 &7.29e-4 &9.77e-4 \\
            &Q3 &2.85e-3 &8.52e-3 &8.23e-3 &6.10e-3 &3.33e-3 &8.63e-3 &2.83e-3 &3.60e-3 \\
            \cmidrule(lr){1-10}
            $\Delta i \:(\circ)$ &Q1 &1.87e-3 &1.90e-3 &2.39e-3 &2.56e-3 &2.89e-3 &4.96e-3 &1.64e-3 &1.88e-3 \\
            &Q2 &9.86e-3 &1.11e-2 &1.66e-2 &1.06e-2 &1.41e-2 &1.72e-2 &7.93e-3 &8.63e-3 \\
            &Q3 &4.54e-2 &5.51e-2 &6.81e-2 &6.31e-2 &4.93e-2 &6.97e-2 &2.72e-2 &3.36e-2 \\
            \cmidrule(lr){1-10}
            $\Delta \Omega\:(\circ)$ &Q1 &3.62e-3 &3.47e-3 &5.39e-3 &3.71e-3 &4.79e-3 &7.47e-3 &2.49e-3 &3.07e-3 \\
            &Q2 &1.78e-2 &2.24e-2 &3.05e-2 &1.90e-2 &2.52e-2 &3.33e-2 &1.43e-2 &1.51e-2 \\
            &Q3 &8.85e-2 &1.01e-1 &1.36e-1 &1.04e-1 &1.19e-1 &1.54e-1 &6.18e-2 &7.96e-2 \\
            \cmidrule(lr){1-10}
            $\Delta \omega\:(\circ)$ &Q1 &- &0.21 &0.14 &0.06 &- &0.6 &0.07 &0.05 \\
            &Q2 &- &1.75 &0.61 &0.34 &- &2.01 &0.25 &0.24 \\
            &Q3 &- &7.93 &2.35 &1.59 &- &8.35 &0.82 &0.75 \\
            \cmidrule(lr){1-10}
            $\Delta f\:(\circ)$ &Q1 &- &0.25 &0.12 &0.06 &- &0.62 &0.07 &0.05 \\
            &Q2 &- &1.73 &0.62 &0.35 &- &2 &0.24 &0.23 \\
            &Q3 &- &8.46 &2.37 &1.64 &- &8.35 &0.85 &0.76 \\
			  \bottomrule
		\end{tabular}
	\end{center}
\end{table*}

\begin{table*}
	\begin{center}
		\caption{SNR experiment - SNR = 4.}\label{Tab:SNR_1}
		\begin{tabular}{cccccccccc}
			\toprule
			  \multirow{4}{*}{Metrics}& \multirow{4}{*}{Q} & \multicolumn{8}{c}{Orbit types}\\
			  \cmidrule(lr){3-10}
                & &   \multicolumn{4}{c}{D-IOD$_{\rm{refine}}$} &   \multicolumn{4}{c}{D-IOD$_{\rm{end-to-end}}$} \\
                \cmidrule(lr){3-6}\cmidrule(lr){7-10}
			    & &   A & B & C & D &   A & B & C & D \\
			 \cmidrule(lr){1-10}
			$\Delta \mathbf{u}$ (pixel) &Q1 &0.43 &0.58 &0.55 &0.54 &0.49 &0.4 &0.37 &0.41 \\
                &Q2 &1.04 &1.4 &1.32 &1.29 &1.14 &0.93 &0.82 &0.91 \\
                &Q3 &2.39 &3.44 &3.26 &3.14 &2.56 &1.96 &1.81 &2.04 \\
                \cmidrule(lr){1-10}
                $\Delta r_p$ (km) &Q1 &1.86 &4.24 &2.21 &1.41 &2.24 &3.08 &1.26 &1.42 \\
                &Q2 &6.04 &21.71 &12.2 &9.6 &6.14 &13.7 &6.47 &6.12 \\
                &Q3 &25.41 &107.32 &62.35 &49.66 &20.38 &58.64 &27.78 &22.08 \\
                \cmidrule(lr){1-10}
                $\Delta e$ &Q1 &2.00e-4 &6.87e-4 &6.44e-4 &6.03e-4 &2.31e-4 &5.79e-4 &4.46e-4 &5.74e-4 \\
                &Q2 &7.90e-4 &3.77e-3 &3.03e-3 &2.84e-3 &9.29e-4 &2.36e-3 &1.45e-3 &1.96e-3 \\
                &Q3 &3.55e-3 &1.43e-2 &1.24e-2 &1.42e-2 &3.33e-3 &8.63e-3 &4.91e-3 &5.88e-3 \\
                \cmidrule(lr){1-10}
                $\Delta i \:(\circ)$ &Q1 &3.04e-3 &4.41e-3 &4.19e-3 &4.85e-3 &2.89e-3 &4.96e-3 &2.67e-3 &2.93e-3 \\
                &Q2 &1.49e-2 &2.64e-2 &2.26e-2 &2.73e-2 &1.41e-2 &1.72e-2 &1.31e-2 &1.46e-2 \\
                &Q3 &5.47e-2 &1.40e-1 &9.11e-2 &1.18e-1 &4.93e-2 &6.97e-2 &4.99e-2 &6.91e-2 \\
                \cmidrule(lr){1-10}
                $\Delta \Omega\:(\circ)$ &Q1 &4.53e-3 &8.86e-3 &8.73e-3 &8.09e-3 &4.79e-3 &7.47e-3 &4.38e-3 &7.11e-3 \\
                &Q2 &2.07e-2 &5.14e-2 &4.95e-2 &4.68e-2 &2.52e-2 &3.33e-2 &1.85e-2 &3.36e-2 \\
                &Q3 &8.43e-2 &2.13e-1 &2.26e-1 &2.17e-1 &1.19e-1 &1.54e-1 &8.48e-2 &1.36e-1 \\
                \cmidrule(lr){1-10}
                $\Delta \omega\:(\circ)$ &Q1 &- &0.66 &0.21 &0.12 &- &0.6 &0.16 &0.11 \\
                &Q2 &- &3.69 &0.9 &0.75 &- &2.01 &0.64 &0.48 \\
                &Q3 &- &16.24 &4.75 &3.22 &- &8.35 &2 &1.42 \\
                \cmidrule(lr){1-10}
                $\Delta f\:(\circ)$ &Q1 &- &0.66 &0.21 &0.11 &- &0.62 &0.15 &0.12 \\
                &Q2 &- &3.69 &0.95 &0.84 &- &2 &0.64 &0.49 \\
                &Q3 &- &16.98 &4.72 &3.26 &- &8.35 &2.04 &1.42 \\
			  \bottomrule
		\end{tabular}
	\end{center}
\end{table*}

\begin{table*}
	\begin{center}
		\caption{SNR experiment - SNR = 3.}\label{Tab:SNR_2}
		\begin{tabular}{cccccccccc}
			\toprule
			  \multirow{4}{*}{Metrics}& \multirow{4}{*}{Q} & \multicolumn{8}{c}{Orbit types}\\
			  \cmidrule(lr){3-10}
                & &   \multicolumn{4}{c}{D-IOD$_{\rm{refine}}$} &   \multicolumn{4}{c}{D-IOD$_{\rm{end-to-end}}$} \\
                \cmidrule(lr){3-6}\cmidrule(lr){7-10}
			    & &   A & B & C & D &   A & B & C & D \\
			 \cmidrule(lr){1-10}
			$\Delta \mathbf{u}$ (pixel) &Q1 &0.55 &0.76 &0.72 &0.75 &0.51 &0.49 &0.46 &0.45 \\
                &Q2 &1.21 &1.8 &1.74 &1.76 &1.17 &1.06 &0.99 &1.05 \\
                &Q3 &2.96 &3.89 &4.26 &3.98 &2.85 &2.16 &2.02 &2.09 \\
                \cmidrule(lr){1-10}
                $\Delta r_p$ (km) &Q1 &2.03 &5.95 &3.74 &2.35 &2.21 &3.56 &1.82 &1.5 \\
                &Q2 &7.57 &26.77 &14.48 &13.97 &6.3 &15.97 &7.34 &6.28 \\
                &Q3 &27.4 &120.67 &76.41 &62.11 &20.1 &54.39 &27.03 &25.69 \\
                \cmidrule(lr){1-10}
                $\Delta e$ &Q1 &2.50e-4 &9.90e-4 &9.34e-4 &9.31e-4 &2.68e-4 &5.49e-4 &5.36e-4 &6.04e-4 \\
                &Q2 &1.09e-3 &4.59e-3 &4.42e-3 &3.94e-3 &9.77e-4 &2.74e-3 &1.77e-3 &2.15e-3 \\
                &Q3 &4.07e-3 &1.76e-2 &1.68e-2 &1.53e-2 &3.08e-3 &8.20e-3 &6.15e-3 &6.36e-3 \\
                \cmidrule(lr){1-10}
                $\Delta i \:(\circ)$ &Q1 &3.27e-3 &6.21e-3 &5.62e-3 &7.25e-3 &3.00e-3 &4.79e-3 &4.54e-3 &3.50e-3 \\
                &Q2 &1.81e-2 &2.95e-2 &2.99e-2 &3.25e-2 &1.35e-2 &1.86e-2 &1.48e-2 &1.70e-2 \\
                &Q3 &6.50e-2 &1.48e-1 &1.13e-1 &1.28e-1 &5.31e-2 &6.89e-2 &5.71e-2 &7.07e-2 \\
                \cmidrule(lr){1-10}
                $\Delta \Omega\:(\circ)$ &Q1 &5.41e-3 &1.68e-2 &1.32e-2 &1.37e-2 &5.67e-3 &7.70e-3 &6.07e-3 &7.53e-3 \\
                &Q2 &2.58e-2 &6.48e-2 &7.40e-2 &6.33e-2 &2.76e-2 &3.47e-2 &2.19e-2 &3.40e-2 \\
                &Q3 &1.07e-1 &2.58e-1 &2.70e-1 &2.61e-1 &1.22e-1 &1.58e-1 &1.13e-1 &1.43e-1 \\
                \cmidrule(lr){1-10}
                $\Delta \omega\:(\circ)$ &Q1 &- &0.95 &0.3 &0.21 &- &0.59 &0.2 &0.13 \\
                &Q2 &- &4.75 &1.47 &0.97 &- &2.23 &0.73 &0.46 \\
                &Q3 &- &16.72 &5.05 &3.71 &- &8.1 &2.37 &1.54 \\
                \cmidrule(lr){1-10}
                $\Delta f\:(\circ)$ &Q1 &- &0.92 &0.33 &0.22 &- &0.59 &0.2 &0.13 \\
                &Q2 &- &4.8 &1.45 &1.01 &- &2.3 &0.72 &0.48 \\
                &Q3 &- &17.84 &5.03 &3.77 &- &8.1 &2.55 &1.49 \\
			  \bottomrule
		\end{tabular}
	\end{center}
\end{table*}

\begin{table*}
	\begin{center}
		\caption{SNR experiment - SNR = 2.}\label{Tab:SNR_3}
		\begin{tabular}{cccccccccc}
			\toprule
			  \multirow{4}{*}{Metrics}& \multirow{4}{*}{Q} & \multicolumn{8}{c}{Orbit types}\\
			  \cmidrule(lr){3-10}
                & &   \multicolumn{4}{c}{D-IOD$_{\rm{refine}}$} &   \multicolumn{4}{c}{D-IOD$_{\rm{end-to-end}}$} \\
                \cmidrule(lr){3-6}\cmidrule(lr){7-10}
			    & &   A & B & C & D &   A & B & C & D \\
			 \cmidrule(lr){1-10}
			$\Delta \mathbf{u}$ (pixel) &Q1 &0.76 &0.97 &0.87 &1.04 &0.7 &0.58 &0.52 &0.55 \\
            &Q2 &1.74 &2.1 &2.11 &2.33 &1.47 &1.29 &1.17 &1.22 \\
            &Q3 &4.42 &4.99 &5.27 &5.66 &3.19 &2.55 &2.4 &2.51 \\
            \cmidrule(lr){1-10}
            $\Delta r_p$ (km) &Q1 &3.22 &7.8 &4.5 &3.28 &2.94 &4.46 &2.04 &1.55 \\
            &Q2 &11.19 &40.88 &22.36 &17.61 &8.05 &17.65 &8.17 &7.2 \\
            &Q3 &40.93 &159.22 &80.32 &63.17 &28.59 &70.07 &30.92 &25.47 \\
            \cmidrule(lr){1-10}
            $\Delta e$ &Q1 &3.90e-4 &1.63e-3 &1.39e-3 &1.79e-3 &4.02e-4 &8.80e-4 &7.02e-4 &6.06e-4 \\
            &Q2 &1.65e-3 &5.95e-3 &6.05e-3 &5.39e-3 &1.44e-3 &2.79e-3 &2.19e-3 &2.25e-3 \\
            &Q3 &5.17e-3 &2.07e-2 &1.90e-2 &1.77e-2 &4.66e-3 &1.06e-2 &6.82e-3 &7.71e-3 \\
            \cmidrule(lr){1-10}
            $\Delta i \:(\circ)$ &Q1 &4.94e-3 &1.05e-2 &8.23e-3 &9.94e-3 &3.68e-3 &6.81e-3 &4.76e-3 &5.27e-3 \\
            &Q2 &2.84e-2 &4.32e-2 &3.90e-2 &4.85e-2 &1.64e-2 &2.75e-2 &1.76e-2 &1.85e-2 \\
            &Q3 &8.14e-2 &1.49e-1 &1.29e-1 &1.63e-1 &7.14e-2 &7.39e-2 &4.89e-2 &7.21e-2 \\
            \cmidrule(lr){1-10}
            $\Delta \Omega\:(\circ)$ &Q1 &9.62e-3 &2.02e-2 &1.96e-2 &1.98e-2 &6.78e-3 &1.08e-2 &6.83e-3 &9.83e-3 \\
            &Q2 &4.30e-2 &8.93e-2 &7.89e-2 &8.64e-2 &3.30e-2 &4.35e-2 &2.84e-2 &3.61e-2 \\
            &Q3 &1.49e-1 &3.29e-1 &3.26e-1 &3.46e-1 &1.54e-1 &1.75e-1 &9.91e-2 &1.48e-1 \\
            \cmidrule(lr){1-10}
            $\Delta \omega\:(\circ)$ &Q1 &- &1.45 &0.46 &0.36 &- &0.85 &0.24 &0.15 \\
            &Q2 &- &5.95 &1.73 &1.43 &- &2.65 &0.74 &0.51 \\
            &Q3 &- &20.83 &5.94 &4.31 &- &8.87 &2.41 &1.66 \\
            \cmidrule(lr){1-10}
            $\Delta f\:(\circ)$ &Q1 &- &1.41 &0.48 &0.33 &- &0.83 &0.25 &0.16 \\
            &Q2 &- &6.21 &1.85 &1.39 &- &2.63 &0.79 &0.52 \\
            &Q3 &- &25.3 &6.15 &4.62 &- &9.29 &2.5 &1.73 \\
			  \bottomrule
		\end{tabular}
	\end{center}
\end{table*}

%% Bibliography
%% Author year style

\end{document}